%% file: MAIN.tex
\documentclass[manuscript,screen]{acmart}
\AtBeginDocument{
  }

\setcopyright{acmlicensed}
\copyrightyear{2023} 
\acmYear{2023}
\acmDOI{10.1145/XXXXXXX.XXXXXXX} 
\acmISBN{978-1-4503-XXXX-X/23/06} 

\usepackage[utf8]{inputenc} 
\usepackage[T1]{fontenc}    
\usepackage{url}            
\usepackage{booktabs}       
\usepackage{amsfonts}       
\usepackage{nicefrac}       
\usepackage{microtype}      
\usepackage{subfig}         

\usepackage{xcolor}         
\usepackage{graphicx}       
\usepackage{textcomp}       
\usepackage{algorithm}      
\usepackage{algpseudocode}  

\usepackage{amssymb}        
\usepackage{mathtools}      
\usepackage{amsthm}         
\usepackage{csquotes}       
\usepackage{multirow}       
\usepackage{multicol}       
\usepackage{color,soul}     
\usepackage{wrapfig}        
\usepackage{float}          
\usepackage{lipsum}         
\usepackage{cleveref}       
\usepackage{tikz, tikz-cd}  
\usepackage{adjustbox}      
\usepackage{hyperref}       

 \usepackage{listings}
\usepackage{xcolor}

\usetikzlibrary{shapes.geometric, positioning, trees}

\newcommand{\wh}{\widehat}
\newcommand{\wt}{\widetilde}
\newcommand{\BZ}{\mathbb{Z}}
\newcommand{\R}{\mathbb{R}}
\newcommand{\G}{\mathcal{G}}
\newcommand{\C}{\mathcal{C}}
\newcommand{\B}{\mathcal{B}}
\newcommand{\N}{\mathcal{N}}

\newcommand{\h}{\mathcal{H}}
\newcommand{\I}{\mathcal{I}}
\newcommand{\X}{\mathcal{X}}
\newcommand{\s}{\mathcal{S}}
\newcommand{\T}{\mathcal{T}}
\newcommand{\D}{\mathcal{D}}
\newcommand{\F}{\mathcal{F}}
\newcommand{\Y}{\mathcal{Y}}
\newcommand{\Z}{\mathcal{Z}}
\newcommand{\K}{\mathcal{K}}
\newcommand{\VR}{\mathcal{R}}

\newcommand{\V}{\mathcal{V}}
\newcommand{\E}{\mathcal{E}}

\newcommand{\W}{\mathcal{W}}

\newcommand{\e}{\epsilon}
\newcommand{\p}{\mathcal{P}}

\newcommand{\PD}{\rm{PD}}
\newcommand{\PB}{\rm{PB}}

\begin{document}

\title{Topological Methods in Machine Learning: A Tutorial for Practitioners}

\author{Baris Coskunuzer}
\email{coskunuz@utdallas.edu}
\orcid{0000-0001-7462-8819}
\affiliation{%
  \institution{University of Texas at Dallas}
  \city{Richardson}
  \state{TX}
  \country{USA}
}
\author{C\"{u}neyt G\"{u}rcan Ak\c{c}ora}
\email{cuney.akcora@ucf.edu}
\orcid{0000-0002-2882-6950}
\affiliation{%
  \institution{University of Central Florida}
  \city{Orlando}
  \state{FL}
  \country{USA}}

\renewcommand{\shortauthors}{Coskunuzer-Akcora}

\begin{abstract}
Topological Machine Learning (TML) is an emerging field that leverages techniques from algebraic topology to analyze complex data structures in ways that traditional machine learning methods may not capture. This tutorial provides a comprehensive introduction to two key TML techniques, persistent homology and the Mapper algorithm, with an emphasis on practical applications. Persistent homology captures multi-scale topological features such as clusters, loops, and voids, while the Mapper algorithm creates an interpretable graph summarizing high-dimensional data. To enhance accessibility, we adopt a data-centric approach, enabling readers to gain hands-on experience applying these techniques to relevant tasks. We provide step-by-step explanations, implementations, hands-on examples, and case studies to demonstrate how these tools can be applied to real-world problems. The goal is to equip researchers and practitioners with the knowledge and resources to incorporate TML into their work, revealing insights often hidden from conventional machine learning methods. The tutorial code is available at \url{https://github.com/cakcora/TopologyForML}.
\end{abstract}



\received{20 September 2024}
\received[revised]{12 March 20xx}
\received[accepted]{5 June 20xx}

\maketitle

\tableofcontents

\section{Introduction} \label{sec:intro}

\input{sections/1-intro}

\section{Background} \label{sec:background}
\input{sections/2-background}

\section{Persistent Homology} \label{sec:PH}
\input{sections/3-PH}

\section{Multiparameter Persistence} \label{sec:MP}
\input{sections/4-MP}

\section{Mapper} \label{sec:mapper}
\input{sections/5-Mapper}

\section{Applications} \label{sec:applications}
\input{sections/6-Applications}

\section{Future Directions} \label{sec:future}
\input{sections/7-Future}

\section{Conclusion} \label{sec:conclusion}
In this tutorial, we have introduced the key concepts of topological methods, particularly focusing on persistent homology and the mapper algorithm, and demonstrated their practical application in machine learning tasks. By providing a clear and accessible roadmap, we aimed to equip readers with the tools and understanding necessary to integrate TDA techniques into their research workflows. The strength of these methods lies in their ability to capture and quantify intricate, multi-scale topological features that are often missed by traditional ML approaches.

As demonstrated through various case studies, including cancer diagnosis, shape recognition, and drug discovery, TDA offers unique insights and interpretability, which are increasingly valuable in today's era of complex data. The integration of persistent homology for feature extraction and mapper for intuitive data visualization opens new pathways for researchers to explore the underlying structures of their data.

Looking forward, the continued development of software libraries, scalable algorithms, and more interpretable models will play a crucial role in making topological machine learning even more accessible to a broader audience. The future directions we highlighted, such as topological deep learning and automated TDA models, provide exciting opportunities for further advancements in the field.

We hope this tutorial serves as a foundational resource for those new to TDA and inspires further exploration of topological methods in machine learning and beyond. By embracing these techniques, researchers can unlock novel insights and push the boundaries of what is possible in data analysis.

\begin{acks}
This work was partially supported by the National Science Foundation under grants  DMS-2202584, 2229417, and DMS-2220613 and by the Simons Foundation under grant \# 579977.
The authors acknowledge the \href{http://www.tacc.utexas.edu}{Texas Advanced Computing Center} (TACC) at The University of Texas at Austin for providing computational resources that have contributed to the research results reported within this paper. 
\end{acks}

\bibliographystyle{ACM-Reference-Format}
\bibliography{main}

\clearpage

\appendix
\input{sections/8-appendix}

\end{document}

%% file: sections/1-intro.tex
As the complexity of datasets has grown in recent years, topological methods have emerged as powerful complements to state-of-the-art ML methods. The advent of ML has revolutionized the way we analyze and interpret complex data, yet there remain challenges in capturing the intrinsic topological structures inherent in such data. Traditional ML techniques, while powerful, often fall short in identifying and leveraging these structures, leading to the potential loss of valuable insights. \textit{Topological Machine Learning} (TML) bridges this gap by integrating concepts from algebraic topology into ML workflows, enabling the discovery of patterns and features that are otherwise elusive. Despite this utility, much of the existing literature on topological methods in ML is highly technical, making it challenging for newcomers to grasp the direct connections to practical applications. 

In this tutorial, we introduce the fundamental concepts of topological methods to the machine learning (ML) community and a broader audience interested in integrating these novel approaches into their research. No prior knowledge of topology or ML is required. Our primary aim is to address this pressing need by providing a practical guide for non-experts looking to employ topological techniques in various ML contexts. To maintain accessibility, we will simplify the exposition and offer references to more detailed technical resources for those interested in further exploration.

In this paper, we teach two cornerstone techniques of TML: \textit{Persistent Homology} and \textit{Mapper algorithm}, and their effective utilization in ML. Persistent homology offers a robust, multi-scale analysis of topological features, allowing researchers to detect and quantify structures such as clusters, loops, and voids across different scales within the data. This capability is particularly beneficial for understanding the intricate relationships and hierarchies that may exist in complex datasets. From an ML perspective, PH offers \textit{a great feature extraction method} for complex datasets, which was impossible with other methods. In this part, we will focus on this aspect of PH, giving hands-on instructions with illustrations on deriving effective topological vectors from complex data. On the other hand, the Mapper algorithm complements this by providing a visual and interpretable summary of high-dimensional data. By constructing a summary graph that mirrors the underlying topology of data, the Mapper algorithm facilitates the exploration and interpretation of data in an intuitive and informative way. This technique is instrumental in uncovering data's geometric and topological essence, making it accessible for practical analysis.

Throughout the paper, we provide comprehensive explanations and step-by-step implementations of these techniques, supported by case studies spanning diverse applications such as cancer diagnosis, shape recognition, genotyping, and drug discovery. We aim to equip researchers and practitioners with the necessary knowledge and tools to integrate TML techniques into their studies, thereby unlocking new avenues for discovery and innovation. By demonstrating the practical utility of persistent homology and the Mapper algorithm, we highlight their potential to reveal insights that traditional methods may overlook, ultimately advancing the field of Machine Learning. 

We note that this paper is not intended to be a survey of recent advances in topological data analysis but rather a tutorial aimed at introducing the fundamentals of the topic to ML practitioners. For comprehensive surveys in topological data analysis, refer to~\cite{hensel2021survey,pun2022persistent,chazal2021introduction,skaf2022topological}. For in-depth discussions of these topics, see the excellent textbooks on TDA and computational topology~\cite{dey2022computational,edelsbrunner2022computational}.

\subsection{Roadmap for the Tutorial}

We recommend reading the entire tutorial for a complete understanding, but readers may skip sections not pertinent to their needs. Here, we give a quick overview of the paper's structure.

In~\Cref{sec:background}, we provide the essential topological background needed to follow the concepts discussed throughout the rest of the paper. We aim to introduce these topological concepts, help build an intuition for TDA approaches, and demonstrate how they can be adapted and applied to various needs. For those unfamiliar with topology, we strongly encourage reading our introductory crash course in~\Cref{sec:topology}. When discussing homology, we start with a non-technical, brief overview~(\Cref{sec:TLDR}), which should suffice for following the rest of the paper. For readers interested in more technical details of homology computation, we offer a more in-depth explanation in~\Cref{sec:homology2}.

In~\Cref{sec:PH}, we introduce \textit{Persistent Homology} (PH) in three key sections: constructing filtrations (\Cref{sec:filtrations}), deriving persistence diagrams (PD) (\Cref{sec:PD}), and applying PDs to ML tasks, including vectorization (\Cref{sec:vectorization}) and neural networks (\Cref{sec:vecNN}). The filtration process is tailored to different data formats, as methods differ significantly based on the type of data—whether it’s point clouds (\Cref{sec:point_cloud}), images (\Cref{sec:image}), or networks (\Cref{sec:graph}). If you focus on a specific data type, you can skip sections on other formats. In each subsection, we also discuss the hyperparameter selection process. Lastly, \Cref{sec:PH_software} covers available software for PH. 

In \Cref{sec:MP}, we give a brief introduction to a niche subfield, \textit{Multiparameter Persistence} (MP), an effective extension of PH. Again, we outline how to tailor the method to specific data formats for multifiltrations, i.e., point clouds (\Cref{sec:MP-point_cloud}), images~(\Cref{sec:MP-image}), and networks~(\Cref{sec:MP-graph}). Next, we outline the state-of-the-art methods on how to integrate MP information in ML pipelines~(\Cref{sec:MP-vec}).

In \Cref{sec:mapper}, we begin with a friendly introduction to the Mapper method~(\Cref{sec:mapper-PC}), an effective TML tool for unsupervised learning. We then cover hyperparameter selection for Mapper in \Cref{sec:mapperParams}, which is key for applications. Although the original Mapper algorithm is intended for point clouds, \Cref{sec:mapper-graph} explores recent advancements that extend its application to images and networks. Lastly, we provide an overview of available software libraries for Mapper in \Cref{sec:mapper-software}.

In \Cref{sec:applications}, we summarize five real-life applications of these methods from published works, namely shape recognition for point clouds (\Cref{sec:shape}), anomaly forecasting for transaction networks~(\Cref{sec:crypto}), cancer detection from histopathological images~(\Cref{sec:histo}), computer-aided drug discovery~(\Cref{sec:drug}), and cancer genotyping from RNA sequencing~(\Cref{sec:mapper-app}).

Finally, in~\Cref{sec:future}, we outline potential future directions to advance TML methods, aiming to improve their practical use in ML and discuss strategies for broadening the application of topological methods to new and emerging fields.

%% file: sections/2-background.tex
In this section, we provide some background that will be used later. We first introduce several mathematical concepts that will be used in the second part, where we describe homology, which is essential for introducing the methods in subsequent sections.

\subsection{A Crash Course on Topology} \label{sec:topology}

Topology, a core discipline in mathematics, studies the properties of shapes and spaces that remain unchanged under continuous transformations such as stretching, crumpling, and bending, but not tearing or gluing. In machine learning, topology provides a powerful framework for examining complex data structures flexibly and intuitively. To provide clarity, we begin by defining several key terms essential for following the paper.

\subsubsection{Topological Space} From a mathematical perspective, topology refers to \textit{the structure of a set}. For instance, consider the set of real numbers between $0$ and $1$, i.e., $\X = \{x\in\R \mid 0 \leq x \leq 1\}$. Many readers might immediately consider the closed interval $[0,1]$, whose shape resembles a stick. However, $\X$ is merely a set with no defined structure yet. This means we do not know which points in $\X$ are close to or far from others, as there is no concept of neighborhoods.

For example, if we define a distance (metric) on $\X$ such as $d_1(x, y) = 1$ for $x \neq y$ and $d_1(x, y) = 0$ for $x = y$, the "shape" of the set $\X$ would be entirely different, resembling an infinite number of points dispersed in a very high-dimensional space. Conversely, if we use a metric like $d_2(x, y) = |x - y|$ on $\X$, we retrieve our familiar closed interval $[0,1]$, a stick of length $1$. Notice that due to the different metrics used, the neighborhood structures and shapes of $(\X, d_1)$ and $(\X, d_2)$ are completely different. This brings us to the first principle in Topology: \textit{the topology of a set} is defined as the complete neighborhood information on the set.  

The principle is not without exceptions: a significant research area called \textit{point-set topology} studying topological spaces that do not necessarily have a metric~\cite{lopez2024point,goubault2013non}. However, these topologies are beyond our scope. Thus, throughout the paper, a \textit{topological space} refers to a set $\X$ (or dataset) equipped with a distance $d(\cdot,\cdot)$, forming what is known as a \textit{metric space} $(\X,d)$. In other words, a set qualifies as a topological space if we can unambiguously identify the neighbors of its points.

Although we might not explicitly reference distance, most datasets inherently possess some form of a metric for detailed analysis. For instance, point clouds use the metric of the space they are embedded in, providing neighboring information. In graphs, adjacent nodes are naturally considered neighbors. Similarly, in images, neighboring pixels are deemed neighbors. This paper treats these examples as topological spaces with their natural topologies. However, specific datasets, such as RNA-sequencing data, lack an inherent metric, requiring users to define a metric to determine which data points are considered close and distant, depending on the context.

\paragraph*{Simplicial Complexes.} One special family of topological spaces used throughout our paper will be {\em simplicial complexes.} A simplicial complex is a mathematical structure used in topology and combinatorics to study the properties of shapes and spaces in a \textit{discrete manner}. It consists of vertices, edges, and higher-dimensional simplices such as triangles, tetrahedra, and their higher-dimensional counterparts, which are glued together in a specific way to form a coherent whole.

\begin{wrapfigure}{r}{3in}
\vspace{-.2in}
    \centering
    \includegraphics[width=\linewidth]{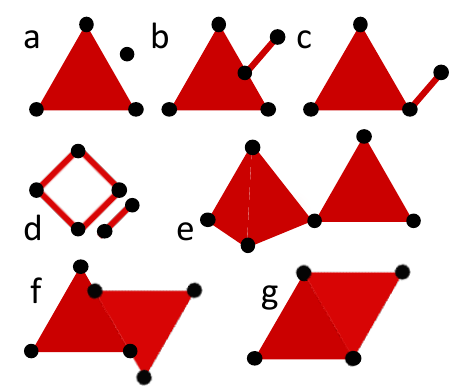}
        \caption{\textbf{Simplicial Complexes.} Among the complexes, only b and f fail to be simplicial complexes, as their simplices do not intersect at complete subsimplices. All others are valid simplicial complexes. \label{fig:scsample}}
    \vspace{-.2in}
\end{wrapfigure}

Each simplex is a generalization of the concept of a triangle to higher dimensions. By this, we mean that the properties and structure of a triangle (such as having vertices, edges, and faces) are extended into higher dimensions, even though the shapes look different as we go up in dimensions. In particular, a $k$-simplex is defined as the convex hull of affinely independent $k+1$ points in $\R^k$. For example, $2$-simplex is a triangle (with its inside filled), e.g., the convex hull of (smallest convex set containing) three points $\{(0,0), (1,0),(0,1)\}$ in $\R^2$. Similarly, $3$-simplex is a tetrahedron (with inside filled), e.g., the convex hull of $\{(0,0,0), (1,0,0),(0,1,0),(0,0,1)\}$ in $\R^3$. A union of simplices is called a \textit{simplicial complex} if any two simplices in the complex either do not intersect or intersect in a complete subsimplex (See \Cref{fig:scsample}).

\paragraph*{Dimension in Topology} One of the most confusing concepts for non-experts in topology is the notion of dimension. To study topological spaces more effectively, we focus on a specific family of spaces that meet certain regularity conditions, known as \textit{manifolds}. A $k$-dimensional manifold is a topological space that locally resembles a ball in $\R^k$, i.e., a small neighborhood of any point looks like (homeomorphic) a ball in $\R^k$. 
Importantly, we do not concern ourselves with the ambient (i.e., surrounding) space in which our manifold resides; we focus solely on its intrinsic properties, e.g., \textit{the material the manifold is made of}. For example, although a circle is often visualized in $\R^2$ as a two-dimensional object, from a topological perspective, it is actually one-dimensional. This is because, at any point on the circle, its local neighborhood resembles a line segment, i.e., it is made of $1$-dimensional material. Consider an ant walking on a circle; it has two options: moving forward or backward. While we might need two coordinates to describe the ant's position on the circle mathematically, from the ant's perspective, it only experiences a continuous path where it can move in either direction. To the ant, the circle feels like an endless line. Similarly, a sphere is a $2$-dimensional manifold (surface) even though it is often visualized in $\R^3$. This is because a small neighborhood of any point on the sphere resembles a piece of in $\R^2$. Notice that with this definition, the dimension of the ambient space becomes irrelevant; only the topological properties define the dimension of a manifold, i.e., a loop (circle) in $\R^2$  or $\R^3$ is called $1$-dimensional.  Similarly, a torus (a hollow donut) and a genus-2 surface (a surface with two holes) are examples of 2-dimensional manifolds. If this discussion leads you to wonder about manifolds with more than two dimensions, we have some discouraging news: they do exist, but visualizing them is not intuitive.

\paragraph*{Boundary.} In topology, understanding the boundary of a manifold is crucial to grasping its structure. A $k$-dimensional manifold $M$ is a space where each point has a neighborhood that resembles an open subset of $\R^k$. The boundary $\partial M$ of $M$ comprises points where these neighborhoods resemble an open subset of the half-space $\R^k_+ = \{ x \in \R^k \mid x_k \geq 0 \}$. This implies that the local structure is similar to part of $\R^k_+$ near each boundary point.

To illustrate, consider a sphere, a 2-dimensional surface without any boundaries. An ant walking on the sphere can indefinitely move in any direction without encountering an edge. In contrast, the unit disk $\D$ in $\mathbb{R}^2$ is a 2-dimensional surface with a boundary, specifically the unit circle $\C$.  An ant residing in $\D$ could not continue its journey once it reaches the boundary $\C$, or the "border" of $\D$. This is denoted as $\partial \D = \C$. Similarly, the unit ball $\B$ in $\R^3$   (the solid ball) is a 3-dimensional manifold with a boundary, and its boundary is the unit sphere $\s$, denoted by $\partial \B = \s$. In this context, we say that the sphere $\s$ bounds the ball $\B$.

If you're puzzled by the statement that \textit{ a sphere has no boundary, whereas the solid ball does}, consider it from the perspective of "its inhabitants." An ant living on the sphere's surface can move freely in any direction without ever encountering a limit (no border), which is why we say the sphere has no boundary. However, a mouse living inside the ball will eventually hit the boundary (border)—the sphere's surface—beyond which it cannot go. That limiting surface is what we call the solid ball's boundary. 

A key principle in topology is that a boundary's boundary is always empty. Mathematically, this is expressed as $\partial (\partial \X) = \emptyset$. Furthermore, the boundary of a $k$-dimensional manifold is generally a $(k-1)$-dimensional manifold with no boundary. For example, the boundary of a disk $\D$ (a 2-dimensional manifold with a boundary) is a circle $\C$ (a 1-dimensional manifold without a boundary). This observation will soon be important when discussing homology.

\subsubsection{Topological Equivalence} Topology primarily focuses on the global properties of shapes rather than their local characteristics. For example, topology is concerned with whether a space is connected or has holes without regard to the size of the object or the holes. 

Topological equivalence can be intuitively understood as follows: two shapes, $\X$ and $\Y$ are topologically equivalent if one can be continuously deformed into the other. Imagine $\X$ and $\Y$ are made of Play-Doh. If you can reshape $\X$ into $\Y$ without tearing, gluing, or collapsing any part of the shape, then they are considered to be continuously deformable into each other.  In this context, "collapsing" refers to reducing a part of the shape to a lower dimension, such as squishing a surface or line down to a point or flattening a 3-dimensional object into a 2-dimensional plane.

In mathematical terms, this concept corresponds to a \textit{homeomorphism} (the prefix "homeo-" comes from the Greek word "homoios," which means "similar" or "like"). In particular, such a deformation from $\X$ to $\Y$ represents a bijective map, which means it creates a one-to-one correspondence between elements of $\X$ and $\Y$, ensuring that each element in $\X$ is paired with a unique element in $\Y$ and vice versa. The map $\varphi:\X \to\Y$ keeps track of how each point $x \in\X$ becomes a point $y \in\Y$ after the deformation, i.e., $\varphi(\X) = y$. The condition of no tearing ensures that this map is continuous as nearby points in $\X$ must map to nearby points in $\Y$. Gluing is the opposite of tearing, so we also require that the inverse map $\varphi^{-1}$ is continuous. In summary, $\varphi:\X \to\Y$ is called a \textit{homeomorphism} if $\varphi$ is a continuous bijection with a continuous inverse.

We previously noted that size does not affect the topological structure. For example, let $\X = \R$, the set of all real numbers, and $\Y = (-1, 1)$, the open interval containing all real numbers between $-1$ and $1$, excluding the endpoints. These two spaces are topologically equivalent via the homomorphism $\varphi:\R\to(-1,1)$ with  $\varphi(x) = \frac{x}{1 + |x|}$. However, $\R$ has an infinite diameter, while $(-1,1)$ has a diameter of only 2. Hence, this is a good example of how the space size does not matter in topology. On the other hand, consider $\X = (0,1) \cup (1,2)$ and $\Y = (0,2)$. While both spaces are similar as sets, they are not topologically equivalent because $\X$ consists of two separate connected components, whereas $\Y$ is a single connected piece. A continuous deformation (homeomorphism) must preserve the number of components.

\paragraph*{Homotopy.} Another important concept in topology is {\em homotopy}, which can be seen as a flexible deformation of one space into another. Unlike homeomorphism, homotopy allows for collapsing but not tearing or gluing. We call two spaces homotopic to each other if such a deformation exists from one to another. For instance, a disk and a point are homotopic because the entire disk can be collapsed into a point by pushing inwards from all directions towards the center. Similarly, a punctured disk ($\mathbf{D}^2-\{(0,0)\}$) and a circle are homotopic, as a punctured disk can be continuously deformed towards its boundary, starting from the puncture point $(0,0)$. Homotopy provides a powerful tool for classifying topological spaces and understanding their fundamental properties, as most topological invariants are homotopy invariants, meaning they yield the same output for two homotopic spaces. For example, the connectivity and the count of holes or cavities do not change under homotopy.

\subsubsection{Topological Invariant} \label{sec:top_invariant} If  two spaces $\X$ and $\Y$ are topologically equivalent, showing that there is a homeomorphism $\varphi: \X \to \Y$ would be sufficient. However, if they are topologically different, one must demonstrate that no such map can exist, which is highly challenging. Mathematicians use invariants to show such inequivalence. An invariant refers to a property or feature of an object that remains unchanged under certain transformations. A \textit{topological invariant} is an invariant that is preserved under homeomorphism. A topological invariant can be a number (number of components, Euler characteristics, Betti numbers), a mathematical object (homology groups, fundamental group, cohomology ring), or a mathematical property (compactness, connectedness).

\begin{wrapfigure}{r}{3in}
\vspace{-.2in}
    \centering
    \subfloat[Sphere]{\includegraphics[width=0.3\linewidth]{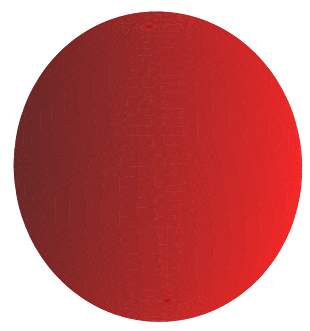}\label{fig:sphere}}
    \subfloat[Cube]{\includegraphics[width=0.3\linewidth]{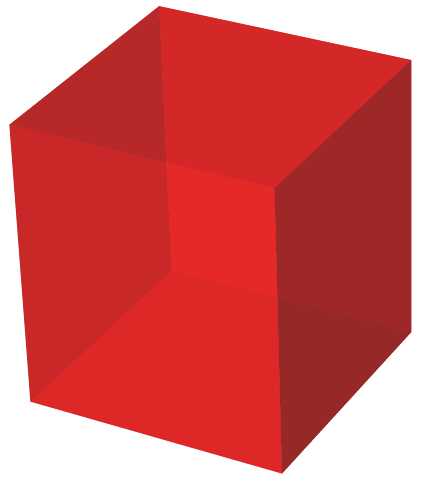}\label{fig:cube}}
    \subfloat[Torus]{\includegraphics[width=0.3\linewidth]{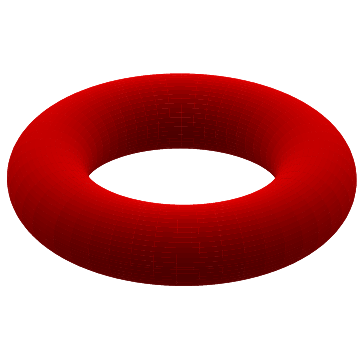}\label{fig:torus}}
    \caption{The sphere and the cube are topologically equivalent, whereas the torus is different from both.}
    \label{fig:sphere_and_torus}
    \vspace{-.1in}
\end{wrapfigure}

In the example above, $\X = (0,1) \cup (1,2)$ versus $\Y = (0,2)$, we used the number of components as a topological invariant to show they are not equivalent. However, more subtle examples require more sophisticated invariants. For instance, a cube and a sphere are topologically equivalent because one can deform a cube into a sphere without tearing or gluing. In contrast, comparing a sphere and a torus (the surface of a donut as shown in ~\ref{fig:torus}) is more complex. They are not topologically equivalent because the torus has "holes." By holes, we mean the loops on the surface, which cannot be shrunk down to a point \textit{without leaving the surface}. From this perspective, a sphere has no hole, however, on a torus, some loops (meridian and longitude) can't be shrunk down to a point in the torus, representing the "holes" in torus (see \Cref{fig:homology}). Despite this difference, proving that a sphere and a torus are not homeomorphic is challenging. Therefore, subtle topological invariants that detect these holes are crucial for distinguishing such spaces. 

\paragraph*{Euler Characteristics} One important and well-known example of such topological invariants is the {\em Euler characteristic}. If $\C$ is a simplicial complex, we define the Euler characteristic as the alternating sum of the count of $k$-simplices in $\C$, i.e., $\chi(\C) = \sum_k (-1)^k n_k$, where $n_k$ is the count of $k$-simplices in $\C$. It turns out the Euler characteristic is a topological invariant. i.e., if $\C$ and $\D$ are homeomorphic, then $\chi(\C) = \chi(\D)$. The Euler characteristic is, in fact, a homotopy invariant as well (see \cite{hatcher2002algebraic}). For example, a sphere $\s$ is topologically equivalent to the surface of a hollow tetrahedron $\F$, which has 4 vertices ($n_0$), 6 edges ($n_1$), and 4 triangles ($n_2$). Therefore, we have $\chi(\s) = \chi(\F) = n_0 - n_1 + n_2 = 4 - 6 + 4 = 2$. Similarly, if one computes the Euler characteristic of a torus $\T$ through a simplicial complex, we find that $\chi(\T) = 0$. Since $\chi(\s) \neq \chi(\T)$, it follows that a sphere is not homeomorphic to a torus.

\subsubsection{Geometry vs. Topology} Before proceeding, let's clarify a common confusion between the terms \textit{topology} and \textit{geometry}. Both are branches of mathematics, but they focus on different aspects of metric spaces.

Geometry focuses on the \textit{local study} of shapes, sizes, and properties of space, as well as how these spaces embed (fit into) higher-dimensional spaces.  For example, a 2-dimensional sphere $\s$ can be embedded into $\R^3$ (or $\R^{10}$) in various shapes and sizes.
Geometry focuses on precise measurements (e.g., angles, distances, curvature) and the relationships between points, lines, surfaces, and solids. It explores how shapes bend and twist, providing tools to understand complex surfaces and spaces through concepts such as curvature and geodesics (i.e., shortest paths).

In contrast, topology examines the \textit{global properties} of space that remain unchanged under continuous deformations. It focuses on qualitative aspects like connectedness and continuity rather than precise measurements. For instance, topology considers a cube and a sphere as topologically equivalent because these shapes can be transformed into one another through continuous deformation despite their geometric differences—one being flat and the other curved. For instance, no matter how we embed a 2-sphere into $\mathbb{R}^{10}$, its topological properties remain unchanged. However, the geometry varies significantly: a round sphere with a radius of 1 differs greatly from one with a radius of 100. Moreover, an irregular sphere would be geometrically distinct from both. 

In data science, geometry usually refers to the local shape characteristics of a dataset, such as distances, curvature, and angles, whereas topology pertains to global characteristics, such as connectedness and the number of holes/cavities. For example, dimension reduction methods like PCA~\cite{pcatutorial} and UMAP~\cite{mcinnes2018umap} are considered geometric methods as they heavily depend on the distances and how the dataset sits in the high dimensional space. In contrast, methods that count the number of components or holes in the space are called topological methods.

\subsection{Homology} \label{sec:homology}

To distinguish topological spaces, the most common method is to use topological invariants such as Euler characteristics~(\Cref{sec:top_invariant}), the fundamental group~\cite{armstrong2013basic}, or homology. Among these, homology is the most versatile and robust invariant that applies to a wide range of spaces such as surfaces (e.g., spheres, tori), simplicial complexes (e.g., triangulated shapes), and manifolds (e.g., higher-dimensional analogs of curves and surfaces).

There are various ways to compute homology (cellular~\cite{hatcher2002algebraic}, simplicial, Morse~\cite{matsumoto2002introduction}), where the outputs are the same, but the computation methods applied are different. To utilize computational tools more effectively,  it's more efficient to use discrete representations of topological spaces, like simplicial complexes. Simplicial homology is particularly suited for TDA because it deals with simplicial complexes, which are sets of vertices, edges, triangles, and their higher-dimensional counterparts. Considering these aspects, TDA mostly employs \textit{simplicial homology} \text{to capture the topological patterns in data}, 
although Morse or cellular homology are used in specific applications within TDA, such as cellular in image classification~\cite{kannan2019persistent, onus2022quantifying}.

 Here, we provide a brief overview (TLDR) of homology, followed by a formal yet accessible introduction. For a detailed, friendly introduction to simplicial homology, refer to~\cite[Chapter 8]{armstrong2013basic}, or for an in-depth study, see~\cite{hatcher2002algebraic,munkres2018elements}.

\subsubsection{TLDR} \label{sec:TLDR}
Homology is a fundamental invariant in topology that captures information about a space's structure by examining its holes/cavities of various dimensions. The focus on holes/cavities may surprise the reader, but holes are preferred because they are fundamental features that significantly influence the structure and properties of a space, as we outline below. To simplify the exposition, we use the concept of a \textit{$k$-hole} in a topological space $\X$. Although this term slightly abuses notation, it refers to a $k$-dimensional submanifold $\s$ in $\X$ that cannot be continuously deformed into a point within the space. In reality, this $k$-hole corresponds to a $(k+1)$-dimensional "cavity" $\Omega$ in $\X$. Since this cavity represents a "missing" region within the space, we describe it by using its boundary $\partial\Omega = \s$, which is non-contractible in $\X$. The $k^{th}$ homology group $\h_k(\X)$ captures these $k$-holes, or $k$-dimensional manifolds in $\X$, that do not bound any $(k+1)$-dimensional region in the space.

 \begin{itemize}
\item $k$-holes are topological invariants, meaning they remain unchanged under continuous deformations, such as stretching or bending, that do not involve tearing or gluing. And homology can detect them.
\item $0$-holes represent the connected components of a space. The dimension (or rank) of $\h_0(\X)$ corresponds to the number of these components in $\X$.
\item $1$-holes correspond to loops in the space that cannot be contracted to a point without leaving the space. The dimension of $\h_1(\X)$ represents the number of such loops ($1$-holes) in the space.
\item $2$-holes correspond to cavities within the space, which can be considered hollow regions enclosed by surfaces (e.g., the interior of a sphere or torus). The dimension of $\h_2(\X)$ represents the number of such cavities ($2$-holes) in the space.
 \end{itemize}

\begin{wrapfigure}{r}{2.8in}
\vspace{-.2in}
    \centering
    \includegraphics[width=\linewidth]{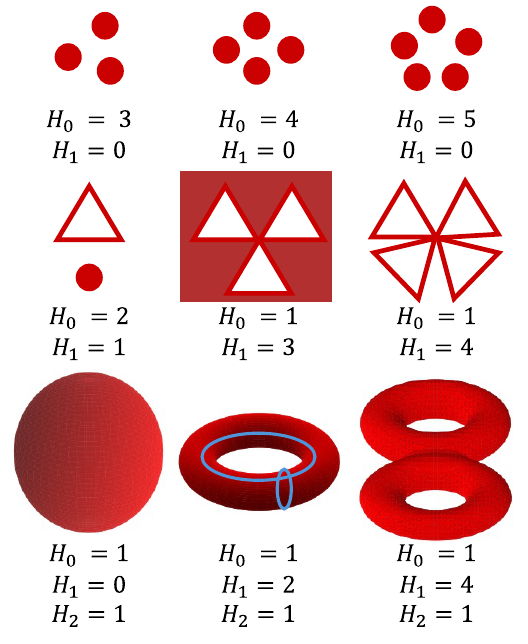}
        \caption{\textbf{Toy examples for homology.} We present the ranks of the homology groups of various topological spaces, i.e., for $\h_0(\X)=\BZ^k$ we write only $\h_0=k$ for simplicity, representing the count of the $k$-dimensional holes in $\X$.  \label{fig:homology}}    
\vspace{-.2in}
\end{wrapfigure}

To make the concept of homology groups more accessible, we will offer a simplified and visual explanation. 
We denote homology groups by $ \h_k(\X) $, where $ \X $ represents the space under consideration, and $ k $ indicates the dimension being analyzed. Independent $k$-holes are generators for the homology group $\h_k(\X)$. Therefore, the rank of $ \h_k(\X) $ indicates the number of $ k $-dimensional \textquote{holes} present in the space $ \X $. These numbers are also called {\em Betti numbers} $\{\beta_k(\X)\}$. e.g., if $\h_2(\X)=\BZ^3$, then we say rank($\h_2(\X))=3$ , or $\beta_2(\X)=3$, meaning $\X$ has three $2$-holes.  Below, we will give some examples for dimensions $0,1$ and $2$ (See \Cref{fig:homology}).

\noindent $\bullet$ \textbf{$\h_0(\X)$:} $0$-dimensional homology represents the connected components in $\X$. If $\X$ has three components, then we say $\h_0(\X)=\BZ^3$ (or $\BZ_2^3$ if used $\BZ_2$-coefficients), and rank$(\h_0(\X))=\beta_0(\X)=3$.

    \smallskip

\noindent $\bullet$ \textbf{$\h_1(\X)$:} $1$-dimensional homology computes the non-contractible loops (holes) in $\X$.
    For example, a sphere $\s$ has no nontrivial loops, as any loop in the sphere bounds a disk \underline{in the sphere}, i.e., $\beta_1(\s)=0$.
    A torus $\T$ (hollow donut), on the other hand, has meridian and longitude circles (see Figure~\ref{fig:homology}), both being non-contractible loops, making the total $\beta_1(\T)=2$. On the other hand, a solid donut $\D$ would have only the longitude circle as noncontractible, $\beta_1(\D)=1$.

    \smallskip
    
\noindent $\bullet$ \textbf{$\h_2(\X)$:} $2$-dimensional homology group captures two-dimensional cavities (voids) in the space $\X$. To count these cavities, we utilize surfaces in the space which does not bound a $3$-dimensional domain in the space. For example, in the unit ball $\B$ in $\R^3$ (solid ball), any closed surface bounds a 3-dimensional domain within $\B$, and there are no cavities within it. Hence, we say rank$(\h_2(\B))=\beta_2(\B)=0$. Similarly, if one removes two disjoint smaller balls from $\B$, say $\B'$, we would have two different cavities, which can be represented by different spheres in $\X$, enclosing these balls. Then, we say $\beta_2(\B')=2$. In~\Cref{fig:homology}, the sphere, the torus, or the genus-$2$ surface have one $2$-dimesional cavities represented by themselves.

We conclude this section with an interesting fact. While we defined the Euler characteristic for simplicial complexes as the alternating sum of the number of $k$-simplices, there is an alternative way using Betti numbers. In particular, if $\Y$ is a topological space, the Euler characteristic is given by $\chi(\Y)=\sum_k(-1)^k\beta_k(\Y)$.

\subsubsection{Computation of Homology} \label{sec:homology2}
We now turn to a formal introduction. Homology is a mathematical operation whose inputs are topological spaces and outputs are groups. In particular, for a given space $\X$, $\h_k(\X)$ represents the $k^{th}$ homology group of $\X$, summarizing the $k$-dimensional non-collapsible submanifolds in $\X$, each representing different $(k+1)$-dimensional "cavity" of $\X$. We will call these "$k$-holes" by abusing notation. 

The concept of homology groups stems from the idea that a $(k+1)$-dimensional hole or cavity in $\X$ is detected by the presence of its $k$-dimensional boundary, which cannot be continuously contracted within $\X$. While this method might seem indirect—using boundaries to infer the existence of cavities—the difficulty arises from the need to identify what is \textit{missing} in $\X$. Here, a fundamental principle of topology comes into play: \textit{the boundary of a boundary is always empty}. If $\Omega$ is a $(k+1)$-dimensional domain in $\X$ with boundary $\s = \partial \Omega$, then the boundary of $\s$ is empty. i.e., $\partial(\partial \Omega) = \partial \s=\emptyset$ Hence, to identify cavities, we first find all $k$-dimensional submanifolds with no boundary in $\X$ ($k$-cycles). Then, by eliminating those that bound a domain in $\X$ ($k$-boundaries), we are left with the true cavities. This process forms the core idea behind homology computation.

To keep the exposition focused, we will describe only \textit{simplicial homology with $\mathbb{Z}_2$ coefficients}, the most common version used in TDA. For other homology settings, refer to~\cite{hatcher2002algebraic}. In particular, simplicial homology involves representing a given space $\mathcal{X}$ as a simplicial complex (a collection of simplices) and performing computations on these simplices. This approach allows us to discretize the problem by focusing on the $k$-dimensional "building blocks" of the topological space, such as vertices for $0$-dimension and edges for $1$-dimension. By considering $k$-submanifolds  (or $k$-subcomplexes) as unions of $k$-simplices, we can identify the special ones that correspond to true cavities in $\mathcal{X}$ by using computational tools. To formalize this concept, we now introduce the relevant mathematical notions.

\paragraph*{i. Representation of $k$-simplices} In a simplicial complex $\X$, we describe $k$-simplices by listing their vertices. For example, a $1$-simplex (an edge) $e$ with endpoints $v_2$ and $v_4$ is denoted as $e = [v_2, v_4]$. Similarly, a $2$-simplex (a triangle) $\tau$ with vertices $v_1$, $v_5$, and $v_7$ is written as $\tau = [v_1, v_5, v_7]$. Since we are using $\mathbb{Z}_2$ coefficients, the order of the vertices does not matter. However, in other versions, such as with $\mathbb{Z}$-coefficients, the order of the vertices would be significant. In general, a $k$-simplex $\Delta$ is represented by its $k+1$ vertices, i.e., $\Delta = [v_{i_0}, v_{i_1}, \dots, v_{i_k}]$. We will call a union of $k$-simplices \textit{a $k$-subcomplex of $\X$}. 

\paragraph*{ii. $k$-chains $\C_k(\X)$} To describe all $k$-dimensional subcomplexes (which correspond to all $k$-submanifolds, with or without boundary) within a simplicial complex $\X$, we define a group $\C_k(\X)$, known as the group of {\em $k$-chains}. Recall that any union of $k$-simplices forms a $k$-subcomplex. For instance, if the simplicial complex $\X$ consists of three edges $\{e_1, e_2, e_3\}$, then all possible $1$-subcomplexes are $\{e_1, e_2, e_3, e_1 \cup e_2, e_1 \cup e_3, e_2 \cup e_3, e_1 \cup e_2 \cup e_3\}$. We represent this collection using group elements, where each $1$-simplex acts as a generator. In particular, the union $e_1 \cup e_3$ is represented by $\sigma_1 = (1,0,1)$, while $e_1 \cup e_2$ is represented by $\sigma_2 = (1,1,0)$. Hence, we obtain the group $\C_1(\X) = \BZ_2^3$, where $0$ element corresponds to the empty set. The group operation is addition, and since $1+1=0$ in $\BZ_2$, we have $\sigma_1 + \sigma_2 = (1,0,1) + (1,1,0) = (0,1,1)$, corresponding to the union $e_2 \cup e_3$. Similarly, if $\X$ contains $m$ $k$-simplices $\{\Delta_1, \Delta_2, \dots, \Delta_m\}$, then $\C_k(\X) = \BZ_2^m$, where a group element like $(1,0,1,\dots,1)$ represents the $k$-subcomplex $\Delta_1 \cup \Delta_3 \cup \Delta_m$ within $\X$.

\paragraph*{iii. Boundary operator $\partial_k$} To identify $k$-holes in a simplicial complex, we must first identify $k$-subcomplexes that have no boundary. This is accomplished by defining a boundary operator $\partial_k: \mathcal{C}_k(\X) \to \mathcal{C}_{k-1}(\X)$, which maps $k$-chains to their $(k-1)$-dimensional boundaries. In particular, $\partial_k$ maps each $k$-simplex to its boundary. For example, if we have a $1$-simplex (edge) $e = [v_0, v_1]$, then $\partial_1[e] = v_0 + v_1 \in \C_0(\X)$, representing the boundary of the edge as the union of its end vertices $\{v_0\} \cup \{v_1\}$. Similarly, for a $2$-simplex (triangle) $\tau = [v_0, v_1, v_2]$, the boundary is given by $\partial_2\tau = [v_0, v_1] + [v_1, v_2] + [v_2, v_0] \in \C_1(\X)$, representing the boundary of the triangle as the union $[v_0, v_1] \cup [v_1, v_2] \cup [v_2, v_0]$. 
To define $\partial_k$ for general $k$-chains, we sum the boundaries of each $k$-simplex within the chain. For instance, if $\sigma = \Delta_1 + \Delta_3 + \Delta_7$ is a $k$-chain, then $\partial_k\sigma = \partial_k\Delta_1 + \partial_k\Delta_3 + \partial_k\Delta_7$. 
A $k$-chain $\sigma$ is said to have \textit{no boundary} if $\partial_k\sigma = 0$. In other words, a $k$-subcomplex with no boundary in $\X$ must map to $0$ under the boundary operator.

\begin{wrapfigure}{r}{0.2\linewidth}
\vspace{-.15in}
    \centering
    \includegraphics[width=\linewidth]{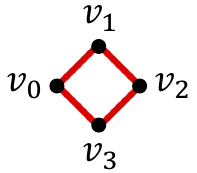}
        \caption{\textbf{Toy example for homology.}   \label{fig:toy1}}    
\vspace{-.15in}
\end{wrapfigure}

The boundary operator $\partial_k: \mathcal{C}_k(\X) \to \mathcal{C}_{k-1}(\X)$ is a linear operator and can be represented as a matrix. For example, if $\mathcal{C}_k(\X) = \mathbb{Z}_2^n$ and $\mathcal{C}_{k-1}(\X) = \mathbb{Z}_2^m$, then $\partial_k$ can be written as an $m \times n$ matrix $\mathbf{A}$, with each column $\mathbf{A}_i$ corresponds to $\partial_k \Delta_i$, where $\Delta_i$ is a $k$-simplex in $\X$. In \Cref{fig:matrix1}, we give an explicit example of a matrix representation of a boundary operator the simplicial complex in \Cref{fig:toy1}.

\paragraph*{iv. $k$-cycles $\Z_k(\X)$} We define a special subgroup $\Z_k(\X)$, named \textit{$k$-cycles}, within $\C_k(\X)$ for $k$-subcomplexes that have no boundary. As previously mentioned, a $k$-subcomplex with no boundary in $\X$ must map to zero under the boundary operator. Hence, we define the subgroup $\Z_k(\X) = \ker \, \partial_k$, which includes all $k$-chains that map to zero.  Recall that $1$-chains with no boundary correspond to loops, while 2-chains with no boundary correspond to closed surfaces, like a sphere or torus.

For example, consider a square-shaped simplicial complex $\X$ (\Cref{fig:toy1}) with four vertices $\{v_0,v_1,v_2,v_3\}$ and four edges $\{e_1,e_2,e_3,e_4\}$  where $e_1=[v_0,v_1],e_2=[v_1,v_2],e_3=[v_2,v_3]$ and $e_4=[v_3,v_0]$. In this case, $\C_1(\X)=\BZ_2^4$ and $\C_0(\X)=\BZ_2^4$. Suppose $\sigma_1=e_1+e_2$; then $\partial_1\sigma_1= (v_0+v_1)+(v_1+v_2)=v_0+v_2$, meaning that $\sigma_1$ represents a 1-subcomplex with a boundary. However, if $\sigma_2=e_1+e_2+e_3+e_4$, then $\partial_1\sigma_2= (v_0+v_1)+(v_1+v_2)+(v_2+v_3)+(v_3+v_0)=2(v_0+v_1+v_2+v_3)=0$, indicating that $\sigma_2$ represents a 1-subcomplex with no boundary. Notice that $\sigma_2$ corresponds to a complete loop in $\X$. See \Cref{ex:toy1} for more details.

\begin{wrapfigure}{r}{1.3in}
\vspace{-.15in}
\footnotesize
$\partial_1 = 
\begin{array}{c}
\partial e_1 \ \partial e_2 \ \partial e_3 \ \partial e_4  \\
\begin{pmatrix}
1 & 0 & 0 & 1 \\ 
1 & 1 & 0 & 0 \\ 
0 & 1 & 1 & 0 \\ 
0 & 0 & 1 & 1 
\end{pmatrix}
\end{array}
\textcolor{red}{
\begin{array}{c}
\ \\
v_0 \\
v_1 \\
v_2 \\
v_3
\end{array}}$
\caption{\footnotesize $\partial_1:\C_1(\X)\to\C_0(\X)$ is represented as a $4\times 4$ binary matrix. Columns represent the edges in $\C_1(\X)$, and rows correspond to the vertices in $\C_0(\X)$. For example,  $\partial e_3= v_2+v_3$ can be read from the third column \label{fig:matrix1}}
\vspace{-.2in}
\end{wrapfigure}

We can also describe the boundary map $\partial_1: \C_1(\X) \to \C_0(\X)$ as a $4 \times 4$ matrix, as shown in \Cref{fig:matrix1}. It is clear that the only element in $\C_1(\X)$ that maps to $(0,0,0,0) \in \C_0(\X)$ is $(1,1,1,1)$, which corresponds to the loop $\sigma_2$. This means $\X$ has only one $1$-cycle, which is $\sigma_2$. Then, $\Z_k(\X)$ is the subgroup of $\C_k(\X)$ generated by the single element $(1,1,1,1)$. If you are unfamiliar with group theory, you can think of $\C_1(\X)$ as a four-dimensional vector space, where $\Z_1(\X)$ is a one-dimensional subspace generated by the vector $(1,1,1,1)$.

\paragraph*{v. $k$-boundaries $\B_k(\X)$} While we identified all $k$-subcomplexes with no boundary, none correspond to true cavities. We must eliminate the ones that bounds a domain in $\X$. To find them, we use again the boundary operator. As $\C_{k+1}(\X)$ represents all $(k+1)$-subcomplexes in $\X$, then the image of $\partial_{k+1}$, i.e., $\B_k(\X) = \partial_{k+1}\mathcal{C}_{k+1}(\X) \subset \mathcal{C}_k(\X)$, represents the ones which bounds a $(k+1)$-domain in $\X$. We call an element in $\B_k(\X)$ a \textit{$k$-boundary}. Recall that $\partial_k(\partial_{k+1} \sigma)=0$ from earlier discussion. This means for any $k$-boundary $\varphi=\partial_{k+1} \sigma$ in $\B_k(\X)$, $\partial_k\varphi=0$. Therefore, we have $\B_k(\X)\subset\Z_k(\X)$. For example, if $\X$ is a simplicial complex formed by only one $2$-simplex $\tau$ with vertices $\{v_0,v_1,v_2\}$, then $\B_1(\X)$ would be a subgroup in $\C_1(\X)$, generated by $\partial_2 \tau=\sigma=(1,1,1)$ while $\Z_1(\X)$ would be the same group, i.e., $\Z_1(\X)=\B_1(\X)$. This means there is only one loop $\sigma$ in $\X$ but it does not represent a hole in $\X$ as it is filled by $\tau$, i.e., $\sigma=\partial \tau$.

\paragraph*{ vi. Homology group $\h_k(\X)$} Now we are ready to define homology. Notice that with the boundary operator, we obtained the following sequence of groups and maps. We can consider these as vector spaces and linear maps. For each $k$, $k$-chains $\C_k(\X)$ represent all $k$-subcomplexes in $\X$, $k$-cycles $\Z_k(\X)\subset \C_k(\X)$ represent $k$-subcomplexes with no boundary, and finally, $\B_k(\X)\subset \Z_k(\X)$ represent the $k$-subcomplexes which bounds a $(k+1)$-domain in $\X$.

\[
\begin{tikzcd}
\cdots \arrow[r, "\partial_{k+2}"] & \C_{k+1}(\X) \arrow[r, "\partial_{k+1}"] & \C_k(\X) \arrow[r, "\partial_k"] & \C_{k-1}(\X) \arrow[r, "\partial_{k-1}"] & \cdots \arrow[r, "\partial_1"] & \C_0(\X)\arrow[r] & 0
\end{tikzcd}
\]

Now, to identify $k$-holes (true cavities), we consider all $k$-cycles in $\X$ ($\Z_k(\X)$), which potentially represent a cavity of $\X$, and, among them, eliminate all $k$-boundaries in $\X$ ($\B_k(\X)$), as they represent the "fake" cavities. Hence, we define $k^{th}$ homology group as the quotient group $$\h_k(\X) = \Z_k(\X)/\B_k(\X) = \text{ker}(\partial_k)/\text{image}(\partial_{k+1})$$
In terms of the sequence above, this quotient $\h_k(\X) = \Z_k(\X)/\B_k(\X)$  effectively counts the $k$-dimensional cycles that correspond to actual holes or cavities in $\X$, not those that are merely boundaries of higher-dimensional regions. From a computational perspective, with this formulation, we only need to compute the kernels and images of a sequence of linear maps $\{\partial_k\}$ (binary matrices) to compute homology.

\begin{wrapfigure}{r}{0.2\linewidth}
\vspace{-.2in}
    \centering
    \includegraphics[width=\linewidth]{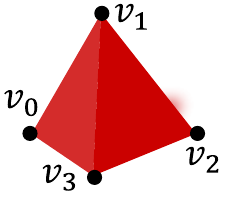}
        \caption{\textbf{Toy example \ref{ex:toy2} for homology.}   \label{fig:toy2}}  
        \vspace{-.15in}
\end{wrapfigure}
To clarify these notions, we give two toy examples for explicit computation of homology.

\begin{example} \label{ex:toy1} Consider the example in \Cref{fig:toy1} where $\X$ is a square-shaped simplicial complex with four vertices $\{v_0, v_1, v_2, v_3\}$ and four edges. For $k \geq 2$, $\C_k(\X) = \{0\}$ since there are no $k$-simplices in $\X$. Both $\C_1(\X)$ and $\C_0(\X)$ are isomorphic to $\BZ_2^4$, as discussed earlier ($k$-cycles above). The group $\Z_1(\X)$ has only one generator, the sum of all edges. Since $\C_2(\X) = \{0\}$, we have $\B_1(\X) = \{0\}$ because $\B_1(\X) = \partial_2 \C_2(\X)$. Therefore, $\h_1(\X) = \Z_1(\X)/\B_1(\X) = \BZ_2/\{0\} = \BZ_2$, meaning that $\h_1(\X)$ has rank 1, corresponding to the entire square loop. The boundary map $\partial_1$ is given in~\Cref{fig:matrix1}.

For $\h_0(\X)$, we need to determine $\Z_0(\X)$ and $\B_0(\X)$. Since $\partial_0$ sends everything to 0, $\Z_0(\X) = \C_0(\X) = \BZ_2^4$. The group $\B_0(\X) = \partial_1 \C_1(\X)$ is generated by the boundaries of the edges: $(v_0 + v_1), (v_1 + v_2), (v_2 + v_3), (v_3 + v_0)$. However, these boundaries are not linearly independent because $v_3 + v_0$ is the sum of the other three boundaries. Thus, $\B_0(\X) \simeq \BZ_2^3$. Consequently, $\h_0(\X) = \Z_0(\X)/\B_0(\X) = \BZ_2^4/\BZ_2^3 = \BZ_2$. Recall that the rank of $\h_0(\X)$ represents the number of connected components and the fact that $\operatorname{rank}(\h_0(\X)) = 1$ confirms that there is only one connected component.  
\end{example}

\begin{wrapfigure}{r}{3.2in}
\vspace{-.1in}
\footnotesize
$\partial_2 = 
\begin{array}{c}
\partial \tau_1\ \partial \tau_2 \ \partial \tau_3 \ \partial \tau_4 \\ 
\begin{pmatrix}
1 & 0 & 1 & 0 \\ 
1 & 1 & 0 & 0 \\ 
0 & 1 & 0 & 1 \\ 
0 & 0 & 1 & 1\\ 
1 & 0 & 0 & 1 \\ 
0 & 1 & 1 & 0   
\end{pmatrix} 
\end{array}
\textcolor{red}{
\begin{array}{c}
\ \\
e_1 \\
e_2 \\
e_3 \\
e_4 \\
e_5 \\
e_6
\end{array}}$ \ 
$\partial_1 = 
\begin{array}{c}
\partial e_1 \ \partial e_2 \ \partial e_3 \ \partial e_4 \ \partial e_5 \ \partial e_6 \\
\begin{pmatrix}
1 & 0 & 0 & 1 & 1& 0 \\ 
1 & 1 & 0 & 0 & 0 & 1 \\ 
0 & 1 & 1 & 0 & 1 & 0\\ 
0 & 0 & 1 & 1 & 0 & 1\\ 
\end{pmatrix} 
\end{array}
\textcolor{red}{
\begin{array}{c}
\ \\
v_0 \\
v_1 \\
v_2 \\
v_3
\end{array}}$
\caption{\footnotesize {\bf Boundary maps for Ex. \ref{ex:toy2}.} The boundary map $\partial_2:\C_2(\X)\to\C_1(\X)$ is represented as a $6\times 4$ binary matrix (left). The top of each column corresponds to the $2$-simplex in $\C_2(\X)$ whose image is represented by that column, while each row's corresponding edge is given next to the column. For example, the boundary of $2$-simplex $\tau_2$, $\partial\tau_2= e_2+e_3+e_6$ by reading the second column in $\partial_2$ matrix. The boundary map $\partial_1:\C_1(\X)\to\C_0(\X)$ is similar (right). \label{fig:matrix2}}
\vspace{-.2in}
\end{wrapfigure}
Second example is in one dimension higher.

\begin{example} \label{ex:toy2} Consider the hollow tetrahedron $\X$ of \Cref{fig:toy2}, composed of four triangular faces: $\tau_1 = [v_0, v_1, v_2]$, $\tau_2 = [v_1, v_2, v_3]$, $\tau_3 = [v_0, v_1, v_3]$, and $\tau_4 = [v_0, v_2, v_3]$. The complex $\X$ contains four 2-simplices (triangles), six 1-simplices (edges), and four 0-simplices (vertices). Let $e_1=[v_0,v_1],e_2=[v_1,v_2], e_3=[v_2,v_3],e_4=[v_3,v_0], e_5=[v_0,v_2]$ and $e_6=[v_1,v_3]$. Therefore, we have $\C_2(\X) = \BZ_2^4$, $\C_1(\X) = \BZ_2^6$, and $\C_0(\X) = \BZ_2^4$.

A straightforward computation shows that the kernel of $\partial_2$, denoted $\Z_2(\X)$, has a single generator: the sum of all the triangles, $\tau_1 + \tau_2 + \tau_3 + \tau_4$. Since there is no 3-simplex in $\X$, we find that $\B_2(\X) = \{0\}$, leading to $\h_2(\X) = \BZ_2$. This result aligns with the fact that rank$(\h_2(\X)) = 1$, indicating the presence of one cavity in $\X$, consistent with $\X$ being a hollow tetrahedron. We 
Furthermore, calculating $\Z_1(\X) = \BZ_2^3$ and $\B_1(\X) = \BZ_2$ shows that these cancel out, yielding $\h_1(\X) = \{0\}$. In other words, any loop in $\X$ is filled, so there is no $1$-hole in $\X$.
Similarly, following the same reasoning as in the example above, we obtain $\h_0(\X) = \BZ_2$ as expected.  
\end{example}

%% file: sections/3-PH.tex
In this section, we introduce Persistent Homology (PH), a foundational technique that played a pivotal role in the emergence of TDA, as developed by Carlsson, Edelsbrunner, Zomorodian, and others in the early 2000s~\cite{edelsbrunner2002topological, carlsson2004persistence, zomorodian2004computing}. PH captures the underlying shape patterns within complex data sets by studying the evolution of topological features across multiple scales.

\begin{wrapfigure}{r}{3in}
\vspace{-.1in}
    \centering
    \includegraphics[width=\linewidth]{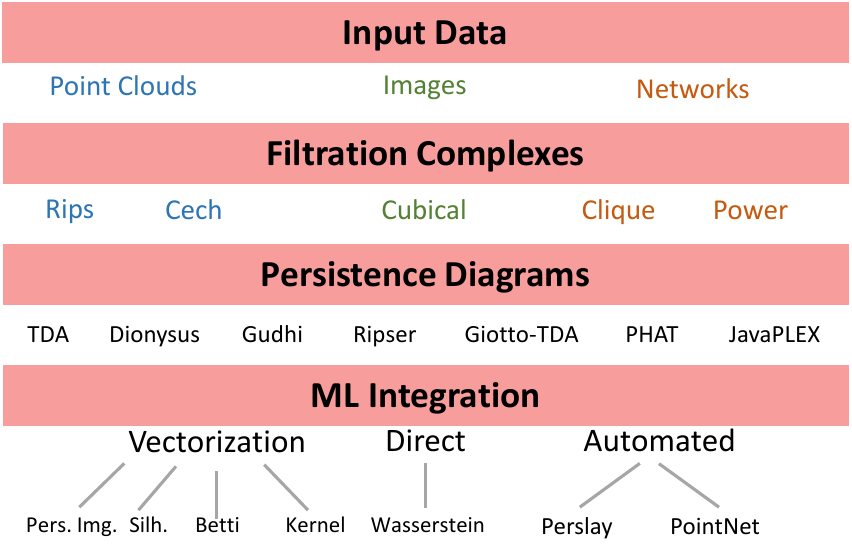}
    \caption{\footnotesize \textbf{PH Pipeline.}  From data acquisition and filtration complex construction to generating persistence diagrams using software libraries. The final step highlights methods for integrating persistence diagrams into downstream ML tasks. \label{fig:pipe}}
    \vspace{-.15in}
\end{wrapfigure}
PH first constructs a nested sequence of simplicial complexes, known as \textit{filtration}, and tracks the birth and death of features, such as connected components, loops, and voids, in this sequence. The resulting multi-scale representation highlights significant features while filtering out noise, making PH a valuable tool in various fields, including medical imaging~\cite{singh2023topological}, biomedicine~\cite{skaf2022topological}, time series analysis~\cite{zeng2021topological}, material science~\cite{obayashi2022persistent}, geography~\cite{corcoran2023topological}, shape analysis~\cite{turkes2022effectiveness} and finance~\cite{ismail2022early}.

In the following, we explain PH in an accessible way, focusing on its key aspects relevant to ML applications. The main idea behind PH is to capture the hidden shape patterns in the data by using algebraic topology tools. PH achieves this by keeping track of the evolution of the topological features ($k$-holes, components, loops, and cavities) created in the data while looking at it in different resolutions.

In simple terms, PH can be summarized as a three-step procedure as follows.

\begin{enumerate}
    \item \textbf{Filtration}: Generate a  nested sequence of simplicial complexes derived from the data.
    \item \textbf{Persistence Diagrams}: Record the evolution of topological features across this sequence.
    \item \textbf{ML Integration}: Transform the persistence diagrams into vectors for efficient use in ML models.
\end{enumerate}

We provide details of these steps in the following sections. Although the second and third steps are conceptually similar across most settings, the first step, constructing filtrations, varies significantly depending on the data type, i.e., point clouds, images, and networks. See \Cref{fig:pipe} for a visual summary of PH pipeline.

\subsection{Constructing Filtrations} \label{sec:filtrations}

In \Cref{sec:background}, we introduced simplicial complexes which enable us to study the topological spaces via computational tools. To study the given data in different resolutions, PH generates a nested sequence of simplicial complexes $\K_1\subset \K_2\subset \dots \subset \K_n$ induced from the data. Such a sequence is called a {\em filtration}. This step can be considered the most crucial for the effectiveness of PH in ML applications. The primary reason is that the filtration process involves examining the data at different resolutions by adjusting a "scale parameter". The choice of this "scale parameter" can greatly influence the performance of the method.

For each data type, there are well-established methods to construct filtrations that have proven highly effective in their respective contexts. These methods vary significantly depending on the data type. While point cloud~\cite{chazal2015subsampling} and image~\cite{yadav2023histopathological} settings use relatively common approaches in their settings, the construction of filtrations in graph settings is notably more versatile~\cite{aktas2019persistence}.

\begin{wrapfigure}{r}{3in}
\vspace{-.15in}
    \centering
    \subfloat[$\epsilon=0.4$]{%
        \includegraphics[width=0.33\linewidth]{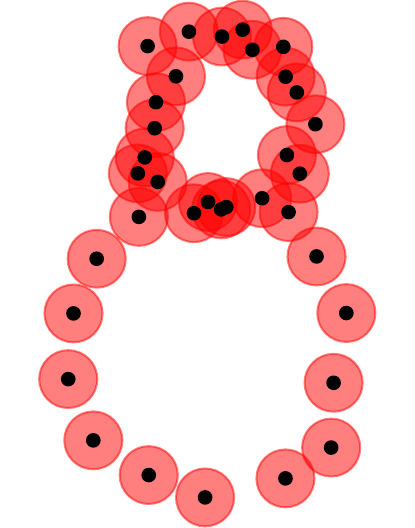}
        \label{fig:p8_04a}
    }
    \subfloat[$\epsilon=0.7$]{%
        \includegraphics[width=0.33\linewidth]{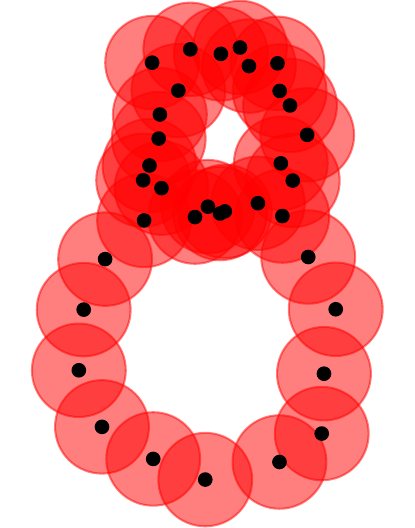}
        \label{fig:p8_07a}
    }
    \subfloat[$\epsilon=2$]{%
        \includegraphics[width=0.33\linewidth]{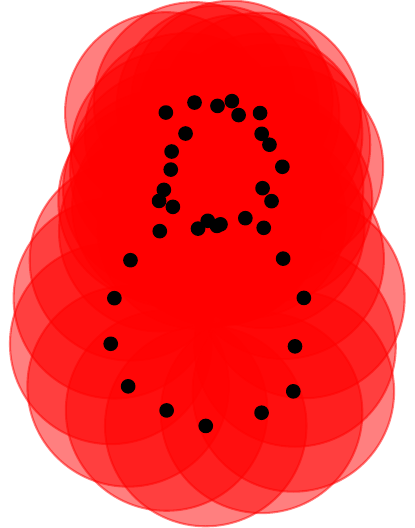}
        \label{fig:p8_2a}
    }
    \caption{ For 8-shaped point cloud $\X$, $\{\N_\e(\X)\}$ filtration steps for thresholds $\e=0.4,\e=0.7$ an $\e=2$. \label{fig:point_filtration} }    
    \vspace{-.15in}
\end{wrapfigure}

\subsubsection{Filtrations for Point Clouds} \label{sec:point_cloud}
As the process is generally similar for various metric spaces, for simplicity, we will describe it for a point cloud $\X$ in $\mathbb{R}^N$ using the Euclidean metric. Let $\X = \{x_1, x_2, \dots, x_m\}$ be a point cloud in $\mathbb{R}^N$. We will define a nested sequence of simplicial complexes $\K_1 \subset \K_2 \subset \dots \subset \K_n$ induced by $\X$.  The central idea here is to build a series of simplicial complexes that progressively capture the topological picture of the point cloud $\X$ in different resolutions as we move from $\K_1$ to $\K_n$. The "nested" part signifies that each simplicial complex is contained within the next one, like a series of Russian Matryoshka dolls.

Before moving on to simplicial complexes, we will define a simpler nested sequence for $\X$. Let $\B_r(x)=\{y\in\R^N\mid d(x,y)\leq r\}$ be the closed $r$-ball around $x$. Then, let $r$-neighborhood of $\X$, $\N_r(\X) = \bigcup_{i=1}^m \B_r(x_i)$ be the union of $r$-balls around the points in $\X$. By declaring $r$ as our scale parameter, we first fix a monotone sequence of threshold values $0 = r_1 < r_2 < \dots < r_n$, where $r_n = \max_{i,j}\{d(x_i, x_j)\}$, the diameter of $\X$. These values intuitively represent the resolution at which we observe the point cloud $\X$. In particular, a smaller value of $r$ indicates a closer, more detailed examination of $\X$, while a larger value of $r$ means observing the point cloud with a broader view from a greater distance, making it difficult to distinguish between points that are close together in $\X$. This naturally gives a nested sequence of topological spaces $\N_{r_1}(\X) \subset \N_{r_2}(\X) \subset \dots \subset \N_{r_n}(\X)$ (See~\Cref{fig:point_filtration}). While these neighborhoods  $\{\N_r(\X)\}$ give a natural sequence for the data, to effectively leverage computational tools, we need to induce a \textit{sequence by simplicial complexes}. There are two common ways to achieve this while preserving the underlying topological information.

\begin{wrapfigure}{r}{3in}
\vspace{-.25in}
    \centering
    \subfloat[Cech Complex]{%
        \includegraphics[width=0.33\linewidth]{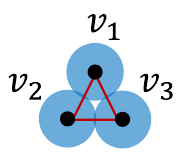}
        \label{fig:cech}    }
       \subfloat[Rips Complex]{%
        \includegraphics[width=0.33\linewidth]{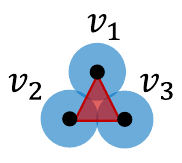}
        \label{fig:rips}    }
        \subfloat[Cech Complex]{%
        \includegraphics[width=0.33\linewidth]{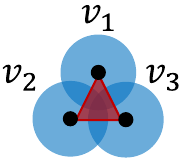}
        \label{fig:cech2}    }
    \caption{\footnotesize \textbf{Comparison of Rips and \v{C}ech Complexes.} In panels~\ref{fig:cech} and \ref{fig:rips}, the \v{C}ech and Rips complexes differ: the \v{C}ech complex does not form a 2-simplex among $v_1$, $v_2$, and $v_3$ due to insufficient ball overlap, while the Rips complex does. A larger radius, as in panel~\ref{fig:cech2}, is needed for the \v{C}ech complex to include this simplex.}
    \label{fig:ripsandcech}
    \vspace{-.25in}
\end{wrapfigure}

\paragraph*{Method 1 - Rips complexes:} For $\X\subset \R^N$, and for $r>0$, the \textit{Rips complex} (aka Vietoris-Rips complex) is the abstract simplicial complex $\VR_r(\X)$ where a $k$-simplex $\sigma=[x_{i_0},x_{i_1},\dots,x_{i_k}]\in \VR_r(\X)$
if and only if $d(x_{i_m},x_{i_n})< r$ for any $0\leq m,n\leq k$. In other words, for $r>0$, if for $k+1$-points are pairwise $r$-close to each other, they form a $k$-simplex in $\VR_r(\X)$.

\paragraph*{Method 2 - \v{C}ech complexes:} Similarly,  for $\X\subset \R^N$, and for $r>0$, the \v{C}ech complex is the abstract simplicial complex $\check\C_r(\X)$ where a $k$-simplex $\sigma=[x_{i_0},x_{i_1},\dots,x_{i_k}]\in \check\C_r(\X)$ if and only if $\bigcap_{j=0}^k B_r(x_{i_j})\neq \emptyset$. Here, the condition to build $k$-simplex is different as we ask for a nontrivial intersection of the $r$-balls of the $k+1$-points.

\begin{algorithm}[b]
\caption{PH Machinery for Point Clouds with Rips Complexes \label{alg:PH_point}}
\begin{algorithmic}[1]
\State \textbf{Input:} Point cloud $P$, Distance metric $d$, Threshold set $\{\epsilon_i\}_{i=0}^n$
\State \textbf{Output:} Topological vector $\vec{\beta}(P, d, \{\epsilon_i\})$
\State $V \gets \{v \in P\}$ \Comment{Vertex set doesn't change with filtration}
\For{$i = 0$ to $n$} \Comment{Filtration index}
    \State \textbf{Vertex set remains the same for all $i$}
    \State $E_i \gets \{(u, v) \mid u, v \in P \text{ and } d(u, v) \leq \epsilon_i\}$ \Comment{Edges with distance $\leq \epsilon_i$}
    \State $\VR_i \gets  (V, E_i)$ \Comment{Rips complex at $\epsilon_i$}
\EndFor
\For{$k = 0$ to $1$} \Comment{Topological dimensions}
    \State Compute persistence diagram $\mathrm{PD}_k(\{\VR_i\}_{i=0}^n)$
    \State Obtain vectorization $\vec{\beta}_k$ from $\mathrm{PD}_k(\{\VR_i\}_{i=0}^n)$
\EndFor
\State \textbf{Return} Concatenation of $\vec{\beta}_0$ and $\vec{\beta}_1$, denoted as $\vec{\beta}(P, d, \{\epsilon_i\}) = \vec{\beta}_0 \mathbin\Vert \vec{\beta}_1$
\end{algorithmic}
\end{algorithm}

The main relationship between our original filtration $\{\N_{r_i}(\X)\}$ and the simplicial complex filtrations arises from the \textit{Nerve Lemma}. This lemma states that the \v{C}ech complex $\check{\mathcal{C}}_r(\X)$ is homotopy equivalent to $\N_r(\X)$ for any $r \geq 0$~\cite{dey2022computational}. Since homotopy equivalent spaces have the same homology, we have $\h_k(\check{\mathcal{C}}_r(\X)) \simeq \h_k(\N_r(\X))$ for any $k \geq 0$. Therefore, from a PH perspective, the filtrations $\{\N_{r_i}(\X)\}_1^n$ and $\{\check{\mathcal{C}}_{r_i}(\X)\}_1^n$ are essentially the same. Furthermore, for $\X \subset \R^N$ with the Euclidean metric, Rips and \v{C}ech complexes are closely related and produce similar topological information as  $\check{\mathcal{C}}_r(\X) \subset \VR_{2r}(\X) \subset \check{\mathcal{C}}_{\sqrt{2}r}(\X)$~\cite{edelsbrunner2022computational}.

While both complexes provide similar filtrations, Rips complexes are more commonly used in practice. This preference is due to the fact that Rips complexes only require the pairwise distances $\{d(x_i, x_j)\}$ among points, which can be easily obtained at the beginning of the process. Hence, constructing Rips complexes is straightforward once these distances are known. In contrast, constructing \v{C}ech complexes requires checking whether collections of $r$-balls have nontrivial intersections. \Cref{fig:ripsandcech} depicts their differences in a toy scenario.

In particular, Rips complexes are the most common filtration type used for point clouds because of their computational practicality, and because of the Nerve Lemma, they capture very similar information produced by the simple neighborhood filtration $\{N_r(\X)\}$ described earlier.

Although we discuss the construction for point clouds in $\R^N$, the same process can be applied to any metric space (any space with a distance function). Furthermore, in some cases, the point cloud $\X$ is given in an abstract setting, where only pairwise distances $\{d(x_i, x_j)\}$ among the points are provided. The Rips complex filtration can be effectively applied to such point clouds. In~\Cref{sec:shape}, we detail the real-life application of this approach for shape recognition.

\paragraph*{Other complex types.} One of the primary challenges in applying PH to point clouds is the computational cost. To mitigate this, \textit{witness complexes} offer a valuable alternative for efficiently analyzing the topological features of point clouds, especially in high-dimensional spaces. Unlike the Rips complexes, which can be computationally expensive, witness complexes reduce complexity by utilizing a \textit{representative subset} of the data points, known as \textit{witnesses}, to construct the simplicial complex~\cite{otter2017roadmap}. This method strikes a balance between computational efficiency and topological accuracy, making it particularly well-suited for large datasets where constructing full complexes would be impractical. By concentrating on representative data points (witnesses), witness complexes facilitate a more manageable and scalable computation of persistent homology, enabling the detection of the underlying shape and features of the point cloud across various scales. In addition to Rips and Čech complexes, other types of complexes, such as \textit{alpha complexes} and \textit{Delaunay complexes}, are particularly useful in lower-dimensional spaces. However, to maintain the focus of this paper, we refer the reader to~\cite{otter2017roadmap} for an in-depth discussion of these complexes.

\begin{figure}[t] 
\centering
    	\includegraphics[width=\linewidth]{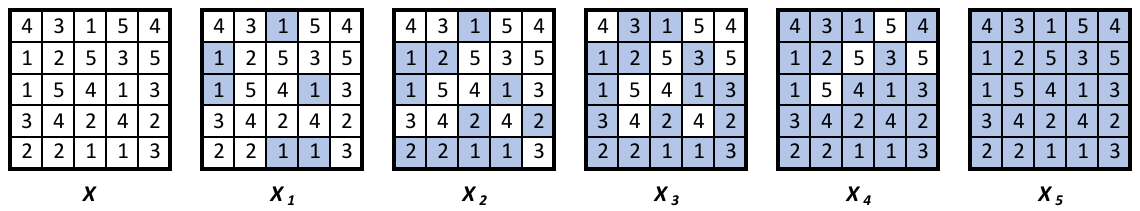}  
 {\caption{For the $5\times 5$ image $\X$ with the given pixel values, \textbf{the sublevel filtration} is the sequence of binary images $\X_1\subset \X_2\subset \X_3\subset \X_4\subset \X_5$. \label{fig:image-filtration}}}
   \vspace{-.2in}
    \end{figure}

\subsubsection{Filtrations for Images} \label{sec:image}

For images, constructing filtrations differs significantly due to the unique structure of image data. To keep the explanation simple, we will focus on 2D images, though the concepts can be extended to 3D and other types of images. Filtration for images typically involves nested sequences of binary images, known as \textit{cubical complexes}~\cite{kaji2020cubical}. Starting with a given color (or grayscale) image $\X$ of dimensions $r \times s$, we first select a specific color channel (e.g., red, blue, green, or grayscale). The color values $\gamma_{ij} \in [0, 255]$ of individual pixels $\Delta_{ij} \subset \X$ are used, where $\Delta_{ij}$ represents a \underline{closed box} (square including its boundary) in the $i^{th}$ row and $j^{th}$ column of the image $\X$.

\begin{figure}[b]
\vspace{-.2in}
    \centering
    \includegraphics[width=\linewidth]{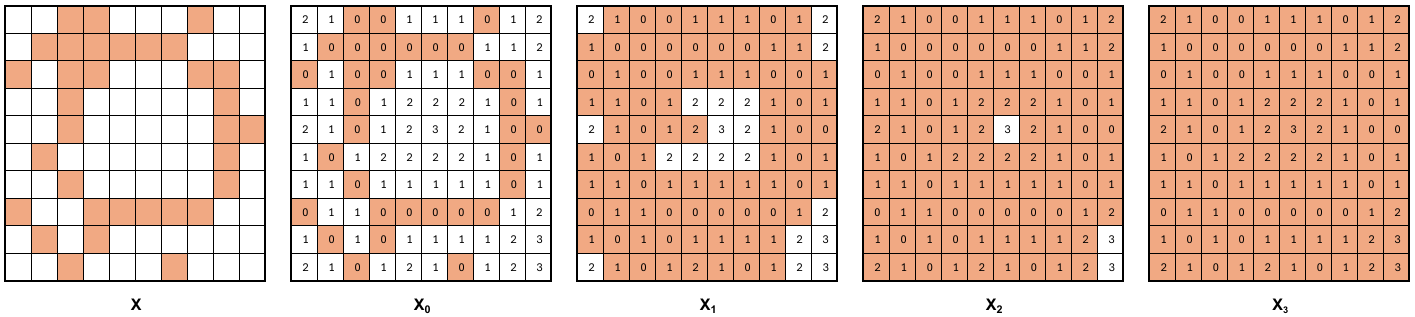}
    \caption{\footnotesize {\bf Erosion Filtration.} For a given binary image $\X$, we define the erosion function, which assigns a value of 0 to black pixels and the distance to the nearest black pixel for white pixels (shown in $\X_0$). From this, we generate a filtration of binary images, $\X_0 \subset \X_1 \subset \dots \subset \X_3$, where each $\X_i$ is obtained by activating pixels that reach a specific threshold value. \label{fig:erosion}}    
\end{figure}
Next, we choose the number  of thresholds "$n$" to span the color interval $[0, 255]$, i.e., $0 = t_1 < t_2 < \dots < t_n = 255$. This determines the length of our filtration sequence, with a typical range being between 50 and 100. Using these thresholds, we define a nested sequence of binary images (cubical complexes) $\X_1 \subset \X_2 \subset \dots \subset \X_n$ such that $\X_m = \{\Delta_{ij} \subset \X \mid \gamma_{ij} \leq t_m\}$ (see \Cref{fig:image-filtration}).

In particular, this involves starting with a blank $r \times s$ image and progressively activating (coloring black) pixels as their grayscale values reach each specified threshold $t_m$. This process is known as {\em sublevel filtration}, applied to $\X$ relative to the chosen color channel (in this case, grayscale). Alternatively, pixels can be activated in descending order, referred to as {\em superlevel filtration}. In this context, let $\Y_m = \{\Delta_{ij} \subset \X \mid \gamma_{ij} \geq s_m\}$, where $255 = s_1 > s_2 > \dots > s_n = 0$, and $\Y_1 \subset \Y_2 \subset \dots \subset \Y_n$ is called superlevel filtration. The persistence homology process involving such cubical complexes has a special name, called {\em cubical persistence}. In \Cref{sec:histo}, we detail a successful application of this method in cancer detection from histopathological images.

While these sublevel and superlevel filtrations are common choices for color or grayscale images, there are other filtration types used for binary images, e.g., erosion, dilation, and signed distance~\cite{garin2019topological}. These filtrations are specific to binary images and effective in capturing the size and other properties of topological features. In~\Cref{fig:erosion}, we give an example of erosion filtration.

\begin{algorithm}[t]
\caption{PH Machinery for Images with Cubical Persistence \label{alg:PH_image}}
\label{alg:cubical}
\begin{algorithmic}[1]
\State \textbf{Input:} Image $\mathcal{X}$, Color channel $F: \mathcal{X} \to \mathbb{R}$, Threshold set $\{\epsilon_i\}_{i=0}^n$
\State \textbf{Output:} Topological vector $\vec{\beta}(\mathcal{X}, F, \{\epsilon_i\})$
\For{$i = 0$ to $n$} \Comment{Filtration index}
    \State $\mathcal{X}_i \gets \{\Delta_{ij} \in \mathcal{X} \mid F(\Delta_{ij}) \leq \epsilon_i\}$ \Comment{Binary Images}
\EndFor
\For{$k = 0$ to $1$} \Comment{Topological dimensions}
    \State Compute persistence diagram $\mathrm{PD}_k(\{\mathcal{X}_i\}_{i=0}^n)$
    \State Obtain vectorization $\vec{\beta}_k$ from $\mathrm{PD}_k(\{\mathcal{X}_i\}_{i=0}^n)$
\EndFor
\State \textbf{Return} Concatenation of $\vec{\beta}_0$ and $\vec{\beta}_1$, denoted as $\vec{\beta}(\mathcal{X}, F, \{\epsilon_i\}) = \vec{\beta}_0 \mathbin\Vert \vec{\beta}_1$
\end{algorithmic}
\end{algorithm}

\subsubsection{Filtrations for Graphs} \label{sec:graph}

Graphs are a widely used application domain for PH because they generate highly effective features that are not easily obtained through other methods. While filtration methods for point clouds and images are relatively standard, graph filtrations offer a variety of choices. The way you construct these filtrations can significantly impact the performance of ML models. Unlike other data formats, there are various methods to construct filtrations from graphs, where the details can be found in~\cite{aktas2019persistence}. We categorize these methods into two groups:

\paragraph*{i. Filtrations through node/edge functions.} This type of filtration is computationally efficient in most cases and is commonly used in applications. The main concept involves defining a filtration function on nodes or edges and using the order dictated by these functions to create a sequence of subgraphs and corresponding simplicial complexes. 

\begin{wrapfigure}{r}{3in}
\vspace{-.15in}
\centering
\includegraphics[width=\linewidth]{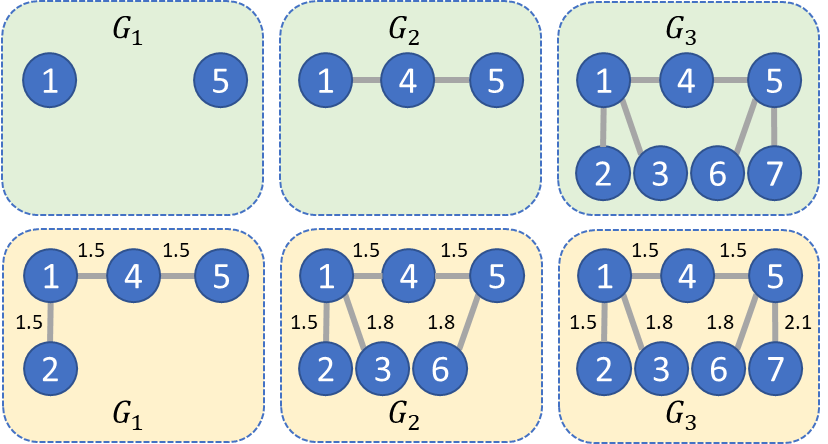}
\caption{\footnotesize \textbf{Graph Filtration.} For $\mathcal{G}=\mathcal{G}_3$ in both examples, the top figure illustrates a \textit{superlevel filtration using the node degree function} with thresholds $3>2>1$, where nodes of degree 3 are activated first, followed by those of lower degrees.  Similarly, the bottom figure illustrates a \textit{sublevel filtration based on edge weights} with thresholds $1.5< 1.8< 2.1$.  \label{fig:graph-filtration}}
\vspace{-.15in}
\end{wrapfigure}

Given a graph $\G=(\V,\E)$, let $f:\V\to\R$ be a \textit{node filtration function}. Common examples of such functions include degree, centrality, betweenness, closeness, and heat kernel signatures (HKS). Additionally, functions may be derived from the domain of the graph, such as atomic number for molecular graphs or account balance for transaction networks. These functions establish a hierarchy among the nodes, ordering them from less important to more important within the graph within the context defined by the filtration function.

To define the resolution of our filtration, we choose a set of thresholds that cover the range of $f$, denoted as $\mathcal{I}=\{\epsilon_i\}_1^n$, where $\epsilon_1=\min_{v \in \V} f(v) < \epsilon_2 < \ldots < \epsilon_n = \max_{v \in \V} f(v)$. For each threshold $\epsilon_i \in \mathcal{I}$, let $\V_i = \{v \in \V \mid f(v) \leq \epsilon_i\}$. Define $\G_i$ as the subgraph of $\G$ induced by $\V_i$, i.e., $\G_i = (\V_i, \E_i)$ where $\E_i = \{e_{ij} \in \E \mid v_i, v_j \in \V_i\}$. In other words, at each threshold, we activate the nodes whose values reach that threshold, along with the edges in the graph between these activated nodes.

This process results in a nested sequence of subgraphs $\G_1 \subset \G_2 \subset \ldots \subset \G_n = \G$ (See~\Cref{fig:graph-filtration}). For each $\G_i$, we define an abstract simplicial complex $\widehat{\G}_i$ for $1 \leq i \leq n$, creating a \textit{filtration} of simplicial complexes $\widehat{\G}_1 \subseteq \widehat{\G}_2 \subseteq \ldots \subseteq \widehat{\G}_n$. Typically, \textit{clique complexes} are used, where the complex $\widehat{\G}$ is obtained by assigning a $k$-simplex to each $(k+1)$-complete subgraph ($(k+1)$-clique) in $\G$~\cite{aktas2019persistence}. The term \textit{clique} refers to a group (clique) of $k+1$ vertices forming a $k$-simplex.  This is called {\em sublevel filtration} for the filtration function $f$. Similarly, one can reverse this process (using a decreasing order for activation of the nodes) to obtain a \textit{superlevel filtration}, where $\V_i = \{v \in \V \mid f(v) \geq \epsilon_i'\}$ where $\{\e_i'\}$ is a monotone decreasing sequence. 

While we have explained the process for node filtration functions, edge filtration functions are also common. This approach is frequently employed in weighted graphs or graphs with key edge functions for downstream tasks. For a graph $\G$, let $g: \E \to \R$ be an \textit{edge filtration function}. In the case of weighted graphs, the weight function $w: \E \to \R$ with $w(e_{ij}) = \omega_{ij}$ serves as an example of such a function. Other common examples include Forman Ricci and Ollivier Ricci curvatures, which assign the (weighted or unweighted) edge a curvature value by interpreting the geometry of the edge's neighborhood~\cite{fesser2024augmentations}. Furthermore, edge filtration functions can be derived from the graph's domain, such as transaction amounts in blockchain networks~\cite{abay2019chainnet} or density in traffic networks~\cite{carmody2021topological}. These functions establish a hierarchy among the edges, similar to node filtration functions.

Again, by choosing the threshold set $\mathcal{I}=\{\e_i\}_{i=1}^n$ with $\e_1=\min_{e_{ij}\in\E}g(e_{ij})<\e_2<\ldots<\e_n=\max_{e_{jk}\in\E}g(e_{jk})$, we define the filtration as follows. For $\e_i\in \mathcal{I}$, let $\mathcal{E}_i=\{e_{jk}\in\mathcal{E}\mid g(e_{jk})\leq \e_i\}$. 
Let $\mathcal{G}_i$ be a subgraph of $\mathcal{G}$ induced by $\mathcal{E}_i$, i.e., $\mathcal{G}_i$ is the smallest subgraph of $\mathcal{G}$ containing the edge set $\mathcal{E}_i$ (Fig. \ref{fig:graph-filtration}). In other words, for each threshold, we activate the edges whose value reaches that threshold, along with the nodes attached to them. Again, this induces a nested sequence of subgraphs $\mathcal{G}_1\subset  \dots\subset\mathcal{G}_n=\mathcal{G}$. Then, one can follow the same steps by using clique complexes as before. 

In \Cref{fig:graph-filtration}, we give a toy example for superlevel (node) and sublevel (weight) filtrations. Here, we list the most common node and edge filtration functions used in applications:

\begin{itemize}
    \item {\em Node Filtration Functions:} Degree, betweenness, centrality, eccentricity, heat kernel signature (HKS), Fiedler values (spectral), node functions from the domain of the data.
    \item {\em Edge Filtration Functions:} edge weights (weighted graphs), Ollivier Ricci, Forman Ricci curvature, 
\end{itemize}

\begin{algorithm}[t]
\caption{PH machinery for graphs (sublevel filtration) \label{alg:PH_graph}}

\begin{algorithmic}
\State \textbf{Input:} Graph $\mathcal{G} = (\mathcal{V}, \mathcal{E})$, Filtration function $f: \mathcal{V} \to \mathbb{R}$, Threshold set $\mathcal{I} = \{\epsilon_i\}_{i=0}^n$
\State \textbf{Output:} Topological vector $\vec{\beta}(\mathcal{G}, f, \mathcal{I})$
\For{$i = 0$ to $n$} \Comment{Filtration index}
    \State $\mathcal{V}_i \gets \{v \in \mathcal{V} \mid f(v) \leq \epsilon_i\}$ \Comment{Vertex set changes with i}
    \State $\mathcal{G}_i \gets$ Induced subgraph of $\mathcal{G}$ with vertex set $\mathcal{V}_i$
    \State $\wh{\mathcal{G}}_i \gets$ Clique complex of $\mathcal{G}_i$
\EndFor
\For{$k = 0$ to $1$} \Comment{Topological dimensions}
    \State Compute persistence diagram $\mathrm{PD}_k(\{\wh{\mathcal{G}}_i\}_{i=0}^n)$
    \State Obtain vectorization $\vec{\beta}_k$ from $\mathrm{PD}_k(\{\wh{\mathcal{G}}_i\}_{i=0}^n)$
\EndFor
\State \textbf{Return} Concatenation of $\vec{\beta}_0$ and $\vec{\beta}_1$, denoted as $\vec{\beta}(\mathcal{G}, f, \mathcal{I}) = \vec{\beta}_0 \mathbin\Vert \vec{\beta}_1$
\end{algorithmic}
\end{algorithm}

\paragraph*{ii. Graph distance-based filtrations.} Next to the sublevel filtrations above, a completely distinct filtration method for graphs uses the distances between nodes. Essentially, it treats the graph as a point cloud, where pairwise distances between nodes are defined by graph distance, and applies Rips filtration (\Cref{sec:point_cloud}) to this point cloud. While this method is more computationally intensive than filtrations via node or edge functions, it can be highly effective for small graphs as it captures distance information and finer topological details.

To apply this method, we need to define graph distances between nodes. In an unweighted graph $\G = (\V, \E)$, a common approach is to set the distance between adjacent nodes to 1 ($\|e_{ij}\| = 1$) and define the distance between $v_i$ and $v_k$ as the length of the shortest path $\tau_{ik}$ in $\G$ with endpoints $v_i$ and $v_k$, i.e. $d(v_i, v_k) = \min \|\tau_{ik}\|$. Thus, each pairwise distance is an integer representing the number of hops needed to travel from $v_i$ to $v_j$ in $\G$. The largest such distance is the diameter of $\G$, denoted as $n$.

After obtaining the pairwise distances $\{d(v_i, v_j)\}$ between the nodes, we treat the graph as a point cloud $\V = \{v_1, v_2, \dots, v_m\}$ with these distances. Since all distances are integers, we use integer thresholds $\{r_i = i\}$ for $0 \leq i \leq n = \text{diam}(\G)$. Now, we are ready to define our filtration. For the filtration index, we use superscripts instead of subscripts, where the reason will become clear later.

To make the exposition simpler, let $\wh{\G}^k$ be the Rips complex corresponding to $r=k$, and let $\G^k$ be the graph corresponding to the $1$-skeleton of $\wh{\G}^k$. Here $1$-skeleton means the graph itself with its nodes and edges, without considering any higher-dimensional elements like faces or solids that might be part of a more complex simplicial structure. For the distance parameter $r = 0$, we have $\wh{\G}^0 = \V$ since there are no edges in the Rips complex. It is easy to see that at the distance threshold of 1, $\mathcal{G}^1 = \mathcal{G}$ because all edges are automatically included in the complex when $r = 1$. Recall that $\wh{\G}^1$ is a Rips complex, hence any complete $j$-subgraph in $\G^1$ generates a $(j-1)$-simplex ($j$-clique) in the Rips complex $\wh{\G}^1$. In particular, $\wh{\G}^k$ is nothing but the clique (or flag) complex of $\G^k$. Furthermore, $\G^k$ is the graph with node set $\V$, and edge set $\E^k = \{e_{ij} \mid d(v_i, v_j) \leq k\}$, i.e., $\G^k=(\V,\E^k)$. Therefore, in the filtration $\wh{\G}^0 \subset \wh{\G}^1 \subset \dots \subset \wh{\G}^n$, all simplicial complexes have the same node set, but at each step, we add new edges and cliques. Finally, for $\wh{\G}^n$, the 1-skeleton is a complete graph with $m$ nodes, and hence the corresponding simplicial complex $\wh{\G}^n$ would be an $(m-1)$-simplex. 

\begin{wrapfigure}{r}{3in}
\centering
\includegraphics[width=\linewidth]{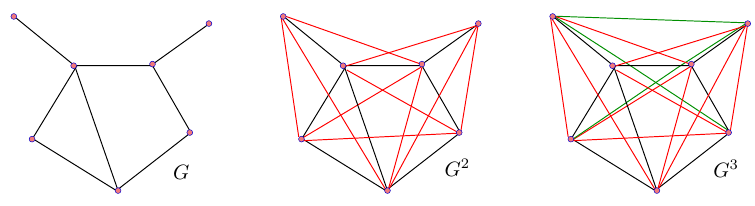}
\caption{\footnotesize {\bf Graph Powers.} A graph $G=G^1$ and its graph powers.
	Red edges are added in $G^2$, and green ones are added in $G^3$.
	Note $\mathcal{G}^3$ is the complete graph on $7$ vertices since $D=\text{diam}(G)=3$. Hence, $\mathcal{G}^3$ would be a complete graph $\mathcal{K}_7$, and all higher powers are the same, i.e.\ $\mathcal{G}^n=\mathcal{G}^3$ for $n\geq 3$. \label{fig:power-filtration}} 
\end{wrapfigure}
The graphs $\{\G^k\}$ are called {\em graph powers} (See~\Cref{fig:power-filtration}), and this filtration is called the {\em power filtration} (or Rips filtration), hence the superscripts. Observe that the power filtration calculates the distances between every pair of vertices in the graph. Therefore, even vertices that are not direct neighbors or appear distant in the graph can still form a simplex in the later stages of the filtration. Note that you don't need to complete the entire filtration. In most cases, the critical insights information lies in the first few steps (e.g., up to $\e=5$ or $10$) for the power filtration. Given the high computational cost, it's both practical and reasonable to stop early.

For a weighted graph $\G = (\V, \E, \W)$, edge {lengths} can be defined using weights $\{\omega_{ij}\}$, where $\|e_{ij}\| = f(\omega_{ij})$ instead of the uniform $\|e_{ij}\| = 1$ used in unweighted graphs. The function $f: \W \to \R^+$ assigns distances based on weights, where a smaller $f(\omega_{ij})$ indicates that nodes $v_i$ and $v_j$ are closely related, and a larger distance indicates they are less related. This approach is particularly useful in financial networks where large edge weights (i.e., amounts) indicate stronger financial connections or dependencies. In~\Cref{sec:crypto}, we elaborate on this method in the context of a real-world application: Crypto-token anomaly forecasting.

Once edge lengths are defined, pairwise distances between any node pair can be calculated as the length of the shortest path as before, i.e., $d(v_i, v_j) = \min \|\epsilon_{ij}\|$, where $\epsilon_{ij}$ is any  such path in the graph with endpoints $v_i$ and $v_j$. After obtaining pairwise distances, the process is the same as for unweighted graphs.

\begin{wrapfigure}{r}{2in}
\vspace{-.1in}
    \centering
    \includegraphics[width=\linewidth]{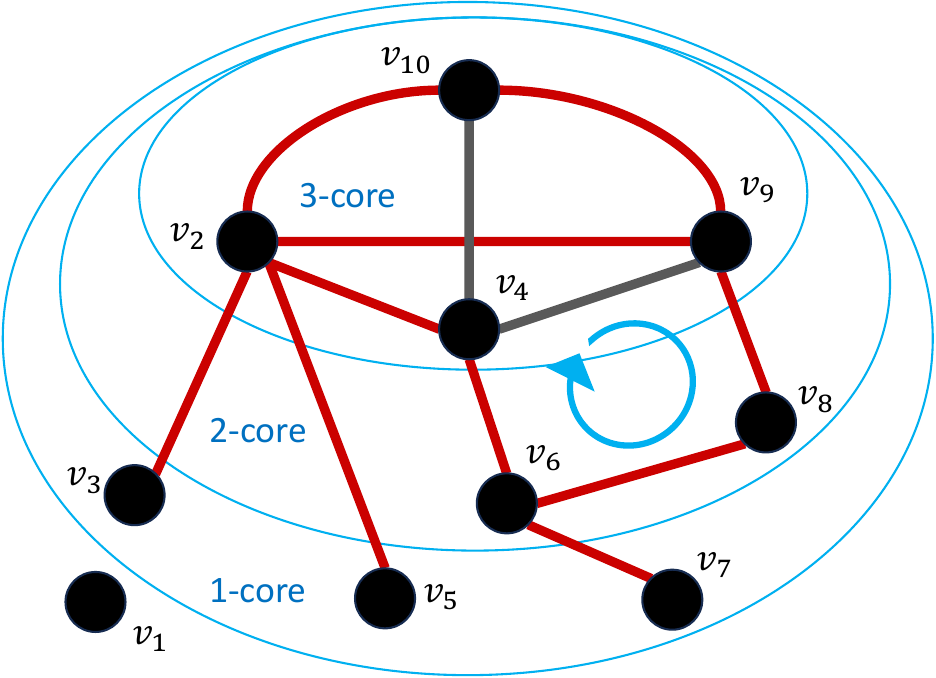}
    \caption{{\bf Coral Reduction.} With CoralTDA, $2$-core of $\G$ has the same persistence diagrams $\PD_k(\G)$ for $k\geq 1$.}
    \label{fig:coraltrick}
    \vspace{-.1in} 
\end{wrapfigure}

\paragraph*{Graph Reduction and PH} Although PH generates highly effective topological embeddings for graph representation learning, scalability remains a challenge. To address this issue, Akcora et al.~\cite{akcora2022reduction} introduced two key methods to significantly reduce the computational costs of the PH process. The first method, \textit{CoralTDA}, leverages the observation that nodes with low graph-core do not contribute to persistence diagrams in higher dimensions. Specifically, they demonstrate that the $(k+1)$-core of a graph, which is a subgraph where each vertex has at least $k$ neighbors, is sufficient to compute the $\PD_k(\G)$ of the original graph. This improvement can be implemented as few as three lines of Python code (see the repository).  The second algorithm, \textit{PrunIT}, introduces an efficient method to reduce the size of graphs without altering their persistence diagrams. In this approach, a vertex \( u \) is said to dominate another vertex \( v \) if the $1$-neighborhood of $u$ contains $1$-neighborhood of $v$, i.e., $\N(u)\supset \N(v)$. The authors show that removing (pruning) a dominated vertex from the graph does not affect the persistence diagrams at any level as long as the dominated vertex enters the filtration after the dominating vertex.

\paragraph*{Which filtration to use for graphs?} We first note that filtrations using node/edge functions and filtrations based on graph distances are entirely different methods, producing distinct outputs. Distance-based filtrations are computationally demanding, but for datasets with small graphs (such as molecular graph datasets), power filtrations can be highly effective. Conversely, for larger graphs, the computational costs necessitate the use of filtration functions. In these cases, the choice of filtration function could be critical for performance~\cite{cai2020understanding}. Typically, relevant node or edge functions specific to the domain of the data are the best choices. If such functions are unavailable, heat kernel signatures (HKS) for node functions and Ollivier Ricci for edge functions are excellent alternatives. Choosing a filtration function can be seen as an outdated method, as letting ML algorithms select or construct the best filtration function for optimal performance is more reasonable. However, in many settings, model interpretability or greater control over the process is needed. In these instances, selecting a relevant filtration function is more suitable for the model.

If the goal is to achieve optimal performance with PH machinery, learnable filtration functions are a promising alternative. There are significant works, along with available code, that can be adapted for various tasks in graph representation learning~\cite{hofer2020graph,zhang2022gefl}.

\subsubsection{Choosing Thresholds} \label{sec:threshold} One of the key steps in effectively applying PH to ML problems is selecting the thresholds $\{\e_i\}_{i=1}^m$ for constructing the filtration. The first decision involves determining the number of thresholds, i.e., $m$. This number can be viewed as \textit{the resolution of the filtration}: a larger $m$ implies higher resolution, while a smaller $m$ indicates lower resolution. Then, a natural question arises: is there any disadvantage to choosing a large $m$? The answer is "Yes". Firstly, the computational cost increases with $m$. Secondly, the key information captured in the data may become diluted in higher dimensions, depending on the vectorization step. 

For graph filtrations, depending on the filtration function, selecting $m$ between 10 and 20 generally yields good results. Once the number of thresholds is set, they can be chosen equally spaced if the filtration function $f: \V \to \R$ (or $g: \E \to \R)$ is appropriate. Another effective method is to examine the distribution of the value set $\{f(v_i)\}$ for all vertices in the graphs across the dataset (or a random subsample), and then select the thresholds as the corresponding quantiles in this distribution. Similarly, for distance-based filtrations, analyzing the distribution of pairwise distances between vertices in a random subsample of the dataset can provide valuable insights for selecting appropriate thresholds.

For sublevel filtrations in the image setting, it’s important to choose thresholds that cover the color range $[0,255]$. Typically, using 50-100 thresholds yields good results. However, if the task only concerns a specific color interval $[a,b]$, it’s advisable to concentrate most or all thresholds within that range. These thresholds don't need to be evenly spaced. In the case of binary image filtrations (e.g., erosion, height, signed distance), thresholds are usually evenly distributed up to the maximum value.

\begin{wrapfigure}{r}{2.8in}
\vspace{-.25in}
    \centering
    \subfloat[$\epsilon=0.4$]{%
        \includegraphics[width=0.4\linewidth]{figures/P8_0.4.pdf}
        \label{fig:p8_04}
    }
        \subfloat[PB up to $\epsilon=0.4$]{%
        \includegraphics[width=0.6\linewidth]{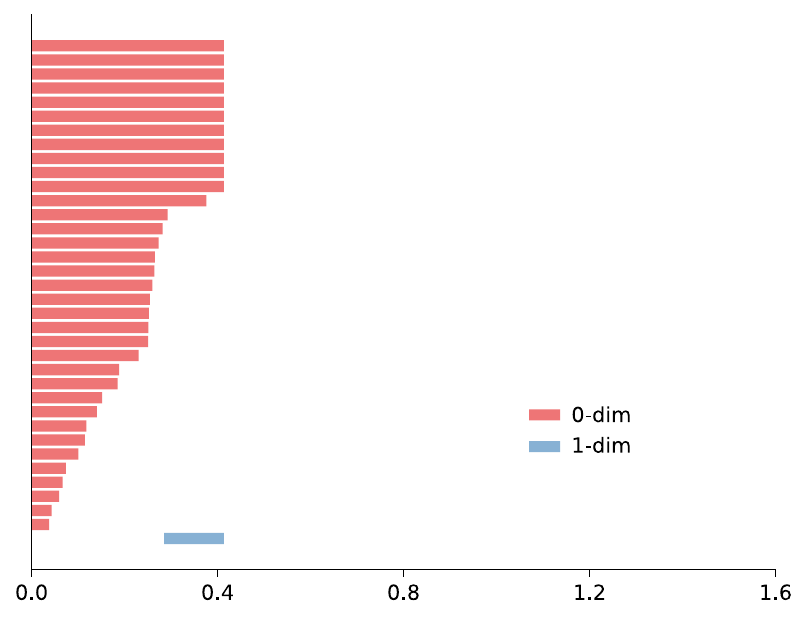}
        \label{fig:pbarcode_04}
    }
    
    \subfloat[$\epsilon=0.7$]{%
        \includegraphics[width=0.4\linewidth]{figures/P8_0.7.pdf}
        \label{fig:p8_07}
    }
        \subfloat[PB up to $\epsilon=0.7$]{%
        \includegraphics[width=0.6\linewidth]{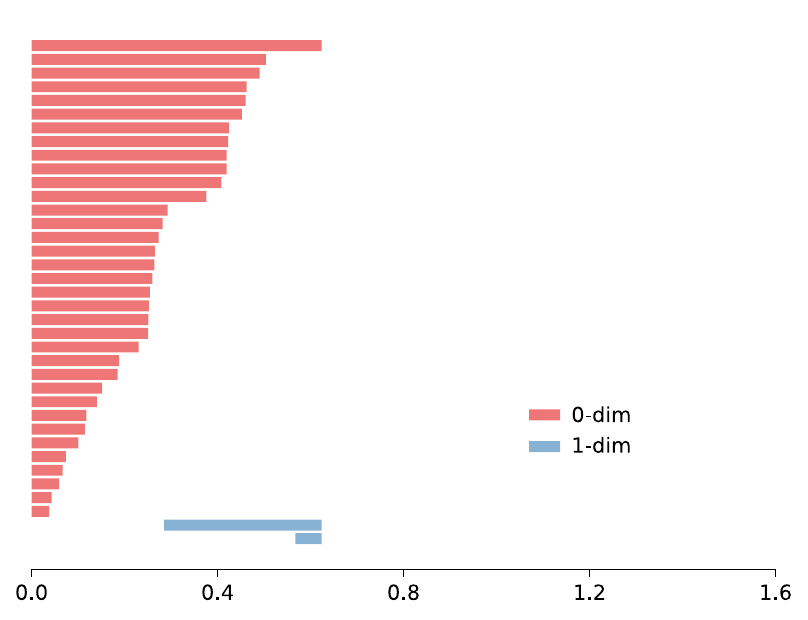}
        \label{fig:pbarcode_07}
    }
    
    \subfloat[$\epsilon=2$]{%
        \includegraphics[width=0.4\linewidth]{figures/P8_2.pdf}
        \label{fig:p8_2e}
    }
    \subfloat[Persistence barcode]{%
        \includegraphics[width=0.6\linewidth]{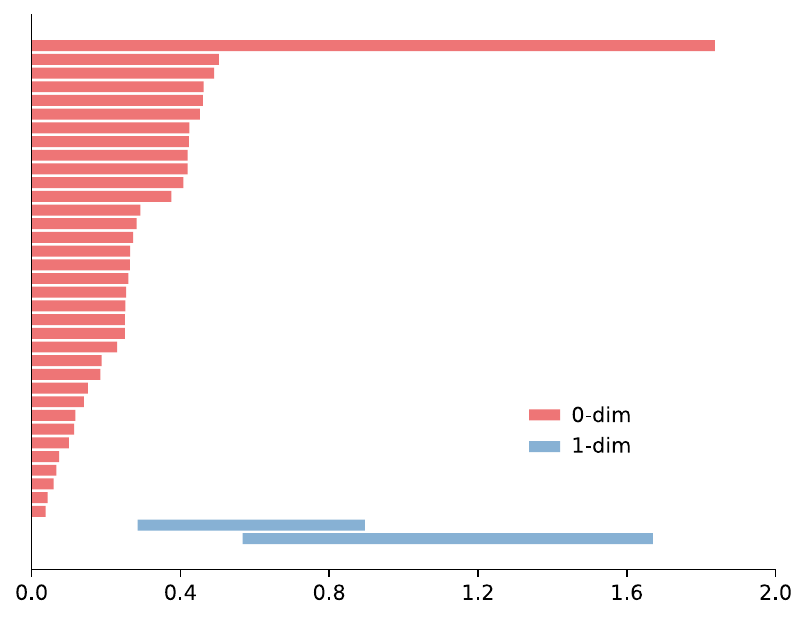}
        \label{fig:pbarcode_2}
    }
    \caption{\footnotesize {\bf Evolution of Persistence Barcode (PB).} In PBs shown on the right, red bars represent $\mathrm{PB}_0(\X)$, and blue bars represent $\mathrm{PB}_1(\X)$. The 8-shaped point cloud $\X$ with 35 points,  initially has 35 components, corresponding to 35 red bars. As $\epsilon$ increases, these components merge. At $\epsilon = 0.4$, only the top 11 red bars remain (b), indicating the number of components in $\mathcal{N}_{0.4}(\X)$ (a). By $\epsilon = 0.7$, the space becomes fully connected (c), and all red bars in $\mathrm{PB}_0(\X)$ terminate, except for the top $\infty$-bar (d). Regarding loops (1-holes), a small loop appears around $\epsilon = 0.3$ (a-b) and persists in $\mathcal{N}_{0.7}(\X)$ (c), while a larger loop emerges around $\epsilon = 0.6$ (d). In $\mathrm{PB}_1(\X)$ (f), two blue bars, $[0.3, 0.9)$ and $[0.6, 1.7)$, correspond to the small and large loops in the 8-shape. \label{fig:point_filtration2} }  
\vspace{-.3in}
\end{wrapfigure}

For point clouds, the number of thresholds directly determines the filtration's resolution. While thresholds are generally evenly spaced, in the presence of outliers, they can be more sparsely distributed after a certain value. Additionally, since computational cost is often a concern with point clouds, it's important to find a balance between the number of thresholds and the associated computational expense.

\subsection{Persistence Diagrams} \label{sec:PD}

After constructing the filtration $\K_1 \subset \K_2 \subset \dots \subset \K_n$ for a data type $\X$, PH systematically tracks the evolution of topological features ($k$-holes) in the filtration sequence and records this information in a \textit{persistence diagram}, which we define next. The nontrivial elements in the homology groups $\h_k(\K_i)$ for $1\leq i\leq n$ represent the $k$-dimensional topological features (or $k$-holes) appearing in the filtration sequence. Furthermore, the inclusion map $\iota: \K_i \hookrightarrow \K_{i+1}$ allows us to determine if a $k$-hole $\sigma$ in $\K_i$ persists in $\K_{i+1}$ through the induced map $\iota_*: \h_k(\K_i) \to \h_k(\K_{i+1})$.

For each $k$-hole $\sigma$, PH records its first appearance in the filtration sequence, denoted $\K_{i_0}$, and its first disappearance in a later complex, $\K_{j_0}$. We define $b_\sigma = \epsilon_{i_0}$ as the \textit{birth time} of $\sigma$ and $d_\sigma = \epsilon_{j_0}$ as the \textit{death time} of $\sigma$, where $\{\epsilon_i\}_{1}^n$ is the threshold set used for the filtration. The difference $d_\sigma - b_\sigma$ is called the \textit{lifespan} of $\sigma$. For example, if a $k$-hole $\tau$ first appears in $\K_3$ and disappears in $\K_7$, we mark the birth time as $b_\tau = \epsilon_3$ and the death time as $d_\tau = \epsilon_7$. The lifespan of $\tau$ is then $\epsilon_7 - \epsilon_3$.

\begin{wrapfigure}{r}{3in}
\vspace{-.15in}
    \centering
    \subfloat[$\epsilon=2$]{%
        \includegraphics[width=0.5\linewidth]{figures/PBarcode_2.pdf}
        \label{fig:p8_2s}    }
    \subfloat[Persistence barcode]{%
        \includegraphics[width=0.5\linewidth]{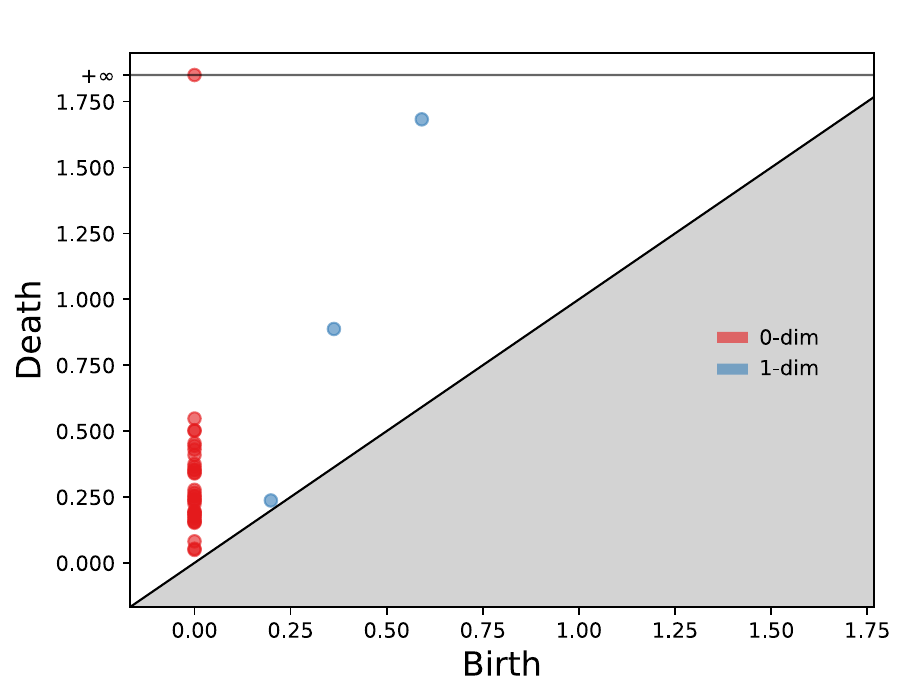}
        \label{fig:pbarcode_2b}}
       \caption{\footnotesize Persistence Barcode (left) and Persistence Diagram (right) for the 8-shaped point cloud $\X$~(\Cref{fig:point_filtration2}). Both representations convey the same information, while the PD plots each bar's birth time as its $x$-coordinate and death time as its $y$-coordinate. For example, the 35 red bars in $\mathrm{PB}_0(\X)$ that begin at 0 are represented as 35 red points along the $x=0$ line in $\PD_0(\X)$, where the $y$-coordinates correspond to the bars' death times. Similarly, the blue points at $(0.3, 0.9)$ and $(0.6, 1.7)$ represent the blue bars $[0.3, 0.9)$ and $[0.6, 1.7)$, respectively. \label{fig:PD}} 
              \vspace{-.15in}
       \end{wrapfigure}

For each nontrivial $\sigma \in \h_k(\K_i)$ for $1 \leq i \leq n$, we represent $\sigma$ with a 2-tuple $(b_\sigma, d_\sigma)$ to denote its birth and death times in the filtration. The collection of all such 2-tuples is called the \textit{persistence diagram} (PD) as depicted in Figure~\ref{fig:PD}. Note that a topological feature with the same birth and death time ($0$ lifespan) is considered a trivial topological feature, and they are represented with diagonal elements in the persistence diagrams. Therefore, the diagonal $\Delta=\{x=y\}\subset \R^2$ is always included in any persistence diagram. 
Then, $k^{th}$ persistence diagram is defined as
$$\PD_k(\X)=\{(b_\sigma, d_\sigma) \mid \sigma \in \h_k(\K_i) \text{ for } b_\sigma \leq i < d_\sigma\}\cup \Delta$$

Since trivial elements $(t,t)$ in PDs can appear multiple times, we take the diagonal $\Delta$ with infinite multiplicity. Essentially, the persistence diagram is a subset of $\R^2$ ($\PD_k(\X) \subset \R^2$), where each point in $\PD_k(\X)$ is either a pair of threshold values $(\epsilon_i, \epsilon_j)$ for some $i < j$, or belongs to the diagonal (trivial element). Here, infinite multiplicity is a technical assumption, which will be important when discussing Wasserstein distance between PDs~(\Cref{sec:wasserstein}).

There is an equivalent concept called the \textit{persistence barcode} (see Figure~\ref{fig:point_filtration2}), which uses bars (half-open intervals) $\{[b_\sigma, d_\sigma)\}$ instead of 2-tuples $\{(b_\sigma, d_\sigma)\}$. The bar notation $[b_\sigma, d_\sigma)$ represents the entire lifespan of a topological feature $\sigma$ as such an interval. In practice, persistence diagrams are more commonly used due to their practicality in applications.

Note that the choice between sublevel and superlevel filtration is irrelevant for the point cloud setting. In the image setting, this choice is not essential, as they essentially convey the same information due to Alexander duality~\cite{hatcher2002algebraic}; when analyzing the topological features of images, the choice of focusing on either the binary image or its complement is not crucial because the topological information of one is directly related to the topological information of the other through this duality. The Alexander duality holds only when the ambient space is quite simple, such as when the space has a structure like an $ m \times n $ rectangle, like images. However, sublevel and superlevel filtrations can yield significantly different results in the graph setting. Carriere et al.~\cite{carriere2020perslay} introduced \textit{extended persistence diagrams} to address this issue.  
This approach enables the birth and death pairs of the persistence diagram to be positioned below the diagonal ($b > d$). By doing so, it captures topological features identified through both superlevel and sublevel filtrations.

\vspace{-.1in}

\subsubsection{Interpretation of Persistence Diagrams}
A persistence diagram $\PD_k(\mathcal{X})$ records the $k$-dimensional topological features of the data $\mathcal{X}$ as a collection of points in $\mathbb{R}^2$. For example, $\PD_0(\mathcal{X})$ records $0$-holes (components), and $\PD_1(\mathcal{X})$ records $1$-holes (loops) appearing in the filtration sequence $\{\mathcal{K}_i\}$. Each point $q_j = (x_j, y_j)$ represents a $k$-dimensional hole $\sigma_j$. For instance, consider a threshold set $\mathcal{I} = \{\epsilon_i = i/10\}_{i=0}^{100}$ with 100 thresholds spanning the interval $[0, 10]$. Suppose we have two points, $q_1 = (0.3, 9.7)$ and $q_2 = (4.2, 4.6)$, in $\PD_2(\mathcal{X})$. These points represent $2$-holes (cavities) $\sigma_1$ and $\sigma_2$ that appear in the filtration $\mathcal{K}_0 \subset \mathcal{K}_1 \subset \dots \subset \mathcal{K}_{100}$.

For $q_1 = (0.3, 9.7)$, the cavity $\sigma_1$ first appears in $\mathcal{K}_3$ and persists until $\mathcal{K}_{97}$, indicating a long lifespan. Conversely, for $q_2 = (4.2, 4.6)$, the cavity $\sigma_2$ first appears in $\mathcal{K}_{42}$ and persists only until $\mathcal{K}_{46}$, indicating a short lifespan. This suggests that $\sigma_1$ represents an important topological feature of the data $\mathcal{X}$, while $\sigma_2$ is likely just \textit{topological noise}.

In general, features with long lifespans (where $y - x$ is large) are located \textit{far from the diagonal} and are considered significant (or "big") features. Features with short lifespans (where $y - x$ is small) are \textit{close to the diagonal} and are considered insignificant (or "small") features. This is why many vectorization techniques try to involve the lifespan information in their computation to give different emphasis to small and big features in the output.

\subsubsection{Wasserstein Distance} \label{sec:wasserstein}

After obtaining the persistence diagrams (PDs) for two datasets, $\mathbf{X}^+$ and $\mathbf{X}^-$, we can assess their topological similarities using these diagrams. One effective approach is to employ a metric in the space of PDs. If the distance between the two persistence diagrams is small, we can conclude that the two datasets exhibit similar topological characteristics; otherwise, they differ.
The most commonly used metric for this purpose is the \textit{Wasserstein distance} (also known as the \textit{matching} or \textit{earth mover distance}) of the Optimal Transport Theory (see an ICML tutorial in~\cite{bunne2023optimal}), which is defined as follows:

Let $\PD(\X^+)$ and $\PD(\X^-)$ be the persistence diagrams for the datasets $\X^+$ and $\X^-$, respectively (we omit the dimensions in PDs for simplicity). Denote $\PD(\X^+) = \{q_j^+\} \cup \Delta$ and $\PD(\X^-) = \{q_l^-\} \cup \Delta$, where $\Delta$ represents the diagonal (indicating trivial cycles) with infinite multiplicity, and $q_j^\pm = (b^\pm_j, d_j^\pm) \in \PD(\X^\pm)$ represents the birth and death times of a topological feature $\sigma_j$ in $\X^\pm$. Let $\phi: \PD(\X^+) \to \PD(\X^-)$ represent a bijection (matching). The presence of the diagonal $\Delta$ on both sides ensures the existence of these bijections even if the cardinalities $|\{q_j^+\}|$ and $|\{q_l^-\}|$ differ. This means that for optimal matching, some points in $\{q_j^+\}$ and $\{q_l^-\}$ are matched to diagonal elements whenever necessary. Then, the $x^{th}$ Wasserstein distance $\W_p$ is defined as
\[
\W_p(\PD(\X^+), \PD(\X^-)) = \min_{\phi} \left( \sum_j \|q_j^+ - \phi(q_j^+)\|_\infty^p \right)^{\frac{1}{p}}, \quad p \in \mathbb{Z}^+.
\]

\begin{wrapfigure}{r}{2.5in}
    \centering
     \includegraphics[width=\linewidth]{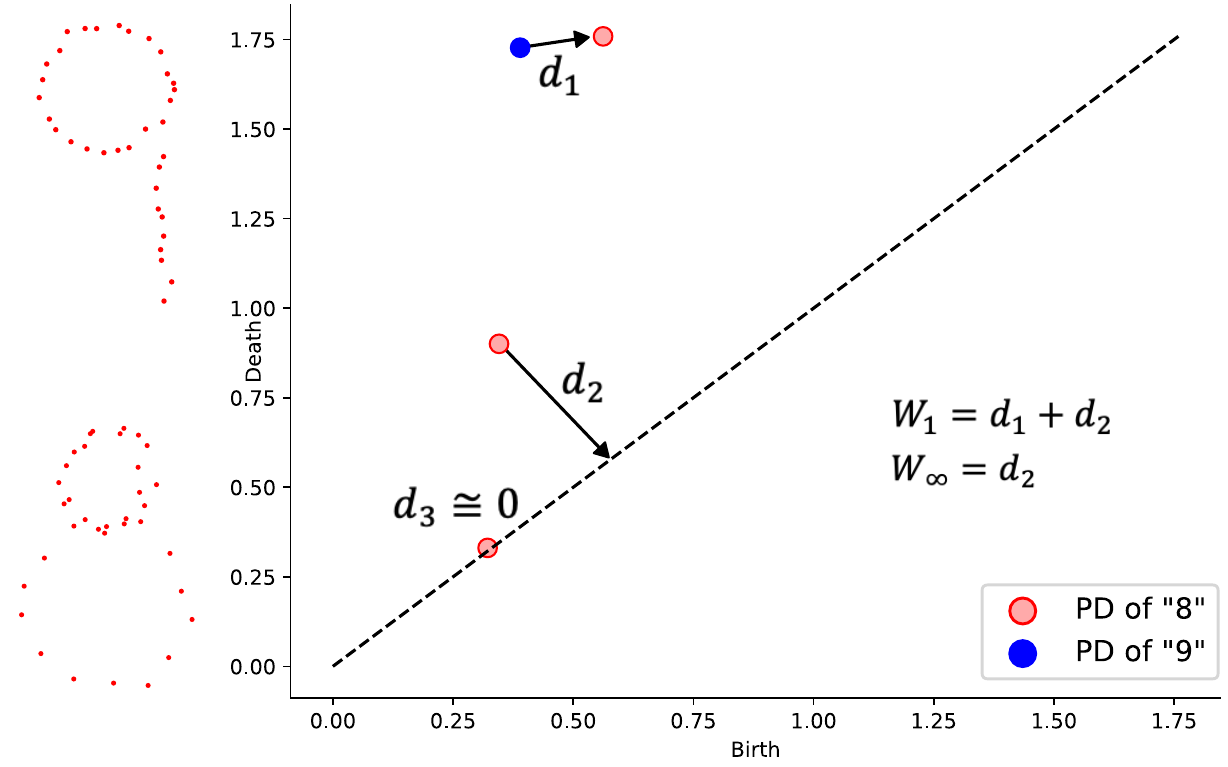}
    \caption{Wasserstein distances $\W_1$ and $\W_\infty$ (bottleneck) between $\PD_1(8)$ and $\PD_1(9)$ for the point clouds of "8" and "9" on the left.}
    \label{fig:wasserstein}
    \vspace{-.1in}
\end{wrapfigure}

Here, $\|(x_1,y_1)-(x_2,y_2)\|_\infty=\sup \{|x_1-y_1|,|x_2-y_2|\}$ is called {\em supremum norm} (or $l^\infty$-norm). When $p=\infty$, $\W_\infty$ is called the {\em bottleneck distance}. i.e.,
$\W_\infty(\PD(\X^+), \PD(\X^-)) = \max_j \{\|q_j^+ - \phi(q_j^+)\|_\infty\}$ (See \Cref{fig:wasserstein}).

In applications, the most common choices for the Wasserstein distance are $ p=1, 2, $ and $ \infty $. The bottleneck distance, unlike the $p$-Wasserstein distance, is insensitive to the number of points in $ \PD_k(\X^\pm) $ and instead focuses on the distance between the farthest points in the optimal matching. For instance, ignoring the diagonal, if $ \PD_k(\X^+) $ has 100 points and $ \PD_k(\X^-) $ has only three points, the bottleneck distance is primarily determined by finding the three closest points in $ \PD_k(\X^+) $ to those in $ \PD_k(\X^-) $ and taking the maximum of these distances. Conversely, when using $ p=1 $ or $ p=2 $, the quantity of points in $ \PD_k(\X^\pm) $ becomes significant as all points contribute to the distances, even if only slightly. Therefore, if there are only a few points in $ \{\PD_k(\X_j)\} $ and the focus is on the distances of the most critical features, the bottleneck distance is preferable. However, if there are many points in $ \{\PD_k(\X_j)\} $ and the primary topological patterns arise from the quantity and location of smaller features, then choosing $ p=1 $ (or $ p=2 $) for the Wasserstein distance would be more suitable.

There are several ways to utilize the Wasserstein distance in applications. One common approach is to bypass the vectorization step, which requires some choices, and directly compare persistence diagrams. For instance, to determine if two datasets have similar shapes or structures, one can compute the persistence diagrams for both datasets and then calculate the Wasserstein distance between these diagrams. A smaller distance indicates more similar shapes. Another application lies in unsupervised learning, where datasets are clustered based on their topological similarity using persistence diagrams, utilizing the Wasserstein distance to measure distances between them.

\subsection{Integrating PDs to ML tasks 1: Vectorizations} \label{sec:vectorization}

To effectively use persistence diagrams (PDs) in ML applications, it is crucial to convert them into numerical or vector formats, allowing for the seamless integration of TDA outputs into standard ML workflows. Although PDs capture the birth and death of topological features within data, their variable size and structure make them challenging to handle directly. For example, in a binary graph classification task with 1000 graphs (400 in Class A and 600 in Class B), PDs might visually highlight differences between the classes. However, applying traditional statistical tools to PDs, which are subsets of $\R^2$, is problematic; for instance, it is not possible to compute averages or confidence intervals for each class. Moreover, directly inputting PDs into ML models is impractical because their variable point count conflicts with the fixed-size input required by most ML algorithms.

Vectorization addresses this by making PDs compatible with traditional ML and statistical techniques. Below, we outline the most common vectorization methods used in practice. For a comprehensive review, refer to~\cite{ali2023survey}.

In the following, for a fixed dataset $\mathcal{X}$, a threshold set $\{\epsilon_i\}_{i=1}^n$, and an induced filtration $\{\mathcal{K}_i\}_{i=1}^n$, we will introduce several vectorization methods. It is important to note that all these vectorization methods are applicable to any type of data since they simply transform a given PD into vectors. However, the effectiveness and common usage of certain vectorization methods can vary depending on the data type and the density or sparsity of the PDs. Although we present various vectorization methods and their configurations here, \textit{those who prefer to avoid the intricacies of vectorization selection and hyperparameter tuning can opt for automated approaches, as detailed in~\Cref{sec:vecNN}}.

\paragraph*{Function vs. Vector Format.} In the following, we introduce several vectorization methods. For each method, we will describe how to convert a given PD into a \textit{vector} or a \textit{function} depending on the setting. Depending on the context, one of these formats might be preferable (visualization, ML input, etc.), however, both formats can be easily converted to each other. For example, if we have a function $f:[\epsilon_1,\epsilon_n]\to\mathbb{R}$ defined over an interval, we can convert $f$ into an $N$-dimensional vector by sampling the values at $f(t_j)$ where $t_j = \epsilon_1 + \frac{j}{N}(\epsilon_n - \epsilon_1)$ and $j$ ranges from 1 to $N$. For each point $t_j$, we calculate $f(t_j)$. The vector $\mathbf{v}_f$ is then  $\mathbf{v}_f=[f(t_1), f(t_2), \dots, f(t_N)]$. Conversely, for a given $N$-dimensional vector $\mathbf{v}=[v_1 \ v_2 \ \dots v_N]$, we can convert it to a function via a step function or linear spline interpolation, e.g., $g_\mathbf{v}:[\epsilon_1,\epsilon_n]\to\mathbb{R}$ such that $g_\mathbf{v}(t_j)=v_j$ for $t_j=\epsilon_1+\frac{j}{N}(\epsilon_n-\epsilon_1)$. Then, for any point $s\in[t_j,t_{j+1}]$, we define the linear extension $g(s)=g(t_j)+\frac{g(t_{j+1})-g(t_j)}{t_{j+1}-t_j}(s-t_j)$. Hence, any of the following vectorizations can be taken as a function or vector depending on the need.

\paragraph*{Topological Dimensions and Final Topological Vector.} In each vectorization method, we convert a persistence diagram $\PD_k(\X)$ into a vector (or array) $\mathbf{v}_k$ where $k$ represents a topological dimension of $k$-holes. In particular, each persistence diagram produces a different vector. The user needs to decide which topological dimensions to be used in the method. After getting an $N$-dimensional vector for each dimension, a common method is to concatenate them to obtain the final topological vector. Hence, if we use $m$ different topological dimensions, we have $(m\cdot N)$-dimensional final vector $\mathbf{v}(\X)=\mathbf{v}_0\mathbin\Vert \mathbf{v}_1\mathbin\Vert \dots\mathbin\Vert \mathbf{v}_{m-1}$. In most applications, the common dimensions used are $k=0$ and $k=1$, and hence the final topological vector is $2N$-dimensional, i.e., $\mathbf{v}(\X)=\mathbf{v}_0\mathbin\Vert \mathbf{v}_1$.

\paragraph*{Betti vectors} One of the most straightforward and interpretable vectorization methods in TDA is the {\em Betti function}. The $k^{\text{th}}$ Betti number of a topological space $\K$ essentially counts the total number of $k$-dimensional holes in $\K$. More formally, it is defined as $\beta_k(\K) = \text{rank}(H_k(\K))$, which is the rank of the $k^{\text{th}}$ homology group of $\K$. For example, $\beta_0(\K)$ represents the number of connected components in $\K$, while $\beta_1(\K)$ indicates the number of $1$-dimensional holes.

\begin{wrapfigure}{r}{0.4\linewidth}
\vspace{-.1in}
\centering
\includegraphics[width= 0.98\linewidth]{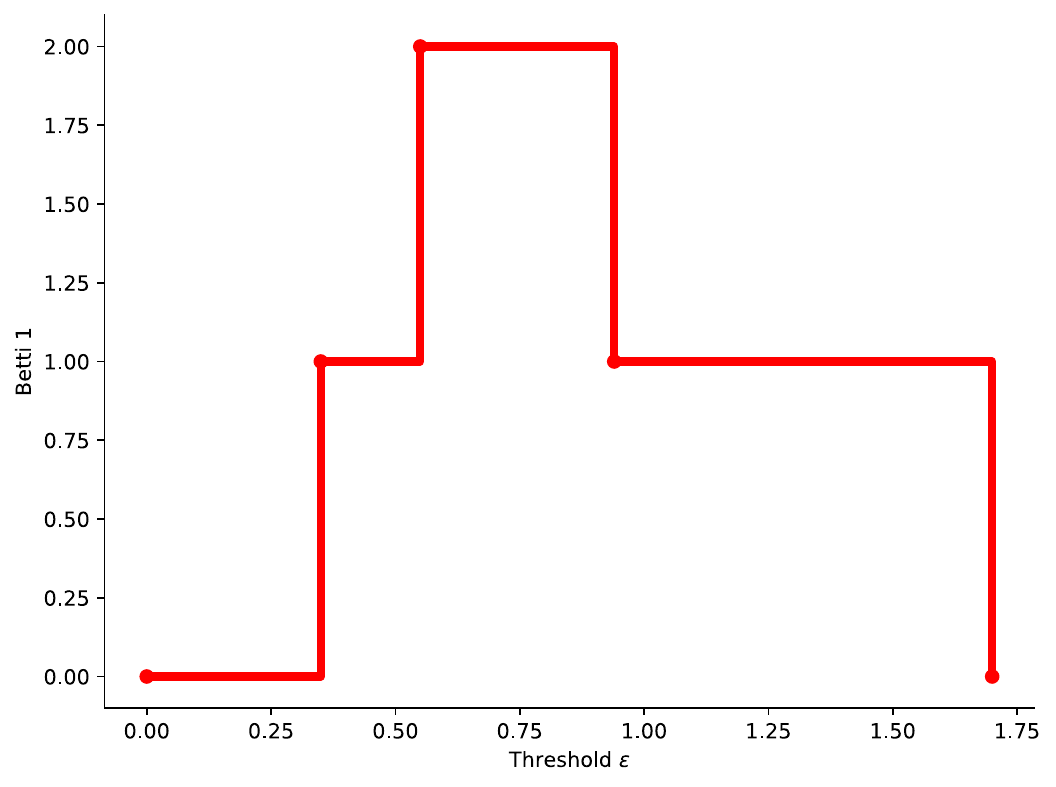}
\caption{\footnotesize \textbf{Betti Function.} The step function represents the first Betti number ($\beta_1$) over the figure-eight-shaped point cloud of Figure~\ref{fig:graph-filtration} as a function of the threshold $\epsilon$. The function shows the number of 1-dimensional holes (loops) in the point cloud. Initially, there are no loops ($\beta_1 = 0$), then a single loop appears at $\epsilon = 0.35$ and persists until $\epsilon = 0.94$, during which another loop appears at $\epsilon = 0.55$ and vanishes at $\epsilon = 0.94$. The final loop disappears at $\epsilon = 1.70$, returning $\beta_1$ to 0. 
 \label{fig:betti_function}} 
 \vspace{-.2in}
\end{wrapfigure}
For given filtration $\{\K_i\}_1^n$, we define $k^{th}$  Betti vector $\vec{\beta}_k(\X)= [ \beta_k(\K_1) \ \beta_k(\K_2) \dots \beta_k(\K_n)]$. Therefore, for each topological dimension $k$, we obtain a Betti vector of the size of a number of thresholds. For example, $k=0,1$ with $n=50$ results in $50$-dimensional vector for each dimension. In ML applications, typically, these vectors are concatenated to create a $100$-dimensional topological vector as output.

\begin{example} \label{ex:betti} Consider the point cloud $\X$ in~\Cref{fig:point_filtration2}. We define a filtration using five threshold values $\wh{\epsilon} = [0, 0.25, 0.75, 1.5, 1.75]$, where $\K_i = \N_{\e_i}(\X)$. The Betti-0 vector $\vec{\beta}_0(\X) = [35, 20, 1, 1, 1]$ is obtained, with $\beta_0(\X) = 35$, $\beta_0(\N_{0.25}(\X)) = 20$, and $\beta_0(\N_\epsilon(\X)) = 1$ for $\epsilon \geq 0.7$. This indicates that $\N_{0.25}(\X)$ consists of 20 connected components. The Betti-0 vector can be easily determined from the persistence barcode $\mathrm{PB}_0(\X)$~(\Cref{fig:pbarcode_2}) by counting the number of red bars intersected by a vertical line $x=\e_i$, representing the number of persistent topological features at the threshold $\e_i$. Thus, with 5 thresholds, we obtain a 5-dimensional vector, $\vec{\beta}_0(\X)$.

Similarly, we can determine the corresponding Betti-1 vector by examining the blue bars in the persistence barcode~(\Cref{fig:pbarcode_2}). We observe no blue bar at $x = 0$, $0.25$, and $1.75$. Furthermore, there are two blue bars at $x = 0.75$ and one blue bar at $x = 1.5$. Therefore, the Betti-1 vector is $\vec{\beta}_1(\X) = [0, 0, 2, 1, 0]$. This indicates that there are no 1-dimensional holes (1-holes) in $\X = \N_0(\X)$, $\N_{0.25}(\X)$, or $\N_{1.75}(\X)$, while $\N_{0.75}(\X)$ contains two 1-holes and $\N_{1.5}(\X)$ contains one. It is also common to use Betti vectors as Betti functions via step functions~(See~\Cref{fig:betti_function}).
Similarly, using 21 equally spaced thresholds $\wh{\e} = \{0, 0.1, 0.2, \dots, 2\}$ would yield 21-dimensional vectors $\vec{\beta_0}$ and $\vec{\beta_1}$.
\end{example}

We note that Betti vectors do not require the computation of persistence diagrams. Therefore, there are computationally more effective ways to produce Betti vectors~\cite{hayakawa2022quantum,ameneyro2024quantum}. Another favorable aspect of Betti vectors is their \textbf{ease of interpretation}. Simply put, $\beta_k(\epsilon_i)$ is equal to the number of $k$-dimensional topological features in $\mathcal{K}_i$.

 \begin{wrapfigure}{r}{2in}
\vspace{-.15in}
\centering
\includegraphics[width= 0.98\linewidth]{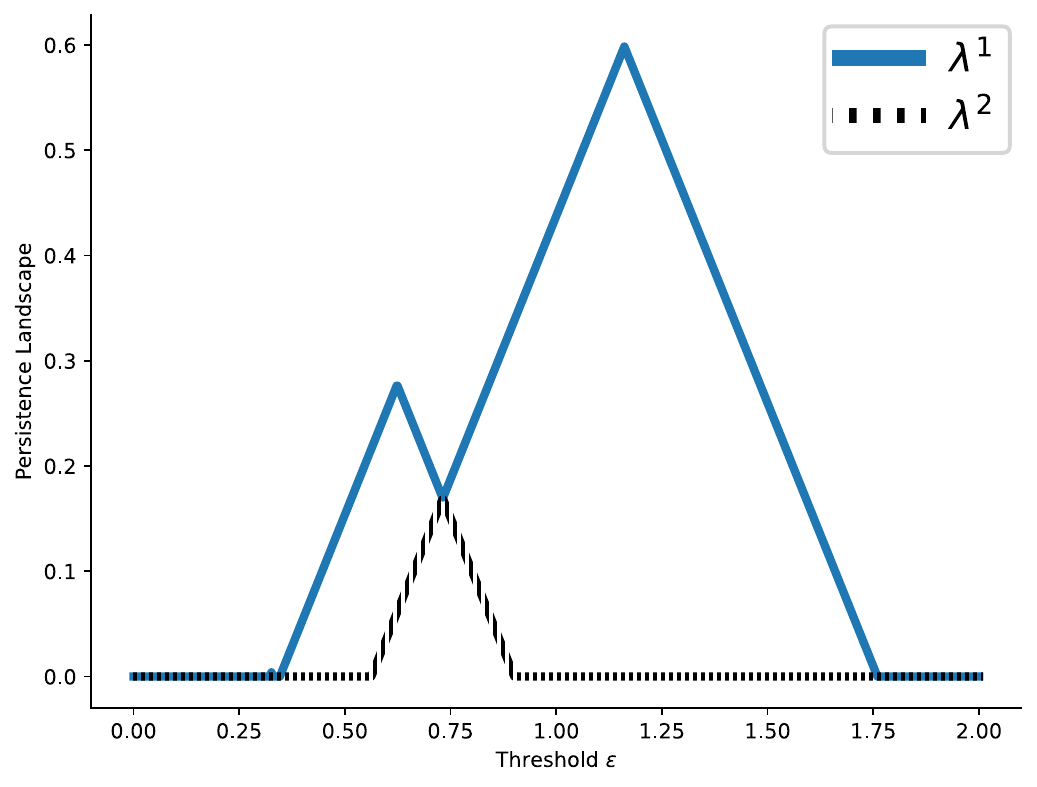}
\caption{\textbf{Persistence Landscape (PL).} The plot shows the PL for $\PD_1(\X)$ for the 8-shaped point cloud filtration in~\Cref{ex:betti}. In the plot, the blue function corresponds to $\lambda^1$, and the orange function to $\lambda^2$. \label{fig:landscape}}
\vspace{-.15in}
\end{wrapfigure}

\paragraph*{Persistence Landscapes} Persistence Landscapes are one of the first vectorization methods in TDA, introduced by P. Bubenik, directly utilizing the lifespan information~\cite{bubenik2015statistical}. In particular, in this vectorization, the points away from the diagonal (large features) are easily distinguished and promoted. For a given persistence diagram $\PD_k(\X)=\{(b_i,d_i)\}$, we first define generating functions $\Lambda_i$ for each $(b_i,d_i)\in \PD_k(\G)$, i.e., $\Lambda_i:[b_i,d_i]\to\R$ is a piecewise linear function obtained by two line segments starting from $(b_i,0)$ and $(d_i,0)$ connecting to the same point $(\frac{b_i+d_i}{2},\frac{b_i-d_i}{2})$. Then, we define several piecewise-linear functions $\{\lambda^m\}$ in the following way. For each $t\in[\epsilon_1,\epsilon_n]$, we check all generating functions $\{\Lambda_i(t)\}$, and we mark $m^{th}$ largest value. In particular, $m^{th}$ \textit{Persistence Landscape} function $\lambda^m(\X):[\epsilon_1,\epsilon_n]\to\R$ for $t\in [\epsilon_1,\epsilon_n]$ is defined as $\lambda^m(\X)(t)=m^{th}\ \max_i\{\Lambda_i(t)\}$ (See~\Cref{fig:landscape}). Note that the first and second persistence landscapes are the most commonly used vectors, and they are used with concatenation in the applications.

  \begin{wrapfigure}{r}{2.5in}
\centering
\includegraphics[width=\linewidth]{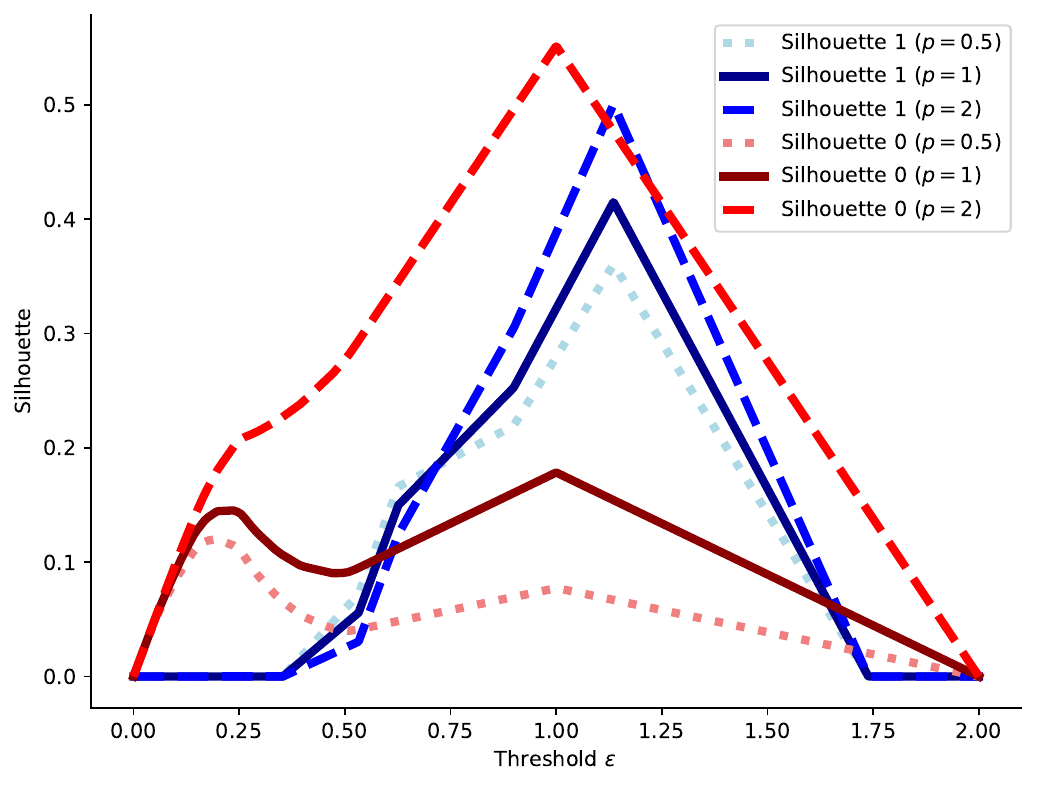}
\caption{\footnotesize \textbf{Silhouette Functions.} The plot shows the silhouette functions with tuning parameters $p=0.5,1,2$ $\PD_1(\X)$ for the 8-shaped point cloud filtration given in~\Cref{ex:betti}. \label{fig:silhouette}}
 
\vspace{-.15in}
\end{wrapfigure}

\paragraph*{Silhouettes} While persistence landscapes are among the first vectorizations to effectively utilize lifespan information, the need to consider $m$ different maxima makes it difficult to use in ML applications.
In \cite{chazal2014stochastic}, Chazal et al. proposed a practical modification called the {\em Silhouette} for persistence landscapes. This modification introduces a tuning parameter $p$ to better utilize the lifespans of topological features. For a persistence diagram $\PD_k(\X)=\{(b_i,d_i)\}_{i=1}^N$, let $\Lambda_i$ be the generating function for $(b_i,d_i)$ as defined in the persistence landscapes. The \textit{Silhouette} function $\psi$ is defined as: $\Psi_k(\X)=\dfrac{\sum_{i=1}^N w_i\Lambda_i(t)}{\sum_{i=1}^m w_i}, \ t\in[\epsilon_1,\epsilon_n],$
where the weight $w_i$ is usually chosen as the lifespan $(d_i-b_i)^p$. Thus, the $p$-silhouette function $\psi^p_k(\X):[\epsilon_1,\epsilon_n]\to \R$ is defined as: 

\hspace{2cm} $\Psi^p_k(\X)=\dfrac{\sum_{i=1}^N (d_i-b_i)^p\Lambda_i(t)}{\sum_{i=1}^N (d_i-b_i)^p}$

\vspace{.1cm}

The tuning parameter $p$ is crucial as it adjusts the silhouette function's emphasis on topological features with varying lifespans. When $p<1$, shorter lifespans $(d_i-b_i)$ are given more weight, emphasizing smaller features. Conversely, shorter lifespans are down-weighted when $p>1$, highlighting larger features. Common choices for $p$ are $\frac{1}{2}$, 1, and 2 (See~\Cref{fig:silhouette}). If the persistence diagram has a few points and the goal is to emphasize significant features, $p=2$ would be a good choice. If there are many points and the key information comes from smaller features, $p=\frac{1}{2}$ is considered more suitable. Similarly, Silhouette functions $\psi^p_k(\X)$ can be converted to vectors for each dimension $k$, and concatenation of these vectors can be used in the application as the topological vector of the dataset.

 \begin{wrapfigure}{r}{2.5in}
 \vspace{-.15in}
\centering
\includegraphics[width= 0.98\linewidth]{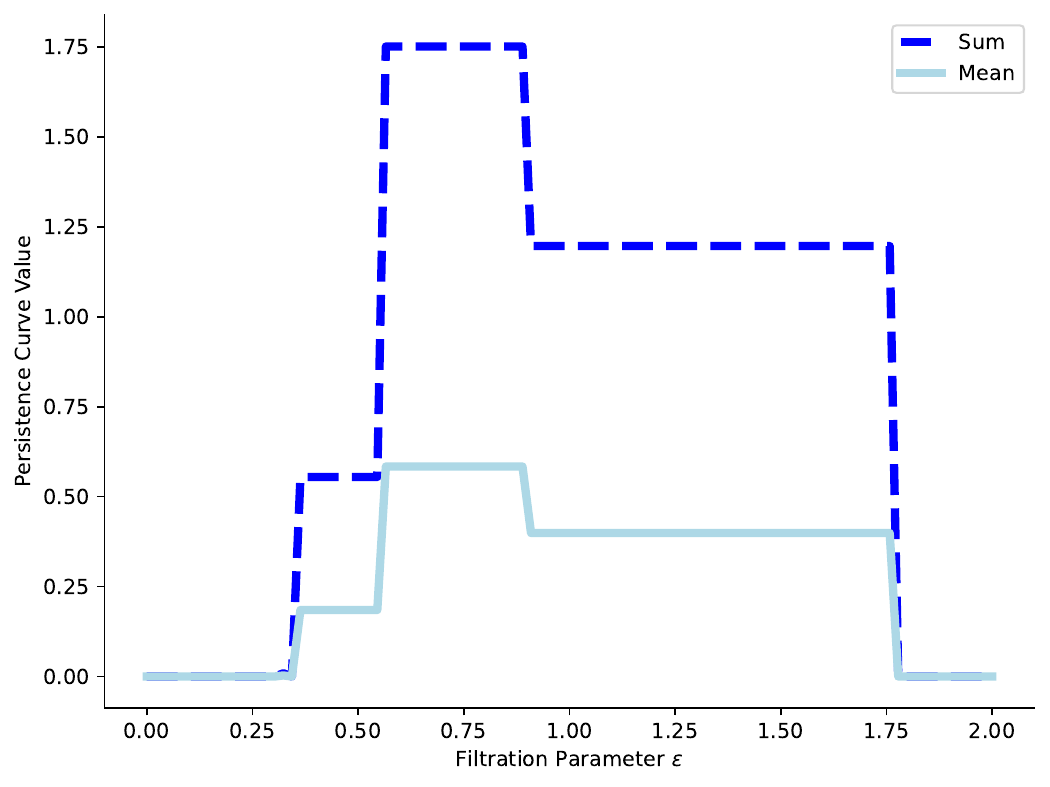}
\caption{\footnotesize \textbf{Persistence Curves.} The plot shows the persistence curves for $\PD_1(\X)$ for the 8-shaped point cloud filtration in~\Cref{ex:betti} by using two different summary statistics: sum and mean. The dashed line represents the persistence curve using the sum of lifespans, which emphasizes the cumulative contribution of all loops. The solid line represents the persistence curve using the mean of lifespans, which highlights the average significance of the loops. This visualization illustrates how different summary statistics influence the interpretation of topological features in the data.}
\label{fig:persistence_curves}
 \vspace{-.15in}
\end{wrapfigure}
\paragraph*{Persistence Curves} The Persistence Curve (PC) framework, introduced by Chung et al.~\cite{chung2019persistence}, offers a structured approach to the vectorization process. The core idea revolves around utilizing a generating function $\psi$, a summary statistic $T$, and a filtration parameter $\epsilon$. From these elements, a PC is defined as: 

$PC(\D, \psi, T )(t) = T ([\psi(\D; b, d, \epsilon) \mid (b, d) \in \D_\epsilon ]), \epsilon \in \R.$ 

Here, $\D_\epsilon$ represents the points within the persistence diagram $\D$ that fall inside a region $\Delta_\epsilon$ varying with the filtration value $\epsilon$.  
The function $\psi$ takes as input a point from the persistence diagram $\D$,  and the filtration parameter, $\epsilon$ and outputs a real number.  This function can be chosen to prioritize certain points in the PD or to emphasize specific features. Finally, the statistic $T$  acts upon a multiset of values, aggregating them into a single real number.  Common examples include sum, mean, or max.

PCs provide a general and unifying framework for vectorization methods.
By selecting different combinations of $\psi$ and $T$, one can generate various functional summaries of the PD, each potentially highlighting different aspects of the data. Many established vectorizations of PDs can be represented within the PC framework, e.g., Betti curves, persistence landscapes, silhouettes, and entropy curves. See~\cite{chung2019persistence} for more details.

\begin{wrapfigure}{r}{3in}
\vspace{-.2in}
    \centering
    \subfloat{%
        \includegraphics[width=0.32\linewidth]{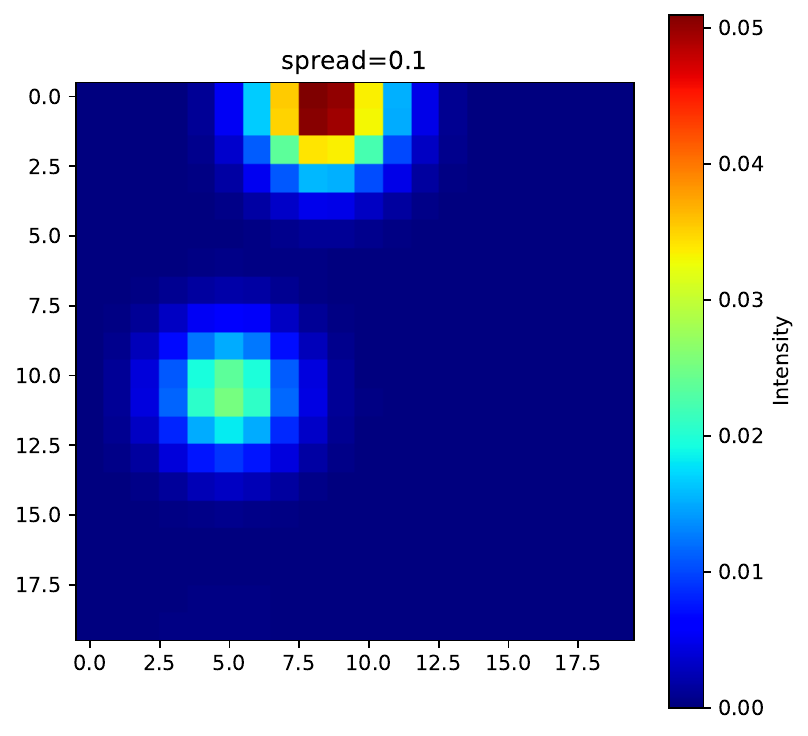}
        \label{fig:p8_04b}}
    \subfloat{%
        \includegraphics[width=0.33\linewidth]{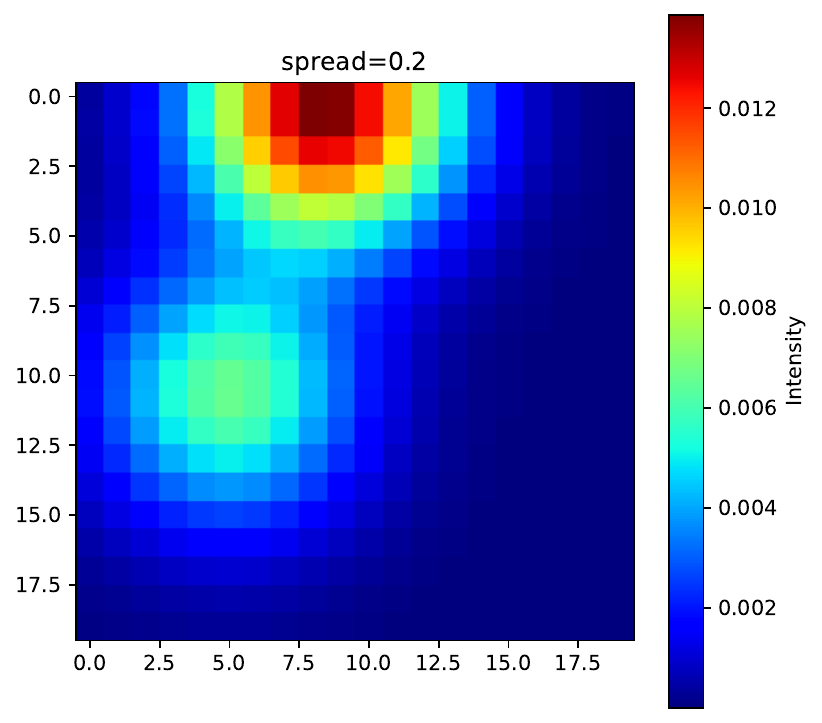}
        \label{fig:p8_07b}}
    \subfloat{%
        \includegraphics[width=0.34\linewidth]{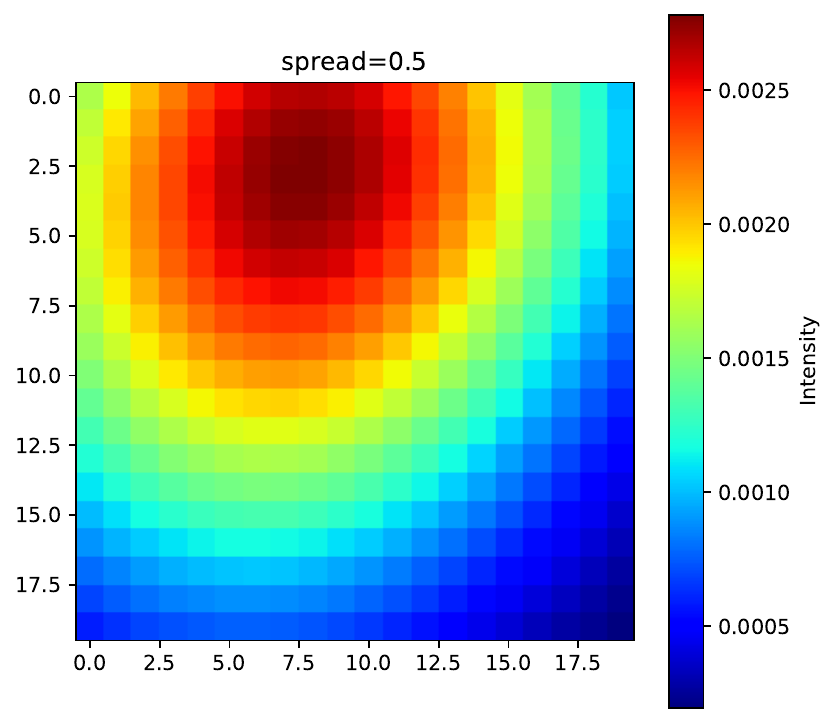}
        \label{fig:p8_2b}}
    \caption{\footnotesize {\bf Persistence images (PI).} PIs are vectorization with 2D (matrix) output. Here, we give examples of PIs for $\PD_1(\X)$ ($8$-shaped point cloud in \Cref{ex:betti}) with different spread values. The spread parameter $\sigma$ controls the standard deviation of the Gaussian functions used to smooth the persistence points in the persistence diagram. As illustrated in the figures, as $\sigma$ increases, the generating Gaussian functions are even more spread out, producing a flatter persistence image.    \label{fig:persistence_image}}
        
        \vspace{-.1in}
\end{wrapfigure}

\paragraph*{Persistence Images} Our next vectorization is Persistence Images, introduced by Adams et al.~\cite{adams2017persistence}. Unlike most vectorizations, Persistence Images, as the name suggests, produce $2D$-arrays (tensors). The idea is to capture the location of the points in the PDs with a multivariable function by using the $2D$ Gaussian functions centered at these points. For $PD(\G)=\{(b_i,d_i)\}$, let $\phi_i$ represent a $2D$-Gaussian centered at the point $(b_i,d_i)\in \R^2$. Then, one defines a multivariable function, \textit{Persistence Surface}, $\wt{\mu}=\sum_iw_i\phi_i$ where $w_i$ is the weight, mostly a function of the life span $d_i-b_i$. To represent this multivariable function as a $2D$-vector, one defines a $k\times l$ grid (resolution size) on the domain of $\wt{\mu}$, i.e., threshold domain of $PD(\G)$. Then, one obtains the \textit{Persistence Image}, a $2D$-vector (matrix) $\vec{\mu}=[\mu_{rs}]$  of size $k\times l$ such that 
$$\mu_{rs}=\int_{\Delta_{rs}}\wt{\mu}(x,y)\,dxdy \quad \text{where } \Delta_{rs}= \text{ pixel with index }rs \text{ in the } k\times l \text{ grid.}$$

Note that the resolution size $k\times l$ is independent of the number of thresholds used in the filtering; the choice of $k$ and $l$ is completely up to the user. There are two other important tuning parameters for persistence images, namely the weight $w_i$ and the variance $\sigma$ (the width of the Gaussian functions). Like Silhouettes, one can choose $w_i=(d_i-b_i)^p$ to emphasize large or small features in the PD. Similarly, the width parameter $\sigma$ determines the sharpness of Gaussian, where smaller $\sigma$ would make the Gaussian functions more like Dirac $\delta$-function, and larger $\sigma$ would make the Gaussians flat. Depending on the context, $\sigma$ can be chosen a constant (e.g. $\sigma=0.1$) or depending on the point $(b_i,d_i)$, e.g., $\sigma_i=k(d_i-b_i)$ for some constant $k>0$.

\paragraph*{Kernel Methods} \label{sec:kernel} Kernel methods provide an alternative to traditional direct vectorization techniques used to transform PDs into formats suitable for ML algorithms. In ML, these methods utilize a mathematical function known as a \textit{kernel} to analyze data and discern patterns. Unlike earlier approaches that directly represent PDs using vectors or functions, kernel methods compute a \textit{similarity score} as an inner product between pairs of PDs in a high-dimensional space without explicitly mapping the data. This approach is advantageous because it adapts well to ML techniques like support vector machines (SVMs) and kernel principal component analysis (KPCA). Therefore, kernel methods are well-suited for tasks such as classification, regression, and principal component analysis.

One common such method is the persistence weighted Gaussian kernel (PWGK)~\cite{kusano2016persistence}. It enhances the Gaussian kernel by incorporating weights based on the significance of topological features. This approach amplifies the influence of important features while reducing noise (i.e., short-lived holes) impact. For instance, it assigns weights to points $p=(b,d)\in\PD(\X)$ proportional to their lifespans $w(p)=(d-b)$. In particular, PWGK is defined as
\[ 
\mathbf{K}(\PD(\X^+), \PD(\X^-)) = \sum_{p \in \PD(\X^+)} \sum_{q \in \PD(\X^-)} w(p) w(q) k(p, q)
\]

where $k(p, q) = \exp\left(-\frac{\|p - q\|^2}{2\sigma^2}\right)$ denotes the Gaussian kernel function.

Another significant approach is the sliced Wasserstein kernel (SWK)~\cite{carriere2017sliced}, which computes the Wasserstein distance between PDs and integrates it into a kernel framework. This method employs Optimal Transport Theory to establish a meaningful metric for comparing PDs. Although kernel methods can yield better results in some settings, they can be computationally intensive and impractical for large datasets due to the high computational costs associated with computing the kernel matrix. In particular, computing kernels takes quadratic time in the number of diagrams, while vectorizing PDs takes only linear time.

\subsubsection{Stability} In most applications, the stability of vectorization is vital for statistical and inferential tasks. Essentially, stability means that a small change in the persistence diagram (PD) should not lead to a significant change in its vectorization. In particular, if two PDs, $\PD_k(\X^+)$ and $\PD_k(\X^-)$, are close, their corresponding vectorizations, $\vec{\beta}_k(\X^+)$ and $\vec{\beta}_k(\X^-)$, should also be close. This ensures that the vectorization process preserves the structural properties of the data. Therefore, when two persistence diagrams $\PD_k(\X^\pm)$ are similar, it implies that the datasets $\X^+$ and $\X^-$ share similar shape characteristics. If these datasets are intuitively expected to belong to the same class, their vectorizations $\vec{\beta}_k(\X^\pm)$ should likewise remain close.

To formalize this concept, we need to define what constitutes a "small/big change" or what it means for PDs to be close. This requires a distance (metric) in the space of PDs, with the most common being the \textit{Wasserstein distance} (or matching distance) as defined in \Cref{sec:wasserstein}. Similarly, we need a metric for the space of vectorizations, such as the Euclidean distance $\|\vec{\beta}(\X^+) -\vec{\beta}(\X^-)\|$ for $\vec{\beta} \in \R^N$. Thus, for a given vectorization $\vec{\beta}$, we call it stable (with respect to the Wasserstein-$p$ metric) if it satisfies the stability equation
\[ \|\vec{\beta}_k(\X^+) -\vec{\beta}_k(\X^-)\|\leq C \cdot \W_p(\PD_k(\X^+), \PD_k(\X^-)) \]
This ensures that if $\W_p(\PD_k(\X^+), \PD_k(\X^-))$ (the distance between PDs) is small, then the distance $\|\vec{\beta}_k(\X^+) -\vec{\beta}_k(\X^-)\|$ between the corresponding vectorizations will also be small.
Among the methods described earlier, persistence landscapes, silhouettes, persistence images, and most kernel methods are stable vectorizations, while Betti functions are generally unstable.

\subsubsection{Choice of Vectorization and Hyperparameters}  
The choice of vectorization method should align with the characteristics of your data and the problem at hand. If your data contains a few prominent topological features that are crucial to the task, Silhouettes with $p \geq 2$ or Persistence Images may be the most suitable options. These methods are also effective when dealing with noisy data, allowing you to filter out less significant features. Conversely, if your data generates a high number of small features, where the task hinges on their location and density—in other words, when the noise itself carries important information—Betti curves, Persistence curves, Silhouettes with $p\leq 0.5$ and kernel methods are likely to yield strong performance. On the other hand, for point cloud data, Turkes et al.~\cite{turkes2022effectiveness} provide valuable insights on how to effectively utilize PH for shape recognition. Lastly, if interpretability is a priority, Betti Curves stands out as the most interpretable vectorization method. Note that most vectorizations are computationally efficient and require minimal time compared to the computation of PDs.

\paragraph*{Hyperparameters: } Each vectorization method comes with its own set of hyperparameters that need to be carefully tuned to maximize performance. In all of them, the choice of \textit{thresholds} is one of the key hyperparameters we discussed in~\Cref{sec:threshold}.

In addition to thresholds, many vectorization methods have \textit{tuning parameters} that are used to adjust sensitivity to topological noise. Typically, topological features with short lifespans (i.e., those close to the diagonal in PDs) are regarded as noise. These features, represented as pairs \(\{(b,d)\}\) with short lifespans \((d-b)\), are generally easy to detect, leading to many tuning parameters being tied to the lifespan.
For example, in persistence landscapes, the number of landscapes (\(m\)) specifies how many maxima are included in the representation. A higher number of landscapes provides a richer depiction of topological features but increases computational complexity. The first few landscapes (e.g., \(1^{st}\) and \(2^{nd}\) landscapes) emphasize features with the largest lifespans.
In silhouettes, the tuning parameter \(p\) in the weight function \(w_i = (d_i - b_i)^p\) determines the emphasis of the vectorization. A higher \(p\) (e.g., \(p \geq 2\)) de-emphasizes features with short lifespans, while a smaller \(p\) (e.g., \(p \leq 0.5\)) increases the importance of topological noise.

For persistence images, the spread (\(\sigma\)) controls the width of the Gaussian kernels applied to each persistence point. A smaller spread results in sharper, more localized features, whereas a larger spread yields smoother images. The resolution parameter sets the number of pixels in the persistence image, thus determining the output dimension of the vectorization. Higher resolution captures finer detail but at a higher computational cost. Similar to silhouettes, a weight function (e.g., linear) can be applied to weigh the contribution of each persistence point based on its significance, such as its lifespan.

\subsection{Integrating PDs to ML tasks 2: Using Neural Networks} \label{sec:vecNN}

While vectorization choices provide greater control over the model and data analysis, there are automated methods to avoid dealing with vectorization choices or hyperparameter tuning. These methods optimize the selection for downstream tasks by either finding the best vectorization within a large vectorization space or working directly with persistence diagrams as point clouds, allowing neural networks to handle the data.

\subsubsection{Vectorization with NN: Perslay} \label{sec:perslay} PersLay, introduced by Carrière et al.~\cite{carriere2020perslay}, is an automated vectorization framework for persistence diagrams, which decides the best vectorization method for the downstream task. The method leverages a neural network architecture that processes PDs via a specialized layer called the PersLay layer. This layer incorporates a variety of vectorization strategies, providing a unified and flexible approach to extracting topological features.

The PersLay layer is composed of several key components: 
\begin{itemize}
    \item {\em Weighting Functions:} These assign importance to points in the persistence diagram. Common choices include exponential weighting and Gaussian weighting, which can be learned during training.
\item {\em Transformation Functions:} These functions, such as linear transformations or learned neural networks, apply to the coordinates of the persistence points to encode meaningful geometric and topological information.
\item {\em Symmetric Functions:} After transformation, symmetric functions like sum, mean, or max aggregate the transformed points into a fixed-size vector, ensuring permutation invariance.
\end{itemize}

With this architecture, PersLay covers a wide range of known vectorization methods within its framework, i.e., Persistence Landscapes, Persistence Images, Silhouettes, Betti Curves, Entropy Functions and other Persistence Curves~\cite{chung2019persistence}. By incorporating various vectorization techniques, PersLay can adapt to diverse data types and topological features, making it a robust and versatile tool for integrating PH into ML workflows. For an alternative method involving slightly different generating functions for learnable vectorizations, see~\cite{hofer2019learning} by Hofer et al.

\subsubsection{PD Learning with NN: PointNet} \label{sec:pointnet} An alternative automated method to leverage PDs in ML tasks without hyperparameter tuning is to directly utilize PointNet~\cite{qi2017pointnet} or similar neural network architectures to process and analyze point cloud data, e.g., DeepSets~\cite{zaheer2017deep}, PointMLP~\cite{ma2022rethinking}. Unlike conventional techniques that transform point clouds into regular grids or structured formats, these networks treat PDs as sets of points in $\mathbb{R}^2$. Perslay and PointNet offer fundamentally different strategies for incorporating PDs into ML models without requiring vectorization or hyperparameter optimization. The choice between these methods depends on the specific task, as there is no universally superior approach.

\subsection{Computational Complexity for Persistent Homology} \label{sec:compComplexity}

The computational complexity of persistent homology (PH) is strongly influenced by the choice of filtration complex, as it is directly related to the size, $|\mathcal{K}|$, of the simplicial complex $\mathcal{K}$, i.e., the number of simplices it contains. Early PH algorithms had a cubic complexity with respect to the number of simplices, i.e., $\mathcal{O}(|\mathcal{K}|^3)$~\cite{morozov2005persistence}. Subsequent advancements have reduced this exponent to $w = 2.376$, i.e., $\mathcal{O}(|\mathcal{K}|^w)$~\cite{otter2017roadmap}. 

Thus, the computational challenge boils down to the number of simplices in a given simplicial complex $\mathcal{K}$. For a point cloud with $N$ points, the worst-case scenario for the Rips or \v{C}ech complex involves up to $2^N$ simplices across all dimensions. However, in PH, the computation of $d$-dimensional topological features requires only simplices up to dimension $d+1$. Higher-dimensional simplices do not contribute to the calculation of $H_d(\mathcal{K})$. For instance, when focusing on $0$- and $1$-dimensional features, only simplices up to dimension 2 are needed. 

In a point cloud of size $N$, for a Rips complex, the number of $0$-simplices (vertices) is $N$, the number of possible $1$-simplices (edges) is $\binom{N}{2} = \frac{N(N-1)}{2}$, and the number of possible $2$-simplices (triangles) is $\binom{N}{3} = \frac{N(N-1)(N-2)}{6}$. Recall that if we include all dimensions, the possible number of simplices would be $2^N$.  Hence, increasing the dimension $d$ for topological features thus significantly raises computational costs, which is why most ML applications limit the maximum dimension for simplices in the filtration ($d+1$)  to 2 or 3, which is enough to calculate up to $1$- or $2$-holes, respectively.

The discussion above primarily applies to Rips and \v{C}ech complexes, but PH computations can be significantly more efficient when using cubical complexes, which are commonly employed in image data. For 2D images, the time complexity of PH is approximately $\mathcal{O}(|\mathcal{P}|^r)$, where $r \approx 2.37$ and $|\mathcal{P}|$ represents the total number of pixels~\cite{milosavljevic2011zigzag}. Practically, this implies that the computational cost of PH scales almost quadratically with the image size. For higher-dimensional images, alternative methods for efficient computation exist~\cite{wagner2011efficient}.

In recent years, several works have been published to improve the scalability and computational efficiency of PH. One approach focuses on developing alternative methods for more efficient computation of persistence diagrams~\cite{kannan2019persistent,guillou2023discrete}, while another aims to sparsify datasets while preserving topological information~\cite{otter2017roadmap,malott2020topology,akcora2022reduction}.

\subsection{Software Libraries for Persistent Homology} \label{sec:PH_software}

Several software libraries provide tools for computing persistent diagrams and vectorizations across different types of data structures, particularly point clouds and graphs (see \Cref{tab:libraries}). Notable among these are GUDHI, DIONYSUS, and RIPSER, which have been compared in a benchmarking study~\cite{somasundaram2021benchmarking}.  

Most TDA libraries require point cloud data as input, rather than image or network (graph) data. GUDHI can process images, but only after they are converted into a suitable format, such as a point cloud or a cubical complex. Giotto-TDA extends the capability to graphs through its VietorisRipsPersistence and SparseRipsPersistence for undirected graphs and FlagserPersistence for directed ones. However, a common challenge in graph-based TDA is the reliance on shortest path (geodesic) distances, which may not always capture the most relevant topological features. Giotto-TDA also supports the input of images and time series data, enabling the computation of various vectorizations. Beyond PH computations, the Persim library from Scikit-TDA offers a suite of tools for further analysis of persistence diagrams, including metrics like the Bottleneck distance and visualizations like Persistence Landscapes and Persistence Images, enhancing the interpretability of TDA results.

In terms of usage, most libraries (e.g., TDA in R) were language-specific bindings of the underlying GUDHI, Dionysus, and PHAT libraries, which were originally implemented in Matlab or C++ for efficiency and acted as the earliest and most established workhorses for TDA. In recent years, Python implementations like Giotto-TDA have become increasingly popular, and the widely used ML library Scikit-learn has contributed to this trend by creating a new set of tools called Scikit-TDA for TDA integration ({\small \url{https://github.com/scikit-tda}}). 

It is worth mentioning that since TDA has been developed and extended primarily by mathematicians, the popularity of R within this community has made the TDA R library particularly important. Mathematicians often release advanced models and methods that can be directly connected to TDA research in R first, leading to a delay in their implementation in other programming languages.

\begin{table}[h!]
  \caption{\textbf{Software Libraries for Persistent Homology.} In the data column, P indicates support for Point cloud data, I for Image data, and N for Network (graph) data.}
    \label{tab:libraries}
    \centering
    \resizebox{1.\linewidth}{!}{
    \begin{tabular}{l c l l l}
        \toprule
        \textbf{Library} & \textbf{Data} & \textbf{Language} & \textbf{Notable Feature} & \textbf{Code URL} \\
        \midrule
        \textbf{Giotto-TDA}~\cite{tauzin2020giottotda} & P,N,I & Python & Scikit integration & \small{\url{https://github.com/giotto-ai/giotto-tda}} \\
        \textbf{Gudhi}~\cite{gudhiUrm} & P,I & C++/Python & Well-established & \small{\url{https://github.com/GUDHI/gudhi-devel}} \\
        \textbf{Dionysus2} & P & C++/Python & Persistence types & \small{\url{https://github.com/mrzv/dionysus}} \\
        \textbf{Ripser.py}~\cite{ctralie2018ripser} & P & Python & Integration with Scikit-TDA & \small{\url{https://github.com/scikit-tda/ripser.py}} \\
        \textbf{Ripser}~\cite{Bauer2021Ripser} & P & C++ & Standalone & \small{\url{https://github.com/Ripser/ripser}} \\
        \textbf{JavaPLEX~\cite{Javaplex}} & P & Matlab/Java & Well-established & \small{\url{https://github.com/appliedtopology/javaplex}} \\
        \textbf{TDA} & P & R & R ecosystem integration & \small{\url{https://cran.r-project.org/package=TDA}} \\
        \textbf{PHAT~\cite{bauer2017phat}} & P & Python/C++ & Well-established & \small{\url{https://bitbucket.org/phat-code/phat/src/master/}} \\
        \bottomrule
    \end{tabular}}
\end{table}

%% file: sections/4-MP.tex
Multiparameter Persistence (often referred to as Multipersistence) introduces a novel concept with the potential to significantly enhance the performance of single-parameter persistence. The concept, introduced in the late 2000s by Carlsson, Zomorodian, and Singh~\cite{carlsson2007theory,carlsson2009computing}, has since been actively investigated for real-world applications~\cite{botnan2022introduction}.

\begin{wraptable}{r}{6cm}
\vspace{-.1in}
\caption{Generic Bifiltration \label{tab:bifiltration}}
\resizebox{1.\linewidth}{!}{
\begin{tabular}{ccccccc}
$\K_{m1}$ & $\subset$ & $\K_{m2}$ & $\subset$ & $\dots$ & $\subset$ & $\K_{mn}$ \\
$\cup$ &  & $\cup$ &  & $\cup$ &  & $\cup$\\
$\dots$ & $\subset$ & $\dots$ & $\subset$ & $\dots$ & $\subset$ & $\dots$\\
$\cup$ &  & $\cup$ &  & $\cup$ &  & $\cup$\\
$\K_{21}$ & $\subset$ & $\K_{22}$ & $\subset$ & $\dots$ & $\subset$ & $\K_{2n}$ \\
$\cup$ &  & $\cup$ &  & $\cup$ &  & $\cup$\\
$\K_{11}$ & $\subset$ & $\K_{12}$ & $\subset$ & $\dots$ & $\subset$ & $\K_{1n}$ \\
\end{tabular}}
\vspace{-.1in}
\end{wraptable}

For original PH, the term \textit{single persistence} is applied because we filter the data in just one direction, $\K_1\subset \K_2\subset\dots\subset\K_n$. The filtration's construction is pivotal in achieving a detailed analysis of the data to uncover hidden shape patterns. For example, for graph setting, when utilizing a single function $f:\V\to\R$ containing crucial domain information (e.g., value for blockchain networks, atomic number for protein networks), it induces a single-parameter filtration as described earlier. However, numerous datasets offer multiple highly relevant domain functions for data analysis. Similarly, in other data types, there are several ways to expand the single filtration into multifiltration to get finer information on the topological patterns hidden in the data. The idea of using multiple filtration functions simultaneously to obtain a finer decomposition of the dataset led to the suggestion of the MP theory as a natural extension of single persistence (SP).

\subsection{Multifiltrations} 

Multifiltration refers to filtrations constructed using multiple scale parameters. A typical example of a bifiltration, denoted as $\{\K_{ij}\}$, is illustrated in Table~\ref{tab:bifiltration}, where each row and column corresponds to a distinct filtration. This bifiltration is generated by simultaneously varying one scale parameter along the horizontal axis and another along the vertical axis. The specific construction of these filtrations depends on the data type, with additional parameters introduced based on the nature of the data. Below, we summarize common methods for different data types. While the explanation focuses on two parameters for simplicity, this approach generalizes to any number of parameters.

\begin{figure}[b]
    \centering
    \includegraphics[width=.9\textwidth]{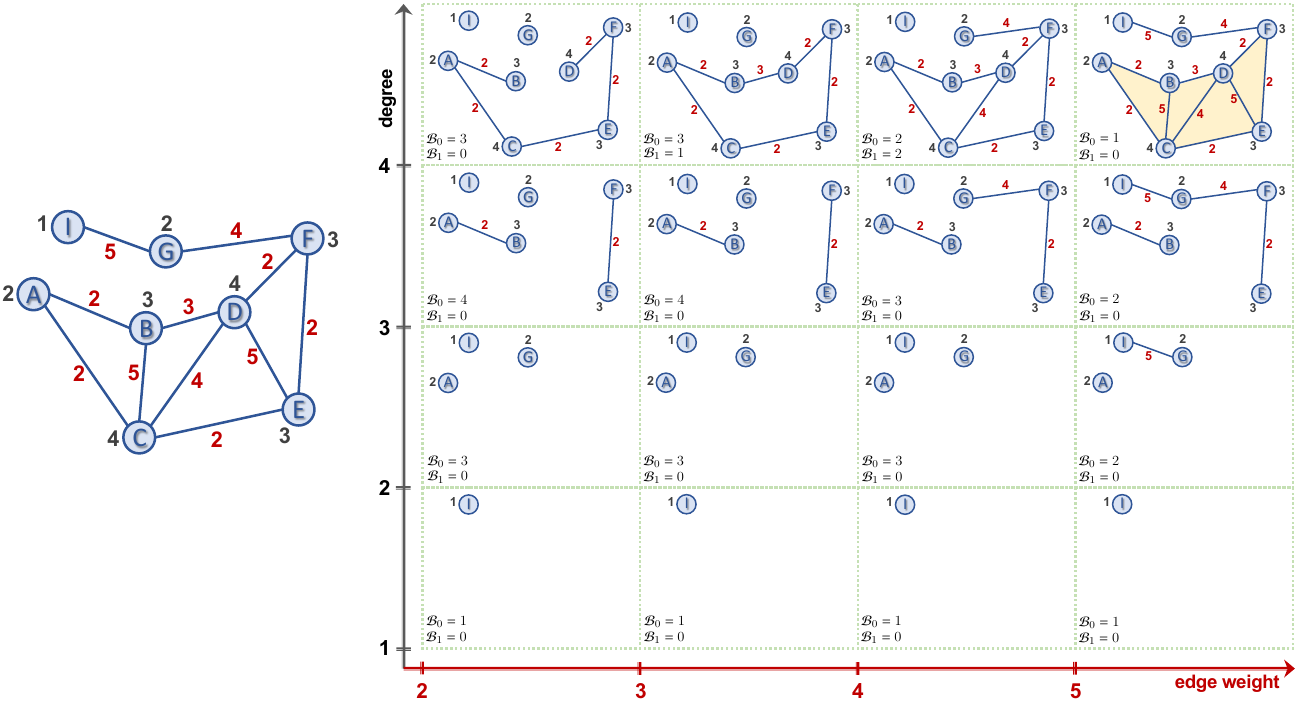}
    \caption{\footnotesize Multidimensional persistence on a graph network (original graph: left). Black numbers denote the degree values of each node whilst red numbers show the edge weights of the network. Hence, shape properties are computed on two filtration functions (i.e., degree and edge weight). While each row filters by degree, each column filters the corresponding subgraph using its edge weights. For each cell, the lower left corners represent the corresponding threshold values. For each cell, $\mathcal{B}_{0}$ and $\mathcal{B}_{1}$ represent the corresponding Betti numbers. The figure is adapted from~\cite{chen2024emp}. \label{Fig:ToyMP}}
\end{figure}
 
\subsubsection{Multifiltrations for Graphs.} \label{sec:MP-graph}

In graph settings, a single filtration function results in a single-parameter filtration $\wh{\G}_1 \subset \dots \subset \wh{\G}_N = \wh{\G}$. However, employing multiple filtration functions enables a more detailed analysis of the data. For instance, two node functions $f: \V \to \R$ and $g: \V \to \R$, which provide complementary information about the network, can be combined using Multiparameter Persistence to generate a unique topological fingerprint. These functions induce a multivariate filtration function $F: \V \to \R^2$ defined by $F(v) = (f(v), g(v))$.

Next, we define sets of increasing thresholds $\{\alpha_i\}_{i=1}^m$ for $f$ and $\{\beta_j\}_{j=1}^n$ for $g$. Then, we have $\V_{ij} = \{v_r \in \V \mid f(v_r) \leq \alpha_i, g(v_r) \leq \beta_j\}$, which can be written as $\V_{ij} = F^{-1}((-\infty, \alpha_i] \times (-\infty, \beta_j])$. Let $\G_{ij}$ be the subgraph of $\G$ induced by $\V_{ij}$, meaning the smallest subgraph of $\G$ generated by $\V_{ij}$. This setup induces a bifiltration of complexes $\{\wh{\G}_{ij} \mid 1 \leq i \leq m, 1 \leq j \leq n\}$, which can be visualized as a rectangular grid of size $m \times n$ (see \cref{Fig:ToyMP}). For more details on applying multipersistence in graph settings, refer to \cite{chen2024emp, loiseaux2024stable}. Additionally, one can combine power filtration with filtration by functions by applying power filtration to each subgraph in the sequence induced by sublevel filtration via functions~\cite{demir2022todd}. In \Cref{sec:drug}, we give details of the utilization of MP method on graph setting for computer-aided drug discovery.

\subsubsection{Multifiltrations for Point Clouds.} \label{sec:MP-point_cloud}

In the context of point clouds, a natural parameter to consider is the radius $r$, as discussed in~\Cref{sec:point_cloud}. However, a point cloud $\p$ often has regions of varying density, with some areas being very dense and others quite sparse. Additionally, a few outliers can significantly distort the topological signature due to the way it is constructed. To address this issue, it is common to use a density parameter in conjunction with the radius parameter when constructing the filtration. This parameter reflects the local density of points in the point cloud and helps identify regions of different densities, distinguishing features significant in dense regions from those in sparse ones.
To achieve this, we first need to compute local densities. For each point $p \in \p$, we compute a density measure $\delta(p)$, which could be based on the number of points within a fixed radius $r_d$ or kernel density estimation~\cite{vipond2020multiparameter,botnan2022introduction}.

Next, we define our threshold set $\{(d_i, r_j)\}$, where $d_1 > d_2 > \dots > d_m$ corresponds to thresholds for density, and $0 = r_1 < r_2 < \dots < r_n = \text{diam}(\p)$ are radius thresholds. We then define the multifiltration as follows. Using $\delta(p)$, we define a nested sequence of subsets $\p_1 \subset \p_2 \subset \dots \subset \p_m = \p$, where $\p_i = \{p \in \p \mid \delta(p) \geq d_i\}$. For each $i_0$, we treat $\p_{i_0}$ as a separate point cloud and apply the Rips filtration process to it. Specifically, for each $i_0$, we construct the Rips filtration $\VR_{i_01} \subset \VR_{i_02} \subset \dots \subset \VR_{i_0n}$. This gives a bifiltration $\{\K_{ij}\}$ as in \Cref{tab:bifiltration}, with $\K_{ij} = \VR_{ij}$ representing the Rips filtration for point cloud $\p_i$ with radius parameter $r_j$. e.g., the first column $\K_{i1} = \p_i$ corresponds to radius $r_1 = 0$.

This multifiltration approach provides more detailed information by capturing topological patterns in both dense and sparse regions, offering a richer understanding of the data's underlying shape, especially when varying densities and scales are crucial for the analysis. For further details, see~\cite{loiseaux2024framework,vipond2020multiparameter}.

\subsubsection{Multifiltrations for Images.} \label{sec:MP-image}

Similarly, in the image setting, if the image is a color image, one can easily employ all color channels at the same time, similar to the graph setting. In particular, a color image has three color functions, denoted as $R$ (red), $G$ (green), and $B$ (blue). Thus, for every pixel $\Delta_{ij}$, there exist corresponding color values: $R_{ij},G_{ij},B_{ij}\in [0,255]$. To proceed, we establish a three-parameter multifiltration with parameters $\{\alpha_m\}_1^{N_R}$, $\{\beta_n\}_1^{N_G}$, $\{\gamma_r\}_1^{N_B}$, where $\alpha_m,\beta_n,\gamma_r\in [0,255]$ are threshold values for color channels $R,G$, and $B$ respectively. By simply defining binary images $\X_{m,n,r}=\{\Delta_{ij}\subset \X \mid  R_{ij}\leq \alpha_m, G_{ij}\leq \beta_n, B_{ij}\leq \gamma_r\}$, we obtain a three parameter multifiltration of size $N_R \times N_G \times N_B$.
Similarly, for grayscale images, one can utilize height, radial, erosion, signed distance, or similar filtration methods~\cite{garin2019topological} along with the color filtration to obtain a multifiltration for color images.

\subsection{MP Vectorization Methods} \label{sec:MP-vec}

By computing the homology groups of the complexes in these multifiltrations, $\{\h_k(\K_{ij})\}$, along with the induced inclusion maps, we obtain the corresponding \textit{multipersistence module}, which can be visualized as a rectangular grid of size $m\times n$. The goal here is to track $k$-dimensional topological features through the homology groups $\{\h_k(\K_{ij})\}$ within this grid. However, as explained in~\cite{botnan2022introduction}, technical challenges rooted in commutative algebra prevent us from transforming the multipersistence module into a mathematical structure like a "Multipersistence Diagram." The key issue is that for any $k$-dimensional hole in the multifiltration, we cannot directly assign a birth and death time due to the partial ordering within the module. For example, if the same $k$-hole $\sigma$ appears at $\K_{1,4}$ and $\K_{2,3}$, neither $(1,4) \nprec (2,3)$ nor $(2,3) \nprec (1,4)$, making it difficult to define a birth or death time for $\sigma$ unless it exists in a perfectly rectangular region. This issue does not arise in single persistence, where any two indices are always comparable, i.e., $i<j$ or $j<i$.

As a result, we do not have an effective representation of the MP module~\cite{botnan2022introduction}. While these technical obstacles prevent this promising approach from reaching its full potential in real-life applications, in the past years, several approaches have been proposed to extract key insights from MP modules by skipping MP representations and directly going to the vectorization from multifiltrations, which we detail below.

The simplest method to derive a vector (or tensor) from a multifiltration involves using the \textit{Hilbert function} of the MP module, which is a generalization of Betti functions. Let $\{\K_{ij}\}$ denote an $m \times n$ multifiltration associated with a given dataset $\X$. First, we construct the multipersistence module ${\h_k(\K_{ij})}$. Next, by computing the Betti numbers of each simplicial complex in the multifiltration, we obtain an $m \times n$ matrix (or tensor) $\wh{\beta}_k(\X) = [\beta_k(\K{ij})]$, i.e., $\beta_k(\K_{ij}) = \text{rank}(\h_k(\K_{ij}))$. These are also called \textit{multipersistence (or bigraded) Betti numbers}. In graph representation learning, this simple vectorization of the MP module has demonstrated remarkable performance~\cite{loiseaux2024stable,chen2024emp,xin2023gril}.

While bigraded Betti numbers offer a straightforward vectorization method for multiparameter (MP) modules, they fail to capture crucial topological information, especially regarding the significance of dominant features or the presence of a few important topological features. More sophisticated techniques are necessary for effective MP vectorization. One widely used strategy involves the "slicing technique," which focuses on studying one-dimensional fibers within the multiparameter domain. A clearer understanding can be obtained by restricting the multidimensional persistence module to these single directions (slices) and applying single persistence analysis. 

In their work \cite{carriere2020multiparameter}, Carriere et al. explored this approach by considering multiple such slices, often referred to as {\em vineyards}, and summarizing the resulting persistence diagrams. Another significant advancement is found in {\em multipersistence landscapes} \cite{vipond2020multiparameter} by Vipond, which extends the concept of persistence landscapes \cite{bubenik2015statistical} from single to multiparameter persistence. 

The vectorization of MP modules is a rapidly evolving area of research. Various techniques have recently been proposed in practical applications, e.g.,  Hilbert decomposition signed measure (MP-HSM-C)~\cite{loiseaux2024stable}, Effective MultiPersistence (EMP)~\cite{chen2024emp}, Generalized Rank Invariant Landscape (GRIL)~\cite{xin2023gril}, and Stable Candidate Decomposition Representations (S-CDR)~\cite{loiseaux2024framework}. Although many recent methods are burdened by high computational costs, the MP-HSM-C method introduced by Loiseaux et al.~\cite{loiseaux2024stable} stands out as the most efficient in terms of speed and performance among current MP vectorization techniques.

%% file: sections/5-Mapper.tex
Next to Persistent Homology, another powerful method in TDA is the \textit{Mapper}, which Singh, Memoli, and Carlsson introduced in the late 2000s~\cite{singh2007topological, Carlsson:2009}.  The Mapper stands out as an effective and versatile tool for extracting insights from high-dimensional datasets, which creates a \textit{simplified graph representation} of data by combining ideas from algebraic topology and data visualization. \textit{While PH is mostly used in supervised learning, Mapper is commonly utilized in unsupervised settings}. In the past decade, it has made critical contributions to various fields involving point clouds in high dimensional spaces, e.g., biomedicine~\cite{skaf2022topological}. This method works by projecting data onto a lower-dimensional space, clustering the points within overlapping intervals, and constructing a topological network that captures the essential structure of the dataset. Through this approach, Mapper can reveal hidden patterns, relationships, and the underlying shape of data, making it particularly valuable for unsupervised learning in fields such as genomics, neuroscience, and social network analysis. Therefore, it is also considered a smart dimension reduction technique, too~(\Cref{fig:mapper}).

\subsection{Mapper for Point Clouds} \label{sec:mapper-PC}
For a point cloud $\X$ in $\R^N$ and a real-valued function $f: \X \to \R$, the Mapper algorithm provides a summary of the data by scanning the clusters in $\X$ in the direction dictated by the function $f$, which is commonly referred to as a \textit{filter function} or \textit{lens}~\cite{dey2022computational}. The output of the Mapper algorithm is the Mapper graph, which is considered a meaningful summary of the data, representing clusters and relations between the clusters in the data. It has been applied in several contexts like diabetes subtype identification~\cite{li2015identification}, ransomware payment detection on blockchains~\cite{bitcoinheist2020}, and cancer genotyping from single-cell RNA sequence data (\Cref{sec:mapper-app}).

The Mapper algorithm generates a graph-based summary of a high-dimensional point cloud. In this new Mapper graph, nodes represent clusters within the {original} point cloud, and edges indicate the interaction between these clusters. In other words, Mapper is a soft clustering approach where a data point may appear in multiple clusters; these clusters are then represented as nodes in the new Mapper graph.

Given a point cloud $\X$ in $\R^N$, we first define a lens $f:\X\to\R$. In general, the lens function can be a natural function derived from the data domain, or it may be computed using a dimensionality reduction technique. The proper selection of hyperparameters for Mapper is crucial, and we will shortly discuss these in \Cref{sec:mapperParams}.

\begin{wrapfigure}{r}{3in}
\vspace{-.2in}
\centering
    \includegraphics[width=\linewidth]{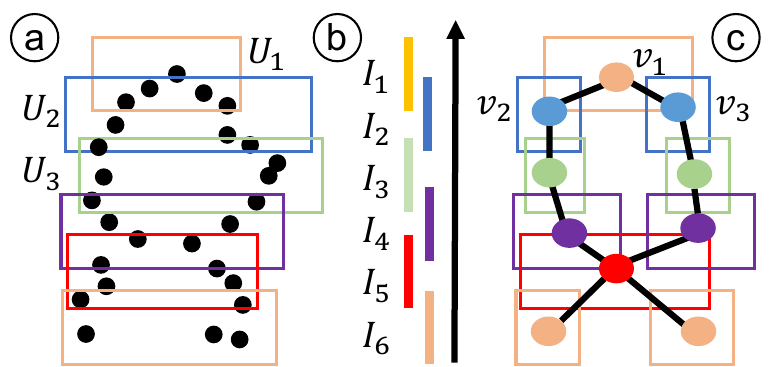}
    \caption{
      {\bf Toy Example of Mapper.} For a point cloud $\X$ (a), we define a lens function \textit{$f:\X\to\R$} (b), and the induced covering defines a Mapper graph where nodes represent clusters and edges represent related clusters (c). \label{fig:toyMapper}    }
\vspace{-.2in}
\end{wrapfigure}

The lens function then decomposes the data into subregions through a cover of the image $f(\X)\subset \R$, allowing for the identification of clusters within each subregion.  
Formally, we cover the image $f(\X)\subset \R$ with a collection of open intervals, i.e. $\I = \{I_k\}_{k=1}^{n}$ where $\bigcup_k I_k\supset f(\X)$. The intervals $I_k=(a_k,b_k)$ are indexed such that $a_k < a_{k+1}$ and typically $b_k \geq a_{k+1}$, allowing for overlap between consecutive intervals.  Figure~\ref{fig:toyMapper}b shows 6 overlapping intervals $\I=\{I_1,\dots, I_6\}$ for a toy example. 

Next, we consider the preimage of each interval, $U_k=f^{-1}(I_k)\subset \X$. In particular, the preimage of each interval refers to the set of points in the original high-dimensional space, such as the data points, mapped into a given interval by the lens function. For example,  in Figure~\ref{fig:toyMapper}, the union of all points from $U_2$ and $U_3$ and lower-part points from $U_1$ are the preimage of the interval $I_2$. We apply a clustering algorithm (e.g., k-means, DBSCAN) to each preimage $U_{k_0}$, dividing it into $m_{k_0}$ clusters $\{C_{k_0l}\}_{l=1}^{m_{k_0}}$. 

For each cluster $C_{kl}$, we create a node $v_{kl}$ in the Mapper graph and let $\V_k=\{v_{k1}, v_{k2}, \dots, v_{km_k}\}$ represent the clusters in $U_k\subset \X$. Thus, $\V=\bigcup_k \V_k$ forms the node set for the Mapper graph $\G$.  In \Cref{fig:toyMapper}, there are 10 nodes in the graph for the 6 preimages: $I_2$, $I_3$, $I_4$, and $I_6$, each resulting in two clusters. Note that the number of clusters within a preimage can vary. For example, representing $U_1$ with a single cluster might result from using a density-based clustering algorithm (e.g., DBSCAN), which identifies dense regions as clusters while treating sparse regions as noise, thereby avoiding the creation of clusters in those less populated areas.

After obtaining the nodes, we define the edges based on the intersections of the clusters. Note that as $I_k\cap I_{k+1}\neq \emptyset$, clusters in subsequent levels might have nontrivial intersections. Then, if the clusters $C_i\cap C_j\neq \emptyset$, we add an edge between the corresponding nodes $v_i$ and $v_j$. i.e., $\E=\{e_{ij}\mid C_i\cap C_j\neq \emptyset\}$ where $e_{ij}$ represents the edge between the nodes $v_i$ and $v_j$. It is also possible to define a weighted graph with weights $\omega_{ij}=\#(C_i\cap C_j)$, the count of the points in the intersection. In this case, the weights reflect the strength of the interaction between the corresponding clusters, indicating how many data points are included in the overlapped area. The final Mapper graph gives a rough summary/sketch of the whole point cloud. 

The Mapper graph not only helps in identifying data clusters but also enables the selection of data points that are similar to a given subset by leveraging its structure~\cite{bayir2022topological}. In this context, nodes that are closer together on the Mapper graph tend to contain more similar data points. This aspect of cluster similarity based on their proximity in the Mapper graph is an area that remains largely understudied and offers significant potential for further exploration. The Mapper graph computations have been detailed in \Cref{alg:mapper}. In~\Cref{sec:mapper-app}, we outline the utilization of Mapper in a real-life application, i.e., cancer genotyping from RNA sequencing.

\paragraph*{Common Filter Functions for Mapper.} While we explain the Mapper algorithm for single-valued filter functions to keep the exposition simpler, in practice it is very common to use multivariate filter functions $F:\X\to \R^2$. One of the most common filter functions used in applications is the \textit{Stochastic Neighborhood Embedding (t-SNE)}~\cite{hinton2002stochastic}, and a variant of this, called \textit{Neighborhood Lens}~\cite{gabrielsson2019exposition}. 

Similarly, it is common to use other dimension reduction techniques as filter map $F:\X\to \R^2$, e.g., Isomap, PCA, or UMAP~\cite{rather2022umap}. Then, for an open covering $\I=\{I_{ij}\}$ of $F(\X)$ in $\R^2$, one can repeat the process of $U_{ij}=F^{-1}(I_{ij})$, and assign a node $w$ for each cluster in $U_{ij}$. Again, assign an edge between the nodes when the corresponding clusters have nontrivial intersections. Note that if the domain of the data offers good filter functions $f,g:\X\to\R$,  one can also utilize multivariate filter functions $F(x)=(f(x),g(x))\in\R^2$ by combining both functions in the process.

\subsubsection{Hyperparameters for Mapper} \label{sec:mapperParams}
For a point cloud $\X$, four main parameter choices must be made to obtain its Mapper graph. 


\begin{enumerate}
    \item \textit{The lens function.} If available, the lens function can be derived directly from the data domain, which in turn greatly enhances the interpretability of the model. In practice, when appropriate lens functions are not readily available from the data domain, dimensionality reduction techniques like t-SNE, UMAP, or PCA are commonly used to create lenses \( f: \mathcal{X} \to \mathbb{R}^2 \). 
    \item \textit{Clustering method.} For each subset $f^{-1}(I_k)$, the clustering method determines the nodes of the Mapper graph. The most common methods employed are DBSCAN and k-means, with their hyperparameters controlling the granularity of the resulting clusters.
    For k-means, it is common to use $<10$ for clusters (e.g., see~\cite[Figure 10]{shamsi2024graphpulse}).
    
    \item {\em Resolution and Overlap.} The resolution and overlap are the parameters for the covering $\{I_k\}_1^n$ of $f(\X)\subset \R$, i.e., $\bigcup_{k=1}^n I_k\supset f(\X)$. 
    The {\em resolution}, $n$, is the number of bins (intervals) $\{I_k\}$ to cover $f(\X)$. The {\em overlap} is the percentage of overlaps of these intervals, i.e. $\frac{|I_k\cap I_{k+1}|}{|I_k|}$.  Note that increasing the resolution gives a finer summary by increasing the number of nodes in the Mapper network and making the clusters smaller (See~\Cref{fig:mapper-bin}). On the other hand, increasing overlap adds more relation (edges) between the nodes (clusters).  In \Cref{fig:toyMapper}, the resolution is 6, and the overlap is the fixed intersection amount (e.g., 20\%) between the intervals $I_k$ and $I_{k+1}$. Typically, it is common to use $10\%-30\%$ overlap and 10-20 intervals (e.g., Figure 10 of GraphPulse~\cite{shamsi2024graphpulse}).
\end{enumerate}

\begin{algorithm}
\caption{Mapper Algorithm}
\label{alg:mapper}
\begin{algorithmic}[1]
\State \textbf{Input:} DataSet $\X$, Filter function $f: \X \to \mathbb{R}$, $\I = \{I_k\}$ covering of $f(\X)$ for a given resolution and overlap.
 
\State \textbf{Output:} Graph $\G$ (nodes and edges)
\State Initialize empty graph $\G$
\State Compute filter values: $f(\X)$
\For{each covering set $I_k$ in $\I$} \Comment{Defining nodes of $\G$}
    \State Compute preimage: $U_k = \{x \in \X \mid F(x) \in I_k\}$
    \State Cluster $U_k$ into clusters $\{C_i\}$
    \State Assign a node for each cluster, and get a set of nodes $\V_k$
    \State Add nodes in $\V_k$ to graph $\G$
\EndFor
\For{each pair of nodes $(v_i, v_j)$ in $\G$} \Comment{Defining edges of $\G$}
    \If{Clusters $C_i$ and $C_j$ have common points}
        \State Add edge $(v_i, v_j)$ to $\G$
    \EndIf
\EndFor
\State \textbf{Return} Graph $\G$
\end{algorithmic}
\end{algorithm}

\subsection{Mapper for Other Data Formats}  \label{sec:mapper-graph}

While the Mapper approach is most commonly applied to point clouds, it can be adapted for other data formats, such as graphs and images. For graphs, this adaptation can be seen as a form of graph coarsening/skeletonization, where the goal is to summarize clusters of nodes within the graph.

In~\cite{bodnar2021deep,hajij2018mog}, Bodnar et al. present a natural extension of the Mapper algorithm to the graph setting. Given a graph $\G=(\V,\E)$, we start with a filter function $f:\V\to\R$ and define a cover $\I = \{I_k\}_{k=1}^{n}$, where $\bigcup_k I_k \supset f(\X)$. The set $\V_k = f^{-1}(I_k)$ represents a subset of nodes, and $\G_k$ is the induced subgraph, i.e., $\G_k = (\V_k,\E_k)$ where $\E_k \subset \E$ are the edges connecting pairs of vertices in $\V_k$. Instead of clustering, we directly use the components (connected subgraphs) in $\G_k$ as the nodes of the Mapper graph.
Specifically, if $\G_{k_0} = \bigcup_{l=1}^{m_{k_0}} \G_{kl}$ has $m_{k_0}$ connected subgraphs, we define the node set $\C_k = \{c_{k_01}, \dots, c_{k_0m_{k_0}}\}$. For example, if $\G_k$ has five connected subgraphs, we define five nodes in the Mapper graph, each representing one connected subgraph in $\G_k$. Let $\W = \{w_1, w_2, \dots, w_M\}$ be the collection of all such nodes where $M = \sum_{k=1}^{n} m_k$. This defines the node set of the Mapper graph of $\G$. 

The edge set is defined similarly: If the subgraphs $\G_{kl}$ and $\G_{(k+1) l'}$ share a common node, then an edge is added between the corresponding nodes in $\W$. In summary, in the original construction, we replace the point cloud with the node set of $\G$, and the clustering algorithm is directly derived from the graph structure.  For this approach, one can utilize the common filtering functions $f:\V\to\R$ from persistent homology (e.g., degree, closeness, betweenness, eccentricity) or can use other popular functions from graph representation learning, e.g., eigenfunctions of graph Laplacian, pagerank~\cite{hajij2018mog}.

Alternatively, if the graph \(\G = (\V, \E, \X)\) includes node attributes, a Mapper graph can be directly derived from the set of node attributes \(\X\). In particular, we treat the node attribute vectors \(\X = \{\X_i\}_1^n\) with \(\X_i \in \R^N\) as a point cloud \(\X \subset \R^N\) and apply the Mapper algorithm as before. This Mapper graph provides a visual summary of the node attribute space, independent of the graph's structure. In other words, the resulting Mapper graphs summarize the graph based on node attributes alone, without incorporating information about node neighborhoods in the original graph. An effective utilization of this approach can be found in~\cite{shamsi2024graphpulse}.

\begin{figure}[h!]
\centering
    \includegraphics[width=\textwidth]{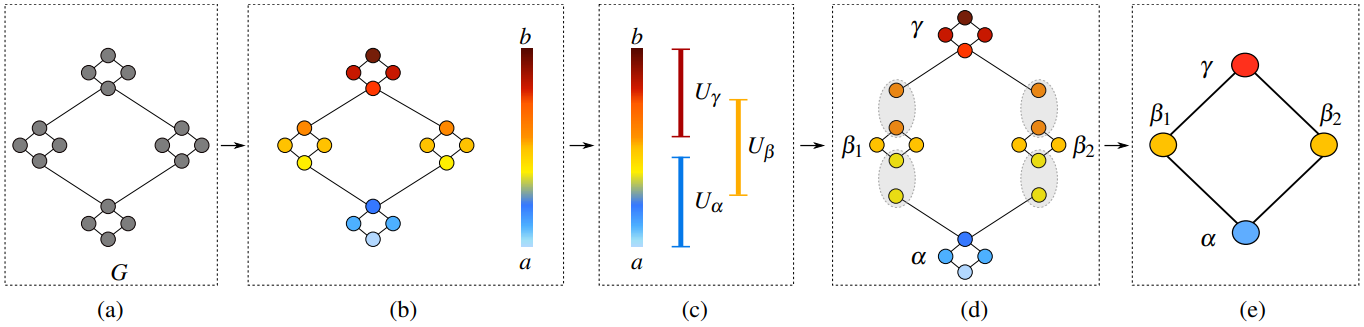}
    \caption{  {\bf Mapper for Graphs.} For a graph $\G$ (a), a filter function $f: \V \to \R$ (b), and the covering $\{U_\alpha, U_\beta, U_\gamma\}$ (c), we obtain the node clusters (connected subgraphs) (d), which define a Mapper network (e). The figure is adapted from~\cite{hajij2018mog}.\label{fig:mapper-graph}}
\end{figure}

The Mapper algorithm on images is analogous to its application in graphs, where clustering is derived from the original image~\cite{robles2017shape}. For a given \(r \times s\) image \(\X\), let \(f: \X \to \R\) represent the color values of the pixels. We define a cover \(\I = \{I_k\}_{k=1}^{n}\) such that \(\bigcup_k I_k \supset f(\X)\). A node is defined for each connected component in \(f^{-1}(\I_k)\), and an edge is created between nodes if the corresponding components have a nontrivial intersection. This Mapper graph summarizes the interaction between different color regions in \(\X\) as a graph.

\subsection{Computational Complexity of Mapper}
The computational complexity of TDA Mapper varies depending on the specific choices of filter functions, clustering algorithms, and dimensionality reduction techniques.

The first step in TDA Mapper involves applying a filter function. If t-SNE is used as the filter, the computational complexity of the Barnes-Hut t-SNE variant is $\mathcal{O}(N \log N \cdot D)$~\cite{van2014accelerating}, where $N$ is the number of data points (e.g., for graphs, the number of nodes $|\mathcal{V}|$) and $D$ is the dimensionality of the data (e.g., for graphs, the number of node attributes). The complexity of t-SNE is largely driven by the computation of pairwise distances in a $D$-dimensional space, followed by optimizing the low-dimensional embedding. While dimensionality $D$ influences the time required for these computations, the overall complexity is typically dominated by the number of data points $N$.

After t-SNE, the data is covered by overlapping intervals, and clustering is performed within each interval. The complexity of this step depends on the clustering algorithm used. For example, if a clustering algorithm with complexity $\mathcal{O}(N^2)$ is used, this step may add significant computational cost. Finally, the Mapper complex is constructed by connecting clusters, which typically has a lower complexity, often $\mathcal{O}(N)$ to $\mathcal{O}(N \log N)$, depending on the number of clusters and the method used to connect them.

The overall computational complexity of TDA Mapper when using t-SNE is dominated by the complexity of t-SNE, which is $\mathcal{O}(N \log N \cdot D)$.  The subsequent steps in the Mapper pipeline add to this complexity, particularly the clustering step. When using $k$-means clustering, the complexity is typically $\mathcal{O}(k \cdot N \cdot D \cdot T)$, where $k$ is the number of clusters, $N$ is the number of data points, $D$ is the dimensionality of the data, and $T$ is the number of iterations. Therefore, the overall complexity can be expressed as $\mathcal{O}(N \log N \cdot D) + \mathcal{O}(k \cdot N \cdot D \cdot T)=\mathcal{O}(N \log N)$. 

\subsection{Software Libraries for Mapper} \label{sec:mapper-software}

Several software libraries provide tools for constructing and analyzing Mapper (see \Cref{tab:mapper-libraries}). Among these, Kepler-Mapper is a popular choice for Python users, particularly within the Scikit-TDA ecosystem, offering HTML outputs that are ideal for creating shareable, interactive visualizations. This makes it a practical tool for users who need to present their findings to non-technical users. 

Giotto-TDA integrates well with the Python environment and offers parameter visualization capabilities, making it an excellent choice for those needing to iteratively refine their Mapper construction. Its interactive features are especially helpful for real-time exploration of the impact of different filter functions and covering parameters.

For users prioritizing performance over interactivity, tda-mapper-python offers a faster computation engine, making it well-suited for large datasets or scenarios where rapid iteration is necessary. However, the lack of interactive features means that users will need to rely on external tools for visual exploration.

TDAmapper in R is an excellent choice for users already embedded in the R ecosystem. It’s a robust standalone solution that is particularly useful for those who prefer the extensive statistical and data manipulation capabilities available in R. However, the lack of interactivity might require additional effort for visualization and exploration.

Finally, Mapper Interactive is designed for those who value hands-on engagement with the data. It offers a highly interactive Python-based environment, making it ideal for exploratory data analysis where immediate feedback and manipulation of the Mapper graph are crucial.

In practice, the choice of library typically depends on the specific requirements of your project, such as whether you prioritize interactivity, computational speed, or seamless integration with other tools in your data analysis workflow. For example, Kepler-Mapper and Giotto-TDA are ideal for exploratory analysis and visualization, while tda-mapper-python and TDAmapper are more suitable for handling large-scale computations and integrating with ecosystems like Python and R, respectively.

\begin{table}[t]
\caption{Summary of TDA Mapper Libraries. \label{tab:mapper-libraries}}
\centering
\resizebox{1.\linewidth}{!}{
\begin{tabular}{l c  l c l}
\toprule
\textbf{Library Name}      & \textbf{Lang.}  & \textbf{Notable Feature}     &     \textbf{Interactive}     & \textbf{Code URL}                                 \\  \midrule 
\textbf{Kepler-Mapper~\cite{KeplerMapper_JOSS}}      & Python              & HTML output  &  \checkmark  & \small{\url{https://github.com/scikit-tda/kepler-mapper} (Scikit-TDA)}        \\  
\textbf{Giotto-tda~\cite{giottotda}}        & Python           & Parameter visuals    & \checkmark & \small{\url{https://github.com/giotto-ai/giotto-tda} }       \\   
\textbf{tda-mapper-python~\cite{simiMapper}} & Python             & Fast computation                          &     $\times$        & \small{\url{https://github.com/lucasimi/tda-mapper-python}} \\  
\textbf{TDAmapper}~\cite{singh2007topological}        & R                           & Standalone                          &           $\times$ & \small{\url{https://github.com/paultpearson/TDAmapper}}     \\  
\textbf{Mapper Interactive~\cite{zhou2021mapper}}         & Python                           & Interactive                          &           $\checkmark$ & \small{\url{https://github.com/MapperInteractive/MapperInteractive}}     \\  
\bottomrule
\end{tabular}}
\end{table}

%% file: sections/6-Applications.tex
In this section, we will provide examples of the applications of TDA methods in machine learning. We provide four examples from published papers and outline their model and how they applied TDA in their project.

\subsection{PH for Point Clouds: Shape Recognition} \label{sec:shape}
\begin{wrapfigure}{r}{3in}
\vspace{-.2in}
\centering
    \includegraphics[width=\linewidth]{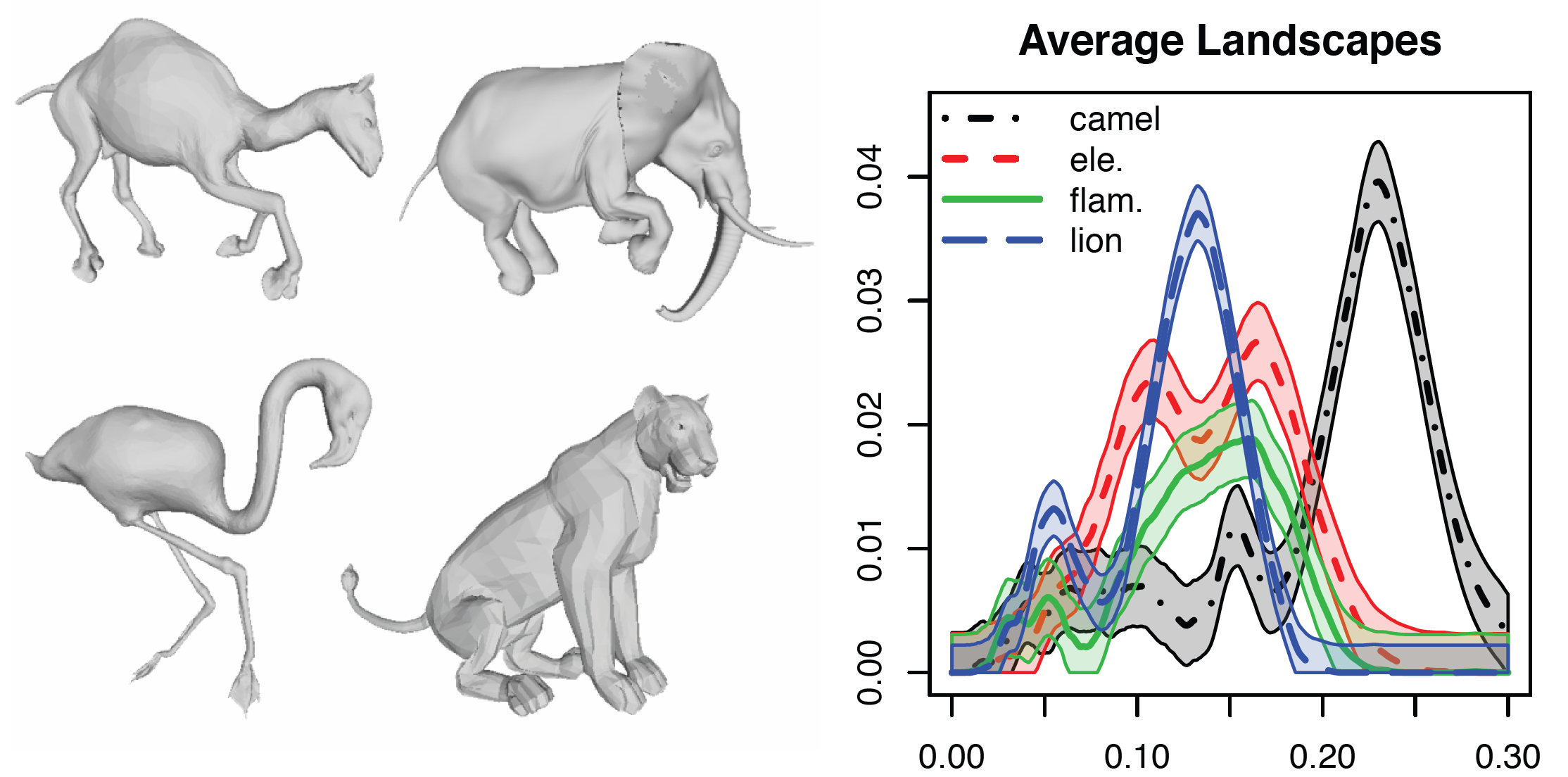}
    \caption{\footnotesize  {\bf Shape Recognition via PH.} In \cite{chazal2015subsampling}, the authors obtained 100 point clouds for each animal by subsampling the surfaces above and analyzed their persistence landscapes. On the right, they present the 95\% confidence bands for the persistence landscapes of these 400 point clouds, demonstrating that each animal's persistence landscape exhibits significantly distinct characteristics. The figure is adapted from~\cite{chazal2015subsampling}. \label{fig:shape}    }
    \vspace{-.2in}
\end{wrapfigure}
In this section, we present an illustrative example of the utilization of PH in shape recognition, based on the work~\cite{chazal2015subsampling}, where Chazal et al. demonstrated the effectiveness of PH in shape recognition. One specific example involved four different animals: an elephant, flamingo, lion, and camel, each representing a unique shape in $\R^3$, where each shape is normalized to have a diameter of one. They began by selecting 500 random points from the surface of each animal, as shown in~\Cref{fig:shape}, resulting in the point clouds $\X_E$, $\X_L$, $\X_F$, and $\X_C$ for the elephant, lion, flamingo, and camel, respectively. Then, they created 100 subsamples of 300 points each from these point clouds, resulting in 400 different point clouds: $E_i \subset \mathcal{X}_E$, $L_i \subset \mathcal{X}_L$, $F_i \subset \mathcal{X}_F$, and $C_i \subset \mathcal{X}_C$ for $1 \leq i \leq 100$.

Next, they performed Rips filtration on each point cloud $\{E_i,L_i,F_i,C_i\}_{i=1}^{100}$ (see \Cref{sec:point_cloud}) and computed the corresponding persistence diagrams for dimension one. By vectorizing these persistence diagrams, they obtained the persistence landscapes for each point cloud, resulting in 400 persistence landscape functions: 
$\{\lambda(E_i), \lambda(L_i), \lambda(F_i),  \lambda(C_i)\}_{i=1}^{100}$.
In~\Cref{fig:shape}, they present the 95\% confidence bands for the true average landscapes (bold curves) of each class. e.g., blue confidence band correspond to 100  persistence landscapes coming from lion point clouds $\{\lambda(L_1), \dots, \lambda(L_{100})\}$. The narrowness of these confidence bands indicates that PH techniques provide robust shape-embedding methods that are minimally affected by noise. Moreover, the distinct confidence bands for each class underscore the effectiveness of PH in shape recognition. We also note that recent work by T\"urke\c{s} et al.~\cite{turkes2022effectiveness} studied the effectiveness of PH for shape recognition in different settings.

\subsection{PH for Graphs: Crypto-Token Anomaly Forecasting} \label{sec:crypto}

Three primary types of problems dominate graph representation learning: graph classification, node classification, and link prediction, as outlined by Hamilton et al.~\cite{hamilton2017representation}. These tasks encompass a broad range of real-life applications, including brain connectivity networks, molecular property prediction, recommender systems, fraud detection and transaction networks. In a related study, Li et al.~\cite{li2020dissecting} employ persistent homology to extract effective feature vectors from weighted and directed Ethereum crypto-token networks, modeled as temporal graphs. The approach is tailored for graph anomaly prediction, a specialized form of graph classification.

The research problem is set as a prediction task where the authors aim to predict whether the absolute price return of an Ethereum token will change significantly beyond a predefined threshold $ |\delta| > 0 $ within the next $ h $ days. This involves analyzing the token's transaction network and its price fluctuations over multiple discrete time graph snapshots. The price information is sourced from external ground truth data, as token prices are determined by trading on blockchain exchanges.

For each snapshot, the authors use transferred token amounts on edges to define distances between adjacent vertices. These functions help  establish the similarity between nodes:
$\omega_{uv} = \left[1 + \alpha \cdot \frac{A_{uv} - A_{min}}{A_{max} - A_{min}}\right]^{-1}$ 
where $ A_{uv} $ represents the amount of tokens transferred between nodes $ u $ and $ v $, and $ A_{min} $ and $ A_{max} $ are the minimum and maximum transferred amounts, respectively. The authors set the parameter $\alpha = 9$ to map these weights to the interval $[0.1, 1]$, thus standardizing the weight values.

\begin{wrapfigure}{r}{3in}
\vspace{-.2in}
    \centering
	\subfloat[\footnotesize Tronix \label{fig:tronix}]{
        \includegraphics[width=.48\linewidth]{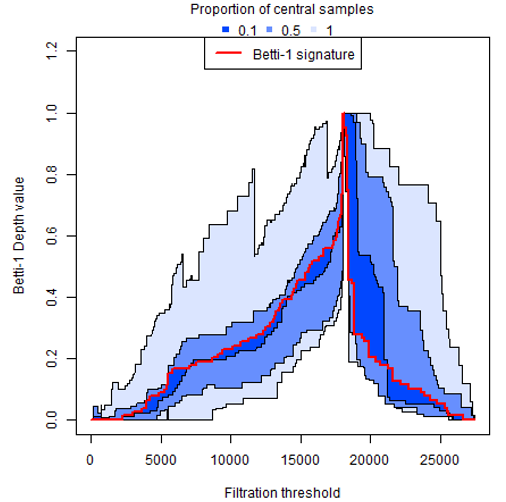}}
	\subfloat[\footnotesize Power Ledger \label{fig:power_ledger}]{
        \includegraphics[width=.5\linewidth]{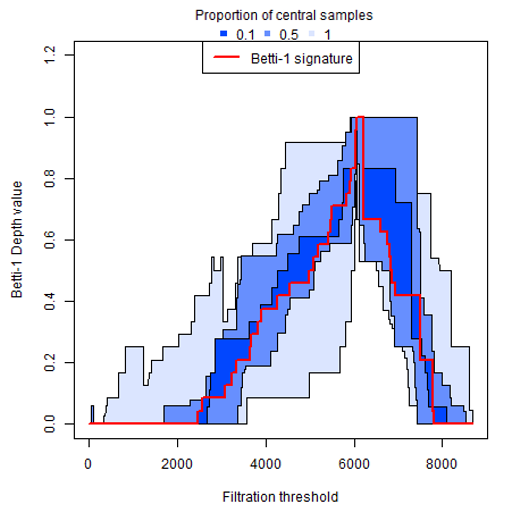}}
    \caption{\footnotesize {\bf Betti pivots.} Comparative functional summaries of the Tronix and Power Ledger token networks over seven days, displaying daily Betti-1 values across various scaling parameters. Each plot captures the evolution of topological features, with the central red lines indicating the Betti pivots. These pivots represent the most stable or 'normal' network structures, providing insights into network behavior and potential anomalies driven by underlying transaction activities. The figure is adapted from~\cite{li2020dissecting}.}
    \label{fig:mbd}
 \vspace{-.1in}
\end{wrapfigure}
By treating the edge weights (similarity measures) as distances between nodes, i.e., \( d(u,v) = \omega_{uv} \), the authors construct a filtration of Rips complexes (\Cref{sec:graph}-ii) from the snapshot graphs. For non-adjacent nodes, the distance is defined by the shortest path in the graph. The main insight here is to extract topological patterns developed in the graphs at multiple scales. This process involves forming a simplicial complex where nodes are connected if their distance is within a threshold \( \epsilon \), reflecting their similarity based on the edge weights. Essentially, nodes with high similarity (those with large transaction amounts between them) are connected early in the filtration, while more distant nodes appear later. 

The authors then compute persistence diagrams from these filtrations and convert them into Betti functions. Additionally, they introduce new functional summaries of topological descriptors, namely Betti limits and Betti pivots, which track the evolution of topological features as the scale parameter \( \epsilon \) changes over the snapshots. \Cref{fig:mbd} illustrates two token networks and their corresponding functional summaries.

To identify which transaction networks indicate anomalous patterns, the authors use Modified Band Depth to assess how central or peripheral each network's topological descriptors are within the observed data set. Figure~\ref{fig:mbd} shows two token networks and their snapshot graphs as represented by functions within these figures.  Snapshot graphs with deeper Betti limits are considered more typical or central.

The authors integrate these novel topological features with conventional network summaries to predict price anomalies. Daily labels are assigned as \textit{anomalous} based on significant price changes anticipated in the near future (e.g., in one or two days). To this end, the authors construct a predictive model by utilizing topological vectors they produce for token networks. \Cref{tab:tokenaccuracy} displays the accuracy metrics for their model, specifically for a prediction horizon of $h=2$ (two days). Achieving an average accuracy of 96\% across ten token networks, they demonstrate the effectiveness of topological features in the anomaly forecasting task.

\begin{table}[t]
    \centering
    \caption{Anomaly forecasting performance of crypto-token prices for two days horizon ($h=2$) for the top-10 tokens~\cite{li2020dissecting}.}
    \begin{tabular}{ccccccccccc}
    \toprule
    \textbf{Token} & Tronix & Omisego & Mcap & Storj & BNB & ZRX & CyberMiles & Vechain & Icon & BAT \\
    \midrule
    \textbf{M4 Accuracy} & 0.975 & 0.990 & 0.913 & 0.962 & 0.979 & 0.978 & 0.961 & 0.954 & 0.908 & 0.965 \\
    \bottomrule
    \end{tabular}
    \label{tab:tokenaccuracy}
\end{table}

\subsection{PH for Images: Cancer Diagnosis from Histopathological Images} \label{sec:histo}

Our example will directly apply persistent homology methods to histopathological images. As noted in \Cref{sec:image}, cubical persistence is the primary method for applying persistent homology in an image context. In Yadav et al. (2023)~\cite{yadav2023histopathological}, Yadav et al. successfully applied cubical persistence to analyze histopathological images. Specifically, for each image $\X$, the authors generated a topological feature vector $\wh{\beta}(\X)$ and utilized standard ML methods on these vectors (image embeddings) for tumor classification.

\begin{wrapfigure}{r}{3in}
\vspace{-.2in}
\centering
    \includegraphics[width=\linewidth]{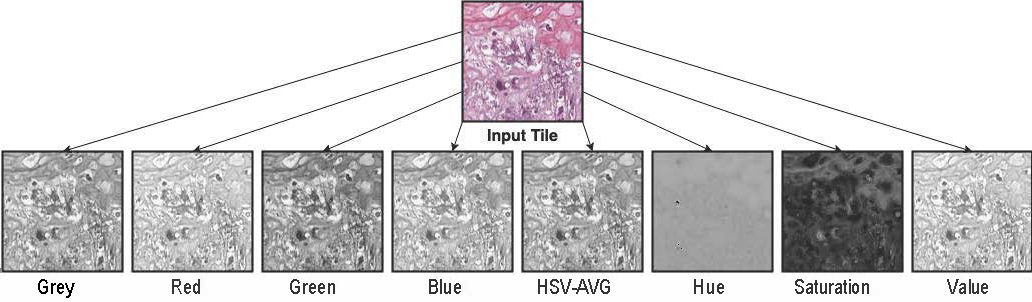}
    \caption{{\bf Eight Color Channels.} Different color channels are used to create sublevel filtrations. The figure is adapted from~\cite{yadav2023histopathological}.    \label{fig:color_channels}}
    \vspace{-.1in}
\end{wrapfigure}
Recall that for a given image $X$ with $r \times s$ resolution, the first step is to create a filtration, which is a nested sequence of binary images ${X_n}$. A common method to create such a sequence is to use the color values $\gamma_{ij}$ of each pixel $\Delta_{ij} \subset X$. Specifically, for a sequence of grayscale values $(t_1 < t_2 < \dots < t_N)$, one obtains a nested sequence of binary images $X_1 \subset X_2 \subset \dots \subset X_N$ such that $X_n = {\Delta_{ij} \subset X \mid \gamma_{ij} \leq t_n}$.

In~\cite{yadav2023histopathological}, the authors construct filtrations for cubical persistence by first extracting eight color channels $\gamma^k_{ij}$ from histopathological images with $1 \leq k \leq 8$, where each superscript $k$ corresponds to one color channel. The first four channels come from the RGB color space: red, green, blue, and grayscale (the average of R, G, and B). Additionally, they utilize the HSV color space, which includes hue, saturation, value, and their average (see~\Cref{fig:color_channels}). Each color channel defines a different filtration $\{\X_n^k\}_{n=1}^N$. In the paper, they set $N=100$.

\begin{figure}[t] 
	\centering
	\subfloat[\scriptsize Colon (Gray Betti-0) \label{fig:colon1}]{
	      \includegraphics[width=0.33\linewidth]{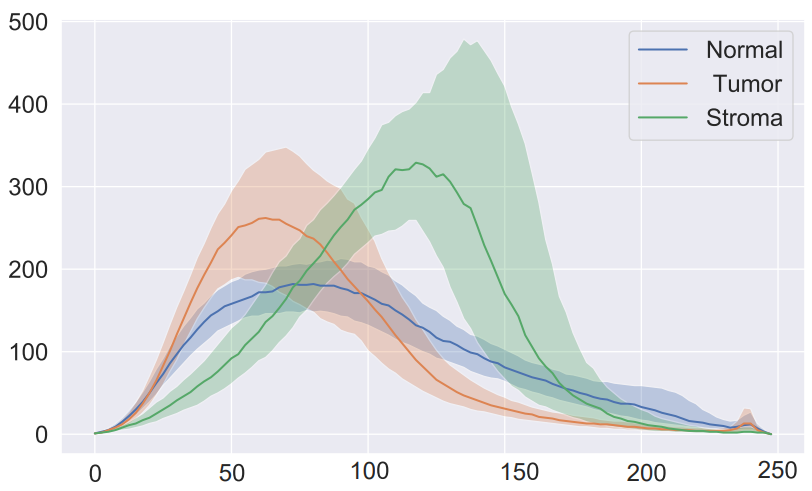}}
	\subfloat[\scriptsize Bone (Gray Betti-0) \label{fig:bone1}]{
		\includegraphics[width=0.33\linewidth]{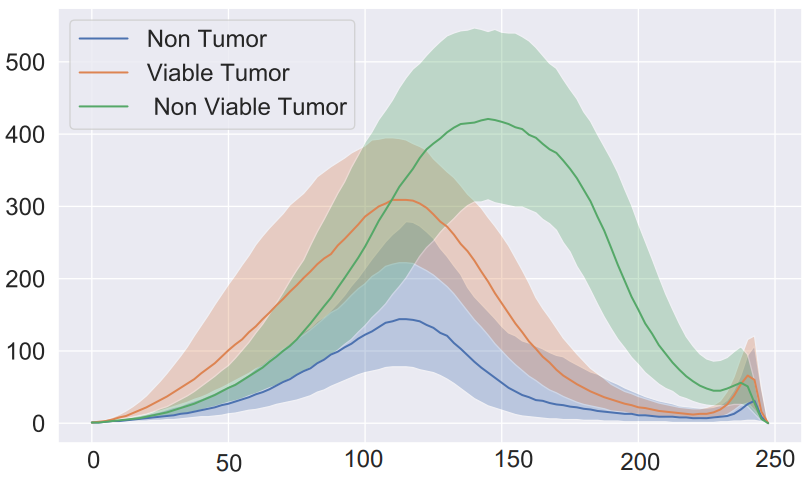}}
	\subfloat[\scriptsize Cervical (Value Betti-0) \label{fig:cervical0}]{
		\includegraphics[width=0.33\linewidth]{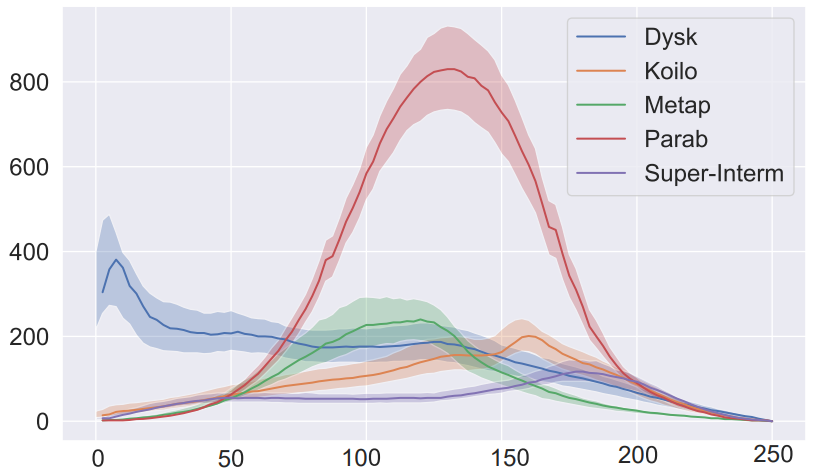}}
	\caption{Median curves and 40\% confidence bands of Betti vectors, one for each cancer type.  $x$-axis represents color values $t\in[0,255]$ and $y$-axis represents $\beta_0(t)$ (Betti-0), i.e., the count of components in the binary image $\X_t$. For details, see~\cite{yadav2023histopathological}.}
	\label{fig:Betti-curves}
 \vspace{-.15in}
\end{figure}
Next, by using these filtrations, they obtain the corresponding persistence diagrams for dimensions $0$ and $1$. Then, by applying Betti vectorization to these persistence diagrams, they obtain $100$-dimensional Betti vectors $\vec{\beta}^k_m=[\beta^k_m(t_1) \dots \beta^k_m(t_{100})]$ where $k$ represents the color, and $m=0,1$ represents the dimension. Hence, $\beta^k_0(t_n)$ is the number of components (Betti-$0$ number) in the binary image $\X^k_n$, and $\beta^k_1(t_n)$ is the number of holes (Betti-$1$ number) in the binary image $\X^k_n$. Considering that there are eight color channels and two dimensions, one obtains 1600-dimensional vector $\vec{\beta}(\X)$ by concatenating these 16 vectors $\{\vec{\beta}^k_m\}$. By extracting extra features by utilizing local binary patterns (LBP) and Gabor filters, they produced another 800 and 400 dimensional features for each image, respectively. Then, they obtain a 2800-dimensional final vector $\wh{\beta}(\X)$. In particular, one can consider this as an image embedding method, and each image is realized as a point in $\R^{2800}$. Next, to improve the performance by removing correlated features, they use a feature selection method for the downstream task and reduce the vector size to 500 dimensions. 

They studied five cancer types, namely bone, breast, cervical, prostate, and colon cancer. In \Cref{fig:Betti-curves}, they provide the median curves and confidence bands for each class for three cancer types. As discussed earlier, while Betti vectors are considered as a weak vectorization method in general, they can be highly effective when the signature comes from the quantity and distribution of small features like these histopathological images. They utilized benchmark datasets for each cancer type consisting of 20K to 60K histopathological images. In~\Cref{tab:histopathology}, we give their results for various sets of feature vectors. Their results indicate that utilizing filtrations with multiple color channels can significantly improve the classification results in some cancer types, e.g., bone, breast, and colon cancers.

\begin{table}[h!]
\centering
\caption{{\bf Cancer diagnosis with PH.} Multiclass accuracy results for various cancer types from histopathological images using a topological ML model with a random forest classifier applied to different feature vector sets. The number of classes is indicated in parentheses. The first five rows correspond to the performance of Betti vectors across different color channels~\cite{yadav2023histopathological}. \label{tab:histopathology}}
\setlength\tabcolsep{3pt}
\begin{tabular}{lcccccc}
\toprule
\textbf{Features} & \textbf{\# Features} &  \textbf{Bone} (3) &   \textbf{Breast} (4)&   \textbf{Prostate} (4)&   \textbf{Cervical} (5)&   \textbf{Colon} (3)\\
    \midrule
Gray & 200               	&	84.6	&	68.8	&	91.1	&	54.6	&	 95.9 \\	
G-RGB & 800               	&	89.4	&	80.0	&	93.2	&	77.5	&	 97.5 \\	
avg HSV &200            	&	82.3	&	60.8	&	84.5	&	60.7	&	 88.0 \\	
HSV+ avg HSV & 800         	&	90.1	&	78.7	&	93.0	&	82.7	&	 97.5 \\	
All Betti features &1600 	&	90.8	&	82.8	&	93.9	&	85.5	&	 97.8  \\	
Betti + Gabor + LBP & 2800       	&	93.7	&	88.4	&	94.7	&	\textbf{92.6}	&	 \textbf{98.5} \\ \midrule	
Feature Selection  & 500    &	\textbf{94.2}	&	\textbf{91.6}	&	\textbf{95.2}	&	91.4	&	 98.4 \\ \bottomrule	
\end{tabular}
\end{table}

\subsection{Multiparameter Persistence: Computer Aided Drug Discovery} \label{sec:drug}

In this part, we outline an effective application of multiparameter persistence in computer-aided drug discovery (CADD), showcasing its use within the graph setting. For an application in the context of point clouds, see~\cite{vipond2021multiparameter, carriere2020multiparameter, loiseaux2024stable}.

Virtual screening (VS), a key technique in CADD, is used to identify potential drug candidates from a vast library of compounds that are most likely to bind to a specific molecular target. Demir et al.~\cite{demir2022todd} employed a multiparameter persistence approach for virtual screening by framing it as a graph classification task. In this method, compounds are represented as graphs, with atoms as nodes and bonds as edges. They adopted a ligand-based approach, wherein a few positive samples are provided for a given protein target, and the goal is to screen the compound library to find compounds that are most similar to these positive samples. Essentially, this approach can be viewed as a topology-based graph ranking problem. While their approach is more technical, here we outline the key concepts of their method.

 \begin{wrapfigure}{r}{0.25\textwidth} 
    \vspace{-0.3cm}
  \begin{center}
    \includegraphics[width=\linewidth]{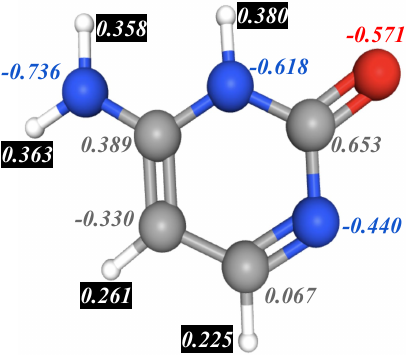}
    \caption{\footnotesize {\bf Cytosine}. Atom types are coded by their color: White=Hydrogen, Gray=Carbon, Blue=Nitrogen, and Red=Oxygen. The decimal numbers next to atoms represent their partial charges. The figure is adapted from~\cite{demir2022todd}.}
    \label{fig:cytosine}
  \end{center}
  \vspace{-0.5cm}
\end{wrapfigure} 

\begin{figure}[b]
\vspace{-.1in}
\centering
\includegraphics[width=0.8\textwidth]{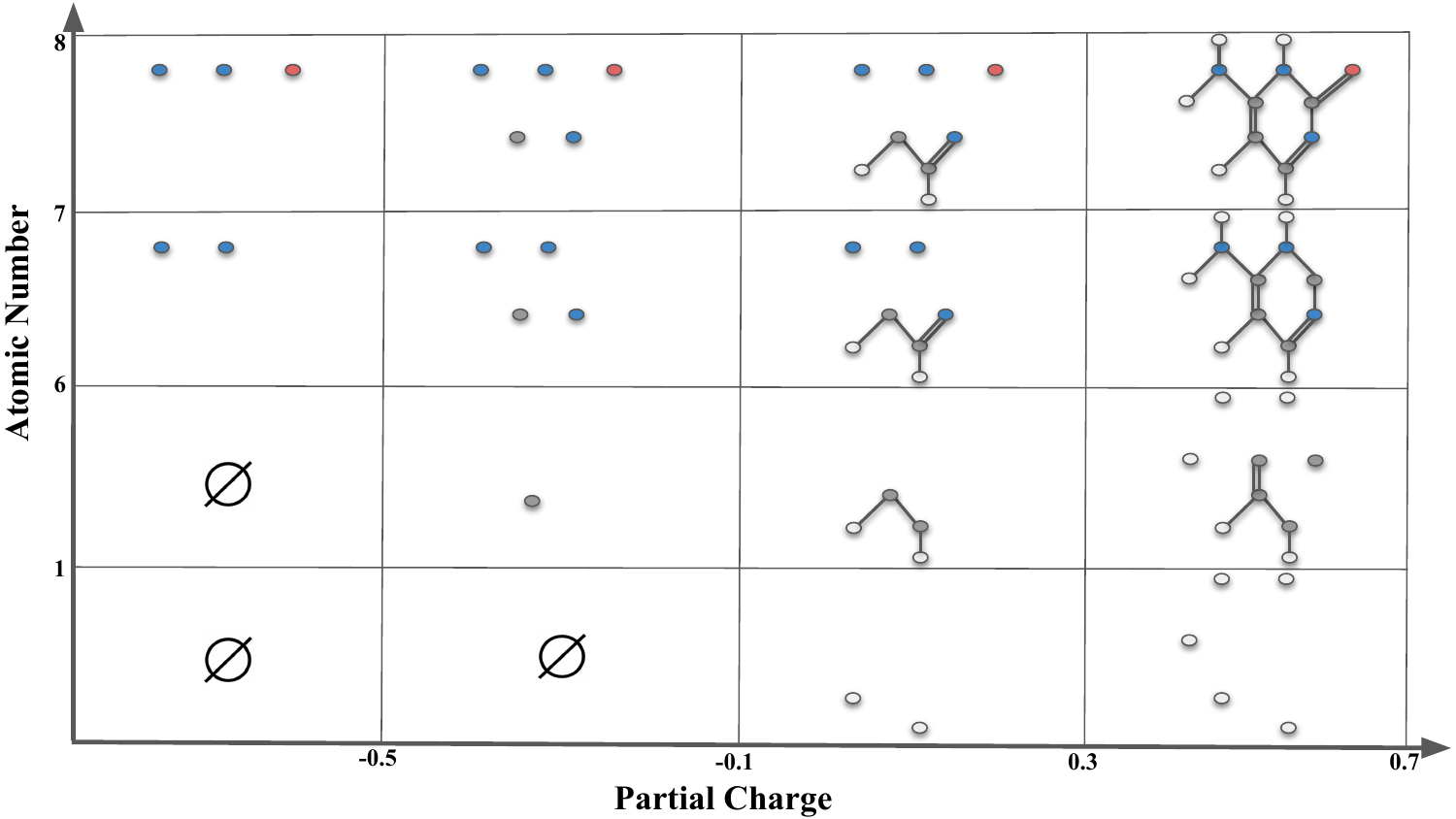}
\caption{\footnotesize \textbf{Sublevel bifiltration of cytosine} is induced by filtering functions atomic charge $f$ and atomic number $g$. In the horizontal direction, thresholds $\alpha=-0.5,-0.1, +0.3,+0.7$ filters the compound into substructures $f(v)\leq \alpha$ with respect to their partial charge. In the vertical direction, thresholds $\beta=1,6,7, 8$ filters the compound in the substructures $g(v)\leq \beta$ with respect to atomic numbers. The figure is adapted from~\cite{demir2022todd}.} 
\label{fig:MP-sublevel}
\end{figure}

Their framework, TODD, generates fingerprints of compounds as multidimensional vectors (tensors), represented as a $2D$ or $3D$ array for each compound. The core idea is to simultaneously employ 2 or 3 highly relevant functions or weights (e.g., atomic mass, partial charge, bond type, electron affinity, distance) to obtain a multifiltration, which decomposes the original compound into substructures using these relevant chemical functions. As detailed in~\Cref{sec:MP-graph}, for a given compound $\G=(\V,\E)$, they used atomic number $A$ and partial charge $P$ as node functions $A:\V\to\R$ and $P:\V\to\R$, as well as bond strength $B$ as an edge function $B:\E\to\R$, to define multifiltrations. An illustration of such a multifiltration for the compound cytosine~\Cref{fig:cytosine} is provided in~\Cref{fig:MP-sublevel}. In this example, the functions are partial charge and atomic number.

After constructing multifiltrations, they obtained compound fingerprints using a slicing technique. In particular, within each multifiltration ${\wh{\G}{ij}}$, they took horizontal slices by fixing a row $i_0$ to get a single persistence filtration ${\wh{\G}{i_0j}}$. For each horizontal slice $i_0$, they then generated a persistence diagram $\PD_k({\wh{\G}{i_0j}})$. These persistence diagrams were vectorized using Betti and Silhouette vectorizations to produce 2D arrays, such as $\mathbf{M}\beta=[\beta(\wh{\G}{ij})]$ (Betti numbers of the clique complex $\wh{\G}{ij}$).

Next, using these 2D arrays, they applied two ML classifiers: Random Forest and ConvNeXt Vision Transformers. Both classifiers performed exceptionally well, surpassing all state-of-the-art models on benchmark datasets. In~\Cref{tab:todd2}, their results are shown for the DUD-E Diverse dataset, which comprises 116K compounds targeting eight proteins. The common metric in virtual screening is \textit{enrichment factor}, which compares the proportion of active compounds found in the top-ranked subset of a model output to the proportion of active compounds in the entire dataset. $EF_{1\%}$   represents the enrichment factor for the top $1\%$ of the dataset.

\begin{table*}[t]
\centering
\caption{\footnotesize {\bf TODD vs. SOTA.} EF 1\% performance comparison between ToDD and the top-performing method among 12 SOTA baseline methods on eight targets from the DUD-E Diverse subset. \label{tab:todd2}}
\setlength\tabcolsep{4.5 pt}
\footnotesize
\begin{tabular}{lccccccccc}
\toprule
\textbf{Model} & \textbf{AMPC} &\textbf{CXCR4} & \textbf{KIF11} &\textbf{CP3A4} &\textbf{GCR} &\textbf{AKT1} &\textbf{HIVRT} &\textbf{HIVPR} &\textbf{Avg.}\\
\midrule
Best Baseline & 39.6 & 61.1 & 54.3 & 44.3 & 40.9 & 89.4 & 43.8 & 65.7 & 47.9 \\
\textbf{ToDD} &42.9 & 92.3 & 75.0  & 67.6 & 78.9 & 90.7  & 64.1  & 92.1 & 73.7 \\
\midrule
Relative gains & 8.3\% & 51.1\% & 38.1\% & 52.6\% & 92.9\%& 1.5\% & 46.3\% & 40.2\% & 52.8\% \\
\bottomrule
\end{tabular}
\end{table*}

\begin{figure}[b]
\centering
    \includegraphics[width=0.8\textwidth]{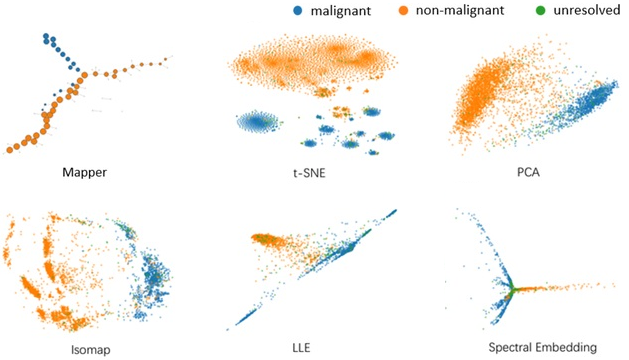}
    \caption{{\bf Visualization Methods}. Visualization of melanoma cells in the expression space with different dimension reduction algorithms. The figure is adapted from~\cite{wang2018topological}. \label{fig:mapper}}
\end{figure}

\subsection{Mapper: Cancer Genotyping from RNA Sequencing} \label{sec:mapper-app}

In this section, we will explore a significant application domain for the Mapper algorithm: the analysis of single-cell RNA sequencing data. Recall that Mapper is particularly effective for analyzing high-dimensional point clouds by generating a low-dimensional graph summary that preserves local relationships (\Cref{sec:mapper}). In the Mapper summary graph, the nodes represent clusters in the point cloud, and the edges between the nodes indicate that the corresponding clusters are nearby (interacting) in the high-dimensional space.

RNA sequencing data is crucial for cancer genotype analysis, though it presents significant challenges. The expression profile of a cell can be mathematically represented as a point in a high-dimensional expression space, where each dimension corresponds to the RNA level of a gene, and the dimension of the space is the number of expressed genes. Points that are close to each other in this space correspond to cells with similar expression profiles. The set of all possible tumors of a cancer type spans a subspace of the expression space. From an ML perspective, this is a highly sparse point cloud in a high-dimensional space. In general, cells are assigned to some cancer subtype (e.g., malignant, benign, etc.), and the aim is to understand the relation of these types in this high-dimensional space.

\begin{figure}[t]
\centering
    \includegraphics[width=0.8\textwidth]{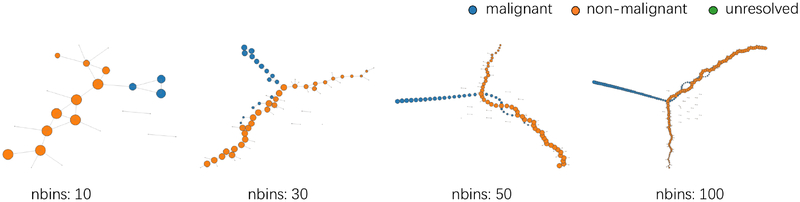}
    \caption{\textbf{Mapper Graph for Varying Bin Sizes.} As the number of bins increases, the clusters become smaller, resulting in an increased number of nodes (clusters). The figure is adapted from~\cite{wang2018topological}. \label{fig:mapper-bin}}
\end{figure}

\begin{wrapfigure}{r}{3in}
\vspace{-.15in}
\centering
    \includegraphics[width=\linewidth]{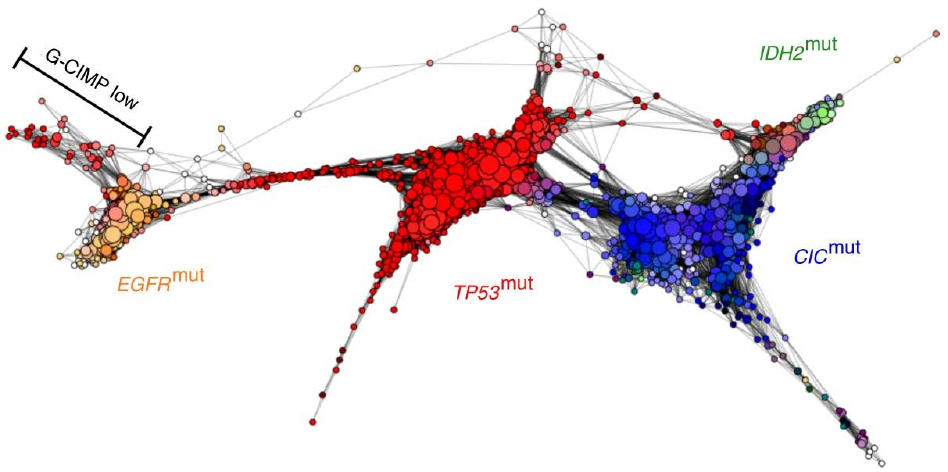}
    \caption{\textbf{Genetic Alterations.} Topological representation of the expression space for low-grade gliomas using Mapper. The structure reveals three main clusters corresponding to distinct glioma subtypes. One cluster, labeled IDH2\textsuperscript{mut}, highlights the significance of the IDH2 gene in differentiating this subtype. This visualization emphasizes the distinct expression profiles that separate these glioma subtypes. This figure is adapted from~\cite{rabadan2020identification}.\label{fig:mapper-genetic}}
\end{wrapfigure}
There are several significant works on cancer genotyping through the use of Mapper techniques~\cite{rabadan2020identification,rizvi2017single,rafique2020topological}. In this paper, we will detail the methods in two works. In the first one~\cite{wang2018topological}, Wang et al. conducted a comparative analysis of Mapper visualization techniques on RNA-sequencing data. They used the melanoma tumor cells dataset GSE72056, which contains 4,645 cells. Of these, 1,257 are malignant, and 3,256 are non-malignant, with some minority classes included. The dataset includes 23,686 expressed genes. From an ML perspective, this data can be represented as a point cloud $\X = \{x^i\}_{i=1}^{4645}$ in very high dimensional space $\R^{23,686}$, where each dimension corresponds to a gene $\{\gamma_j\}_{j=1}^{23,686}$. In particular, for a cell $x^i \in \X$ with $x^i = (x^i_1, x^i_2, \dots, x^i_{23,686})$, $x^i_j$ represents the RNA level of gene $\gamma_j$ in cell $x^i$. This is computed as $x^i_j = \log_2\left(1 + \frac{\text{TPM}_{ij}}{10}\right)$, where $\text{TPM}_{ij}$ is the transcript-per-million (TPM) for gene $\gamma_j$ in cell $x^i$. Notice that the number of dimensions far exceeds the number of points. From an ML perspective, this presents a highly challenging setup. In~\Cref{fig:mapper}, the authors illustrate different dimension reduction techniques on this dataset. In~\Cref{fig:mapper-bin}, they show how the graph changes when the number of bins (resolution) is varied.

In the second work~\cite{rabadan2020identification}, Rabadan et al. applied the Mapper algorithm to identify somatic mutations relevant to tumor progression. They analyzed mutation and RNA expression data for 4500 genes across 4476 patients from 12 tumor types, including low-grade glioma (LGG), lung adenocarcinoma, breast invasive carcinoma, and colorectal adenocarcinoma. This study led to the identification of 95 mutated cancer genes, 38 of which were previously unreported. They identified three large expression groups with oligodendrogliomas (enriched for CIC and IDH2 mutations), IDH1-mutant astrocytomas (enriched for TP53 mutations), and IDH1-wild-type astrocytomas (enriched for EGFR mutations)~(\Cref{fig:mapper-genetic}). In their analysis, they used the neighborhood lens (\Cref{sec:mapper}) as the filter function and scanned the resolution and gain parameter space to determine the optimal parameters~(\Cref{fig:mapper-bin}).

%% file: sections/7-Future.tex
In the preceding sections, we covered the primary topological methods employed in machine learning. In this section, we will explore future directions for advancing these methods, aiming to enhance the practicality of TDA in ML. We will also discuss strategies for expanding the application of topological methods into new and emerging areas.

\subsection{Topological Deep Learning}

While TDA is already an effective tool for feature extraction from complex data, recent research underscores its potential to augment deep learning models by providing complementary insights. Deep learning algorithms typically emphasize local data relationships, whereas TDA offers a global perspective, delivering supplementary information. 
Furthermore, TDA methods have also been integrated into deep learning algorithms to optimize its process by regulating the loss functions and other layers~\cite{hu2019topology,wang2020topogan}. 
In recent years, there has been significant progress in integrating TDA with deep learning in domains such as image analysis~\cite{singh2023topological,clough2020topological,peng2024phg}, biomedicine~\cite{luo2024improving}, graph representation learning~\cite{immonen2024going,chen2021topological}, genomics~\cite{amezquita2023genomics}, cybersecurity~\cite{guo2022gld} and time-series forecasting~\cite{coskunuzer2024time,shamsi2024graphpulse}. This is a rapidly emerging field, with several recent papers surveying the current state-of-the-art~\cite{papamarkou2024position,zia2024topological,hajij2022topological,papillon2023architectures}.

 \subsection{TDA and Interpretability}

TDA, though often used as a feature extraction technique in machine learning, provides a powerful framework for enhancing model interpretability by capturing the inherent geometric and topological structures of data. Tools like PH and Mapper excel at identifying and quantifying features such as clusters, loops, and voids across multiple scales, unveiling intricate patterns that might escape traditional analytic methods~\cite{krishnapriyan2021machine,cole2021quantitative,mukherjee2022determining}. By utilizing these topological insights into ML pipelines, TDA offers a unique perspective on decision boundaries, feature interactions, and model behavior. For instance, Mapper facilitates the visualization and interpretation of complex, high-dimensional datasets by projecting them into lower-dimensional topological spaces\cite{nicolau2011topology,purvine2023experimental,rabadan2019topological,spannaus2024topological}. This transformation uncovers relationships within the data that are otherwise difficult to grasp, making it easier to interpret the reasoning behind certain predictions or classifications. Ultimately, by incorporating TDA, researchers and practitioners can clarify the decision-making processes of machine learning models, promoting greater transparency and trustworthiness in AI systems.

\subsection{Fully Automated TDA Models}

While PH and Mapper have demonstrated effectiveness across several machine learning domains, their successful application still requires considerable expertise, such as hyperparameter tuning and selecting appropriate filtration functions. To make PH and Mapper more accessible to the broader ML community, there is a pressing need for end-to-end algorithms that automate these processes. For PH, this automation can be tailored to specific data formats, such as point clouds, images, or graphs. In each case, the fully automated algorithm would handle the selection of filtration functions, thresholds, vectorization methods, and other hyperparameters for optimal performance in downstream tasks. For instance, an automated PH model designed for graph data could automatically choose the best filtration function (learnable), appropriate threshold values, and vectorization techniques to achieve optimal results in a node classification task. Similarly, a fully automated Mapper algorithm that selects filtration functions, resolution, gain, and other hyperparameters would significantly enhance its usability and accessibility to the ML community. For PH, there are works for learnable filtration functions, which find the best filtration functions in graph setting~\cite{hofer2020graph,zhang2022gefl}. On the other hand, For Mapper, there is promising progress in this direction with Interactive Mapper~\cite{zhou2021mapper}. The broader adoption of TDA hinges on developing practical software libraries that streamline its use for various applications.

\subsection{Scalability of TDA methods}

Another key barrier to the widespread adoption of TDA methods in various application areas is their high computational cost. Although TDA performs well on small to medium datasets, scaling these methods to larger datasets poses significant challenges due to computational demands. Several strategies have been introduced to alleviate these costs, varying by data type. For graph datasets, recent research \cite{akcora2022reduction, yan2022neural} has proposed practical techniques that substantially lower the computational burden of PH. Similarly, for point clouds, recent works~\cite{de2022ripsnet, zhou2022learning, malott2020topology} present algorithms that improve the efficiency of PH computations. While these advances are promising, there remains a critical need to enhance the scalability of PH for large graph and point cloud datasets. In contrast, cubical persistence is already computationally efficient for image datasets and can be applied to large datasets without significant cost concerns. Even in the case of 3D image datasets \cite{gonzalez2016fast}, PH provides a cost-effective alternative to deep learning approaches. Similarly, scalability issues are minimal for Mapper when hyperparameters are appropriately tuned.

%% file: sections/8-appendix.tex
\section{Notation Table}

\begin{table}[h!]
\footnotesize
\centering
\caption{Notation and Main Symbols \label{notations}}
\begin{tabular}{ccc}
\toprule
\textbf{Notation} & \textbf{Description} & \textbf{Details} \\
\midrule
$(\X, d)$ & Topological space with metric $d$ & $d(x, y)$ defines the distance between points $x$ and $y$ \\
$\chi(\C)$ & Euler characteristics of $\C$ & Topological invariant to characterize $\C$ \\
$\h_k(\X)$ & $k$-th homology group of $\X$ & Describes $k$-dimensional holes in $\X$ \\
$\C_k(\X)$ & $k$-chains of $\X$ & All $k$-subcomplexes in $\X$ \\
$\Z_k(\X)$ &  $k$-cycles of $\X$ & $k$-subcomplexes in $\X$ without boundary \\
$\B_k(\X)$ & $k$-boundaries of $\X$ & Boundaries of $(k+1)$-subcomplexes \\
$\partial_k$ & Boundary operator & Maps $\C_k(\X)$ to $\C_{k-1}(\X)$\\
$\I=\{\e_i\}$ & Threshold set & Used to define filtration \\
$\{\K_i\}$ & Filtration & Nested sequence of simplicial complexes \\
$\PD_k(\X)$ & $k$-th persistence diagram of $\X$ & Visualizes persistent features \\
$\PB_k(\X)$ & $k$-th persistence barcode of $\X$ & Alternate view of persistence \\
$\N_r(\X)$ & $r$-neighborhood of $\X$ & Forms covering space \\
$\VR_r(\X)$ & Rips complex of point cloud $\X$ & Built from pairwise distances \\
$\check\C_r(\X)$ & \v{C}ech complex of point cloud $\X$ & Constructed using open balls \\
$\G=(\V, \E)$ & Graph, vertices, edges & $\V$ are vertices, $\E$ are edges \\
$\W=\{w_{ij}\}$ & Edge weights of $\G$ & Assigns weight $w_{ij}$ to edge $e_{ij}$ \\
$\G^k$ & $k$-th power of graph $\G$ & Higher-order graph structure \\
$\wt{\G}$ & Clique complex of $\G$ & Contains all maximal cliques \\
$\Delta_{ij}$ & Pixel at $(i,j)$-coordinate & A pixel in an $m\times n$-size image \\
$\W_p(\cdot, \cdot)$ & Wasserstein distance & Measures distance between PDs  \\
$\W_\infty(\cdot, \cdot)$ & Bottleneck distance & Measures distance between PDs \\
$\vec{\beta}_k(\X)$ & Betti vector for $\PD_k(\X)$ & Summarizes homological features \\
$\beta_i(\X)$ &  $i^{th}$ Betti number of $\X$ & Rank of $\h_i(\X)$ \\
$\{\K_{ij}\}$ & Bifiltration & Nested sequence of spaces in two parameters\\
\bottomrule
\end{tabular}
\end{table}

\section{Dataset Resources}

As we develop this tutorial to introduce topological methods to the ML community, we also seek to highlight their real-world applications to the mathematics community. For those eager to gain hands-on experience with topological techniques using real-world datasets, Table~\ref{tab:datalibraries} lists the commonly utilized datasets. For task-specific datasets, you can explore the collections at~\url{https://paperswithcode.com/datasets} and \url{https://www.kaggle.com}.

\begin{table}[h!]
  \caption{\textbf{Datasets Resources.} The list of commonly used benchmark datasets in ML.}
    \label{tab:datalibraries}
    \centering
    \resizebox{1.\linewidth}{!}{
    \begin{tabular}{l c l l l}
        \toprule
        \textbf{Format} & \textbf{Dataset} & \textbf{Task} &  \textbf{Dataset URL} \\
        \midrule
 & ModelNet & Shape classification & \url{https://modelnet.cs.princeton.edu}\\
{\bf Point}& UCR& Time Series Classification& \url{https://www.cs.ucr.edu/~Eeamonn/time_series_data_2018/}\\
{\bf Cloud}& Broad Institute & Single Cell Genotyping & \url{https://singlecell.broadinstitute.org/single_cell} \\
& 3D-PointCloud & Various & \url{https://github.com/zhulf0804/3D-PointCloud}\\
\midrule
\multirow{5}{*}{\bf Graph} & Small Benchmarks & Graph Classification & \url{https://paperswithcode.com/task/graph-classification}\\
 & Small Benchmarks & Node Classification & \url{https://paperswithcode.com/task/node-classification}\\
  & Small Benchmarks & Link Prediction & \url{https://paperswithcode.com/task/link-prediction}\\
& OGB (Large Benchmarks) & Graph, Node, Link & \url{https://ogb.stanford.edu}\\
& TUDatasets (Small Benchmarks)& Graph, Node, Link & \url{https://chrsmrrs.github.io/datasets}\\
& TGB (Temporal Graphs) & Dynamic Prediction& \url{https://tgb.complexdatalab.com/}\\
\midrule
\multirow{4}{*}{\bf Image} &MNIST & Handwritten Digits & \url{https://yann.lecun.com/exdb/mnist}\\
&MedMNIST & Medical Image Classification & \url{https://medmnist.com/}\\
&Fashion MNIST &  Image Classification & \url{https://www.kaggle.com/datasets/zalando-research/fashionmnist}\\
& Various Collections& Image Classification& \url{https://paperswithcode.com/datasets?task=image-classification}\\
        \bottomrule
    \end{tabular}}
\end{table}